\documentclass[twoside,11pt]{article}
\usepackage{blindtext}
\usepackage{jmlr2e}
\usepackage[utf8]{inputenc}
\usepackage[T1]{fontenc}

\usepackage{microtype}
\usepackage{graphicx}
\usepackage{booktabs}
\usepackage[dvipsnames]{xcolor}

\definecolor{diff}{rgb}{0,0.0,1}

\usepackage{amsmath}
\usepackage{amssymb}
\usepackage{bm}
\usepackage{comment}
\usepackage{graphicx}
\usepackage{subcaption}
\usepackage{xcolor}
\usepackage{tikz}
\usetikzlibrary{shadows}
\usepackage{url}            
\usepackage{booktabs}       
\usepackage{amsfonts}       
\usepackage{nicefrac}      
\usepackage{microtype}      
\usepackage{mathtools}
\usepackage{dsfont}
\usepackage{mathrsfs} 

\usepackage{hyperref}
\usepackage{enumerate}
\usepackage{mathtools}
\DeclarePairedDelimiter{\ceil}{\lceil}{\rceil}

\usepackage{lastpage}

\def\xb{\boldsymbol{x}}

\def\zb{\boldsymbol{z}}

\def\Zb{\boldsymbol{Z}}

\def\Wb{\boldsymbol{W}}
\def\Cb{\boldsymbol{C}}
\def\Ub{\boldsymbol{U}}
\def\Bb{\boldsymbol{B}}
\def\Qb{\boldsymbol{Q}}

\def\zetab{\boldsymbol{\zeta}}

\def\Ab{\boldsymbol{A}}
\def\Hb{\boldsymbol{H}}

\def\vb{\boldsymbol{v}}

\def\EE{\mathbb{E}}
\def\Con{\mathcal{C}}

\def\Fb{\boldsymbol{F}}
\def\Risk{\mathcal{R}}

\def\Ball{\mathbb{B}}

\def\rhoz{\rho_{\zb}}
\def\rhoa{\rho_{a}}
\def\rhoW{\rho_{W}}

\def\Pcal{\mathcal{P}}
\def\Hilb{\mathcal{H}}

\newcommand{\xbg}[1]{\xb'_{#1}}

\def\rhoz{\rho_{\zb}}
\def\rhoa{\rho_{a}}
\def\rhoW{\rho_{W}}

\def\Pcal{\mathcal{P}}
\def\Hilb{\mathcal{H}}

\newcommand{\summ}[2]{\sum_{#1=1}^{#2}}

\newcommand{\sigbig}[1]{\sigma \big ( #1 \big )}
\newcommand{\sig}[1]{\sigma ( #1 )}
\newcommand{\sigpbig}[1]{\sigma' \big ( #1 \big )}

\def\Rbb{{\mathbb{R}}}
\def\Nbb{{\mathbb{N}}}

\def\domX{\mathcal{X}}

\def\xb{\boldsymbol{x}}

\def\zb{\boldsymbol{z}}

\def\Zb{\boldsymbol{Z}}

\def\Wb{\boldsymbol{W}}
\def\Cb{\boldsymbol{C}}
\def\Ub{\boldsymbol{U}}

\def\Xib{\boldsymbol{\Xi}}
\def\zetab{\boldsymbol{\zeta}}

\def\Ab{\boldsymbol{A}}
\def\Hb{\boldsymbol{H}}

\def\vb{\boldsymbol{v}}

\def\taub{\boldsymbol{\tau}}
\def\EE{\mathbb{E}}
\def\Gfun{\kappa}

\def\Rad{\text{Rad}}

\def\rhoz{\rho_{\zb}}
\def\rhoa{\rho_{a}}
\def\rhoW{\rho_{W}}

\def\Pcal{\mathcal{P}}
\def\Hilb{\mathcal{H}}

\def\Uilb{\mathcal{U}}

\def\EE{\mathbb{E}}
\def\PP{\mathbb{P}}

\def\Risk{\mathcal{R}}

\newcommand{\xbtrk}[1]{\xb_{#1}}

\def\ddt{\tfrac{d}{dt}}

\def\rhoz{\rho_{\zb}}
\def\rhoa{\rho_{a}}
\def\rhoW{\rho_{W}}

\newcommand{\expenu}[2]{\mathcal{E}_{#1} \left \{ #2 \right \}}
\newcommand{\expenus}[2]{\mathcal{E}_{#1} \{ #2 \}}
\newcommand{\expenub}[2]{\mathcal{E}_{#1} \big \{ #2 \big \}}
\newcommand{\expenuB}[2]{\mathcal{E}_{#1} \Big \{ #2 \Big \}}

\def\EEE{\mathbb{E}}
\newcommand{\EEEs}[1]{\EEE [ #1 ]}
\newcommand{\EEEb}[1]{\EEE \big [ #1 \big ]}
\newcommand{\EEEB}[1]{\EEE \Big [ #1 \Big ]}
\newcommand{\EEEbb}[1]{\EEE \bigg [ #1 \bigg ]}
\newcommand{\EEEBB}[1]{\EEE \Bigg [ #1 \Bigg ]}
\newcommand{\EEElr}[1]{\EEE \left [ #1 \right ]}

\newcommand{\EEElb}[2]{\EEE^{(#1)} \big [ #2 \big ]}

\def\Rbb{{\mathbb{R}}}
\def\Frob{\text{F}}

\newcommand{\zbjt}[2]{\zb_{#1, #2}}

\newcommand{\hitl}[3]{h_{#1, #2}^{(#3)}}
\newcommand{\hitlxb}[4]{h_{#1, #2}^{(#3)}(#4)}
\newcommand{\bitl}[3]{b_{#1, #2}^{(#3)}}
\newcommand{\qitl}[3]{q_{#1, #2}^{(#3)}}
\newcommand{\qitlxb}[4]{q_{#1, #2}^{(#3)}(#4)}
\newcommand{\Wijtl}[4]{W_{#1, #2, #3}^{(#4)}}
\newcommand{\Htl}[2]{\Hb_{#1}^{(#2)}}
\newcommand{\Qtl}[2]{\Qb_{#1}^{(#2)}}
\newcommand{\Btl}[2]{\Bb_{#1}^{(#2)}}
\newcommand{\Xitl}[2]{\boldsymbol{\Xi}_{#1}^{(#2)}}
\newcommand{\Hdtl}[2]{H_{#1}^{(#2)}}
\newcommand{\Qdtl}[2]{Q_{#1}^{(#2)}}
\newcommand{\Bdtl}[2]{B_{#1}^{(#2)}}
\newcommand{\Xidtl}[2]{\Xi_{#1}^{(#2)}}

\newcommand{\btl}[2]{b_{#1}^{(#2)}}

\newcommand{\tilait}[2]{\tilde{a}_{#1, #2}}
\newcommand{\tilzbjt}[2]{\tilde{\zb}_{#1, #2}}

\newcommand{\tilhitl}[3]{\tilde{h}_{#1, #2}^{(#3)}}
\newcommand{\tilhitlxb}[4]{\tilde{h}_{#1, #2}^{(#3)}(#4)}
\newcommand{\tilbitl}[3]{\tilde{b}_{#1, #2}^{(#3)}}
\newcommand{\tilqitl}[3]{\tilde{q}_{#1, #2}^{(#3)}}
\newcommand{\tilqitlxb}[4]{\tilde{q}_{#1, #2}^{(#3)}(#4)}
\newcommand{\tilWijtl}[4]{\tilde{W}_{#1, #2, #3}^{(#4)}}

\newcommand{\kappaker}[5]{\gamma_{#1, #2}^{(#3)}(#4, #5)}

\def\ladder{\mathcal{F}}
\newcommand{\Ladder}[1]{\mathcal{F}^{(#1)}}

\newcommand{\Hilbl}[1]{\Hilb^{(#1)}}
\newcommand{\Hilbmu}[1]{\Hilb_{#1}}

\newcommand{\zetamt}[1]{\zeta_{m, #1}}

\def\Law{\text{Law}}
\def\barron{\mathcal{B}}
\def\Comp{\mathscr{C}}
\newcommand{\CompL}[1]{\Comp^{(#1)}}
\newcommand{\CompLp}[2]{\CompL{#1}_{#2}}
\newcommand{\DompLp}[2]{\mathscr{D}^{(#1)}_{#2}}
\def\Conti{\mathcal{C}}

\usepackage[capitalize,noabbrev]{cleveref}

\usepackage[textsize=tiny]{todonotes}

\newtheorem{assumption}[theorem]{Assumption}

\title{Neural Hilbert Ladders: Multi-Layer Neural Networks in Function Space}

\author{\name Zhengdao Chen \email zhengdao.c3@gmail.com \\
      \addr Google Research\\
      Mountain View, CA 94043}

\date{}

\firstpageno{1}

\jmlrheading{25}{2024}{1-\pageref{LastPage}}{9/23; Revised
3/24}{3/24}{23-1225}{Zhengdao Chen}
\ShortHeadings{Neural Hilbert Ladders}{Chen}

\begin{document}

\editor{Lorenzo Rosasco}
\maketitle

\begin{abstract}
To characterize the function space explored by neural networks (NNs) is an important aspect of learning theory. In this work, noticing that a multi-layer NN generates implicitly a hierarchy of reproducing kernel Hilbert spaces (RKHSs)---named a neural Hilbert ladder (NHL)---we define the function space as an infinite union of RKHSs, which generalizes the existing Barron space theory of two-layer NNs. We then establish several theoretical properties of the new space. First, we prove a correspondence between functions expressed by $L$-layer NNs and those belonging to $L$-level NHLs. Second, we prove generalization guarantees for learning an NHL with a controlled complexity measure. Third, we derive a non-Markovian dynamics of random fields that governs the evolution of the NHL which is induced by the training of multi-layer NNs in an infinite-width mean-field limit. Fourth, we show examples of depth separation in NHLs under the ReLU activation function. Finally, we perform numerical experiments to illustrate the feature learning aspect of NN training through the lens of NHLs.
\end{abstract}


\section{Introduction}
\label{sec:intro}
The empirical success of deep learning has inspired many efforts to understand neural networks (NNs) theoretically, from aspects including approximation, optimization and generalization. In the basic setting of supervised learning, NNs can be seen as parameterizing a particular family of functions on the input domain, from which a solution can be found through training. Then, to explain what is special about NNs, it is crucial to investigate the space of functions (a.k.a. hypothesis class) they represent.

As modern NNs often involve a huge number of parameters, it is worthy of understanding the space of functions that can be represented by NNs with \emph{unlimited} width. As a foundational result, the universal approximation theorem (e.g., \citealp{cybenko1989approximation, hornik1989multilayer}) shows that, given enough width, NNs are capable of approximating virtually all reasonable functions, which suggests the vastness of this space.
A more interesting question, though, is to also find a \emph{complexity measure of functions} that quantify their representation cost in terms of the rate of approximation error, which would yield insights on what kind of functions are more naturally represented by NNs. This question has been studied fruitfully in the literature for shallow (a.k.a. two-layer) NNs \citep{barron1993universal, bengio_convex_2005, bach2017breaking, ma2018priori}, but remains mostly open for multi-layer NNs.

Meanwhile, to study the sample complexity of learning NNs, prior works have proved generalization guarantees that are based on \emph{not} the number of parameters but certain norms of them (e.g., \citealp{bartlett1998sample, neyshabur2015norm}), which could serve as a complexity measure of NNs from the \emph{generalization} point of view. Then, an important question is whether there is a complexity measure associated with width-unlimited multi-layer NNs that unifies the two perspectives of approximation and generalization.

Another central aspect of deep learning is the training of NNs, which involves a non-convex optimization problem but can often be solved sufficiently well by variants of gradient descent (GD). Although remarkable progress has been made to prove optimization guarantees for various settings, it remains intriguing what kind of exploration in \emph{function space} is induced by the training of NNs. The Neural Tangent Kernel (NTK) analysis provides a candidate theory via a linearized approximation of NN training \citep{jacot2018neural}, which implies that NNs represent functions in a particular reproducing kernel Hilbert space (RKHS) determined by their initialization. However, the NTK theory is unable to model the \emph{feature learning} that occurs in the training of actual NNs \citep{chizat2019lazy, woodworth2020kernel}, which is crucial to the success of deep learning.

Hence, this work is motivated by the following questions that are critical yet largely open: 
\begin{itemize}
    \item \emph{How to characterize the function space explored by the training of multi-layer NNs with arbitrary width?}
    \item \emph{Can we associate with it a complexity measure of functions that governs both approximation and generalization?}
\end{itemize}
In this work, by viewing an $L$-layer NN as a ladder of RKHSs with $L$-levels, we propose a function space $\Ladder{L}$ and a complexity measure $\CompL{L}$ that satisfy the following properties:
\begin{enumerate}[i.]
    \item \emph{Width unlimited}: $\Ladder{L}$ contains all functions that can be represented by an $L$-layer NN with arbitrarily-wide hidden layers; \label{item:1}
    \item \emph{Approximation guarantee}: Any function $f$ in $\Ladder{L}$ can be approximated by an $L$-layer NN with a cost that depends on $\CompL{L}(f)$; \label{item:2}
    \item \emph{Generalization guarantee}: Generalization errors can be upper-bounded for learning in $\Ladder{L}$ with $\CompL{L}$ under control; \label{item:3}
    \item \emph{Depth dependence}: There exist choices of $L$ and the activation function under which $\Ladder{L}$ is strictly smaller than $\Ladder{L+1}$. \label{item:5}
    \item \emph{Feature learning}: Gradient descent training of $L$-layer NNs in a feature-learning regime corresponds to a learning dynamics in $\Ladder{L}$; \label{item:4}
\end{enumerate}
To our knowledge, this is the first proposal satisfying all of the properties above---or even just \eqref{item:1} and \eqref{item:4} together---thus opening up a new perspective in the understanding of deep NNs. 

The rest of the paper is organized as follows. In Section~\ref{sec:nhl}, we introduce the neural Hilbert ladder (NHL) model and the NHL spaces and NHL complexities that it gives rise to. In Section~\ref{sec:real_approx}, we prove static correspondences between multi-layer NNs and NHLs, verifying \eqref{item:1} and \eqref{item:2}. In Section~\ref{sec:gen}, we prove generalization bounds for learning NHLs via a Rademacher complexity analysis, verifying \eqref{item:3}. In Section~\ref{sec:depth}, we show examples of depth separation in the NHL spaces under the ReLU activation function, verifying \eqref{item:5}. In Section~\ref{sec:training}, we show that the training of multi-layer NNs in the mean-field limit translates to a particular learning dynamics of the NHL, verifying \eqref{item:4}, and it is further illustrated by numerical results on synthetic tasks. Prior literature is discussed in Section~\ref{sec:related}.
\footnote{Part of this work was presented at the \emph{40th International Conference on Machine Learning} \citep{chen2023multi}.}

\section{Background}
\subsection{Basic Notations and Definitions}
We use bold lower-case letters (e.g. $\xb$ and $\zb$) to denote vectors and bold upper-case letters (e.g. $\Ub$ and $\Hb$) to denote random variables or random fields. $\forall m \in \mathbb{N}_+$, we write $[m] \coloneqq \{1, ..., m\}$. When the indices  $i$, $j$, $t$ and $s$ and the variables $\xb$ and $\xb'$ appear without being further defined, they are considered by default as under the universal quantifiers ``$\forall i, j \in [m]$'', ``$\forall t, s \geq 0$'' and ``$\forall \xb, \xb' \in \domX$''. 

Suppose $\mathcal{U}$ is a measurable space, and we let $\Pcal(\mathcal{U})$ denote the the space of probability measures on $\mathcal{U}$. $\forall \mu \in \Pcal(\mathcal{U})$, we let $L^2(\mathcal{U}, \mu)$ denote the space of square-integrable functions on $\mathcal{U}$ with respect to $\mu$, and $\forall \xi \in L^2(\mathcal{U}, \mu)$, we write $\| \xi \|_{L^2(\mathcal{U}, \mu)} \coloneqq (\int |\xi(u)|^2 \mu(du))^{1/2}$. If $\Ub$ is a $\mathcal{U}$-valued random variable, we let $\Law (\Ub) \in \Pcal(\mathcal{U})$ denote its law and let $\EEE[\phi(\Ub)] = \int \phi(u) [\Law (\Ub)](du)$ denote the expectation of any measurable function $\phi: \mathcal{U} \to \Rbb$ applied to $\Ub$. Suppose additionally that $\mathcal{U}$ is equipped with a norm (or quasi-norm) $\| \cdot \|_{\mathcal{U}}$. We define $\Ball(\mathcal{U}; M) \coloneqq \{ u \in \mathcal{U}: \| u \|_{\mathcal{U}} \leq M \}$ for $M > 0$, write $\Ball(\mathcal{U}) \coloneqq \Ball(\mathcal{U}; 1)$ for the unit ball in $\mathcal{U}$, and let $\hat{\mathcal{U}}\coloneqq \{ u \in \mathcal{U}: \| u \|_{\mathcal{U}} =1 \}$ denote the unit sphere in $\mathcal{U}$. 

For $N \in \mathbb{N}_+$, we let $\text{Lip}(\Rbb^N)$ denote the space of functions on $\Rbb^N$ with Lipschitz constant at most $1$. 
We say a function $f: \Rbb \to \Rbb$ is  \emph{non-negatively homogeneous with degree} $1$ (or \emph{$1$-homogeneous} for short)  if $\forall u \in \Rbb$, $a \geq 0$, $f(au) = a f(u)$. An example is the \emph{regularized linear unit} (\emph{ReLU}), defined as $\text{ReLU}(u) = \max\{u, 0\}$, a popular choice for the activation function in NNs.

Let $\domX$ denote the \emph{input domain}, which we assume throughout the paper to be a compact subset of $\Rbb^d$. We let $\mathcal{C}$ denote the space of continuous functions on $\domX$ equipped with the sup-norm $\| f \|_\infty \coloneqq \sup_{\xb \in \domX} |f(\xb)|$ and the associated Borel sigma-algebra. For $\nu \in \Pcal(\domX)$ and $f \in \mathcal{C}$, we write $\expenu{\xb \sim \nu}{f(\xb)} \coloneqq \int_\domX f(\xb) \nu(d\xb)$. The $d \times d$ identity matrix is denoted by $I_d$.

\subsection{Multi-Layer Neural Network (NN)}
\label{sec:nn} 
We consider an $L$-layer \emph{(fully-connected) NN} with width $m$ as expressing a function on $\domX$ of the following form:
\begin{equation}
\label{eq:f_m}
    f_{m}(\xb) \coloneqq \frac{1}{m} \summ{i}{m} a_i \sigbig{h_{i}^{(L-1)}(\xb)}~,
\end{equation}
where we define $h_i^{(1)}(\xb) \coloneqq \zb_i^{\intercal} \cdot \xb = \summ{j}{d} z_{i,j} x_{j}$, 
and $\forall l \in [L-2]$~,
\begin{equation}
\label{eq:h_i}
    h_i^{(l+1)}(\xb) \coloneqq \frac{1}{m} \summ{j}{m} W_{i, j}^{(l)} \sigbig{h_j^{(l)}(\xb)}~.
\end{equation}
Here, $\sigma: \Rbb \to \Rbb$ is the \emph{activation function}, and each $z_{i,j}, W^{(l)}_{i, j}$ and $a_i$ is a weight parameter of the \text{input} layer, the $l$th \text{middle} layer and the \text{output} layer, respectively. For simplicity, we will omit the bias term in the main text and discuss its inclusion in Appendix~\ref{app:bias}. We refer to $h_i^{(l)}$ as the \emph{pre-activation function} represented by the $i$th neuron in the $l$th hidden layer.

The $1/m$ factor in \eqref{eq:h_i} is often called the \emph{mean-field} scaling, which allows large $m$ limits to be considered while the parameters stay scale-free. Unlike the NTK scaling, the mean-field scaling allows feature learning to occur, including in the infinite-width limit \citep{yang2021feature}. The comparison with NTK and other scaling choices are further discussed in Section~\ref{sec:related}.

\begin{figure}[t]
    \centering
    \includegraphics[scale=0.14]{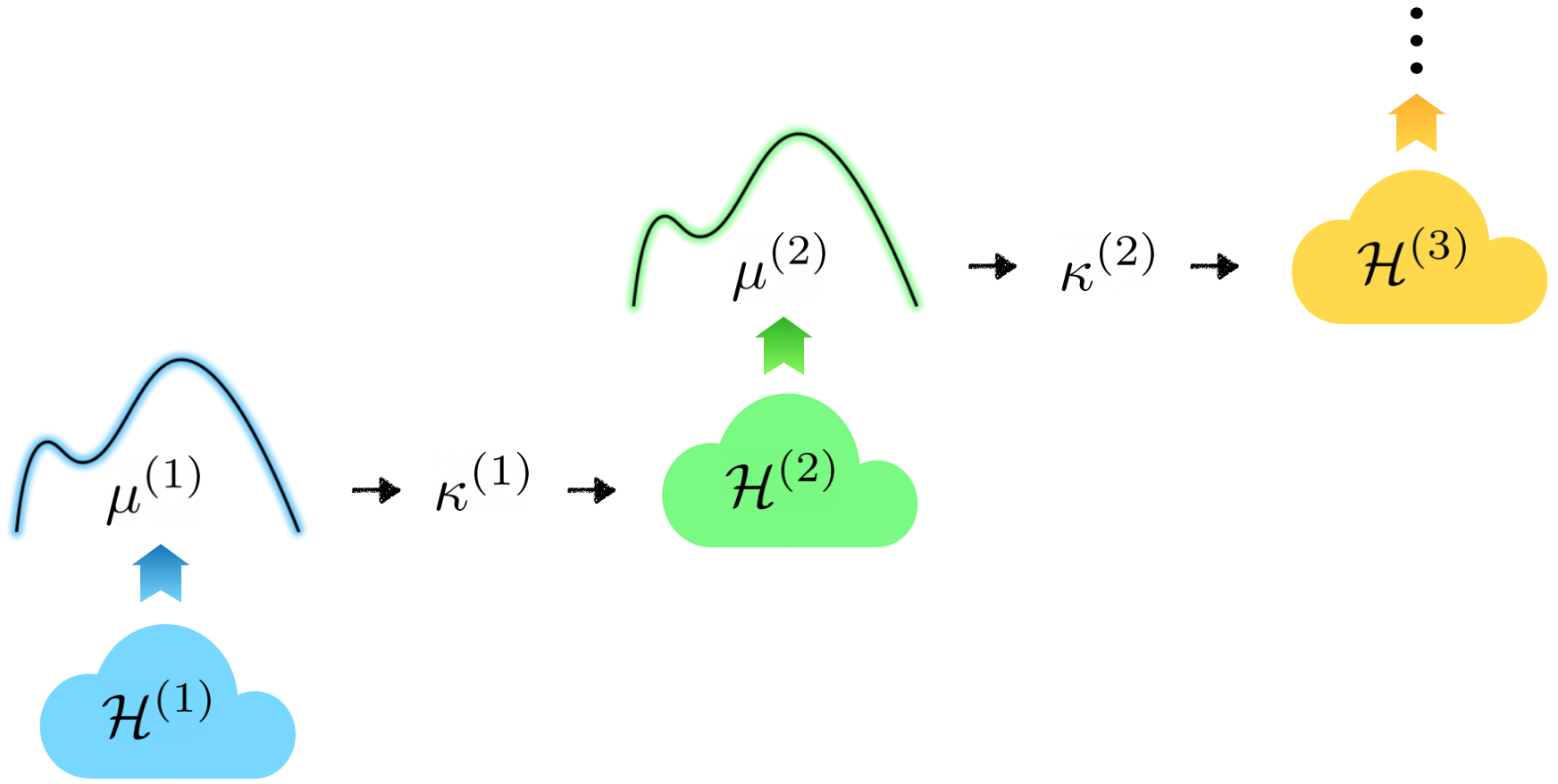}
    \caption{\textbf{Illustration of an NHL}, as defined in Definition~\ref{def:nhl}. Each $\Hilbl{l}$ is an RKHS; each $\mu^{(l)}$ is a probability measure on $\Hilbl{l}$; each kernel function $\kappa^{(l)}$ is defined by $\mu^{(l)}$ through \eqref{eq:kappa_mu} and, in turn, defines $\Hilb^{(l+1)}$ as its RKHS.}
    \label{fig:nhl}
\end{figure}
\subsection{Distributional View of Shallow NN}
\label{sec:shallow_mf}
When $L=2$, the model defined by \eqref{eq:f_m} reduces to the following expression of a shallow NN:
\begin{equation*}
    f(\xb) = \frac{1}{m} \sum_{i=1}^m a_i \sigbig{\zb_i^{\intercal} \cdot \xb}~,
\end{equation*}
which can be equivalently represented as $f(\xb) = \int a \sigbig{\zb^{\intercal} \cdot \xb} \mu_{m}(da, d\zb)$, if we define $\mu_{m}(da, d\zb) \coloneqq \frac{1}{m} \sum_{i=1}^m \delta_{a_i}(da) \delta_{\zb_i}(d\zb)$ as a probability measure on $\Rbb \times \Rbb^d$ that is the average of $m$ Dirac measures. More generally, we may consider functions that can be represented in a distributional form as
\begin{equation}
\label{eq:shallow_measure}
    f(\xb) = \int a \sigbig{\zb^{\intercal} \cdot \xb} \mu(da, d\zb)~,
\end{equation}
for some $\mu \in \Pcal(\Rbb \times \Rbb^d)$. In particular, the \emph{Barron norm} \citep{ma2022barron} of a function $f$ can be defined as
\begin{equation*}
    \| f \|_{\barron} \coloneqq \inf_{\mu} \int |a| \| \zb \| \mu(da, d\zb)~,
\end{equation*}
with the infimum taken over all $\mu \in \Pcal(\Rbb \times \Rbb^d)$ such that \eqref{eq:shallow_measure} holds. One can further define the \emph{Barron space} to contain all functions with a bounded Barron norm, which is equivalent to the $\mathcal{F}_{1}$ space considered by \citet{bach2017breaking} if $\sigma$ is $1$-homogeneous. 

Favorable theoretical properties of the Barron space have been derived in prior literature, including approximation guarantees and Rademacher complexity bounds \citep{barron1993universal, bach2017breaking, ma2022barron}. Meanwhile, the training dynamics of a two-layer NN can be understood through the Wasserstein gradient flow of the underlying probability measure $\mu$ \citep{mei2018mean, chizat2018global, rotskoff2018parameters, sirignano2020mean_lln}. Hence, through the distributional perspective, prior works have developed a promising picture of shallow NNs by viewing it as learning in the Barron space. A central motivation of this work is to extend this theory to NNs with more than two layers.

\subsection{Reproducing Kernel Hilbert Space (RKHS)}
A \emph{Hilbert space} is a vector space equipped with an inner product, $\langle \cdot, \cdot \rangle$, such that the norm defined through $\| \cdot \| \coloneqq \langle \cdot, \cdot \rangle$ makes it a complete metric space. Of particular interest to learning theory is a type of Hilbert spaces consisting of functions on the input domain, which we define below.

Let $\kappa: \domX \times \domX \to \Rbb$ be a symmetric and positive semi-definite function, which we call a \emph{kernel function}. It is associated with a particular Hilbert space on $\domX$, whose definition, existence and uniqueness are given by the following foundational result, often called the Moore-Aronszajn theorem \citep{aronszajn1950theory, cucker2002mathematical}:
\begin{lemma}[Moore-Aronszajn]
\label{lem:aron}
    There exists a unique Hilbert space, $\Hilb$, consisting of functions on $\domX$ and equipped with the inner product $\langle \cdot , \cdot \rangle_{\Hilb}$, which satisfies the following properties:
    \begin{enumerate}
        \item $\forall \xb \in \domX$, $\kappa(\xb, \cdot) \in \Hilb$;
        \item $\forall f \in \Hilb$, $\forall \xb \in \domX$, $\langle f, \kappa(\xb, \cdot) \rangle_{\Hilb} = f(\xb)$;
        \item the span of the set $\{ \kappa(\xb, \cdot) \}_{\xb \in \domX}$ is dense in $\Hilb$.
    \end{enumerate}
\end{lemma}
\noindent The Hilbert space that satisfies these properties is called the \emph{Reproducing Kernel Hilbert Space} (RKHS) associated with (a.k.a. reproducing) the kernel function $\kappa$. 
The RKHS plays an important role in classical learning theory as well as in mathematics and physics, and we refer the readers to \citet{cucker2002mathematical, mohri2018foundations} for further background. In Section~\ref{sec:related}, we will review the prior efforts in understanding NNs through kernels and RKHSs.

\section{Neural Hilbert Ladder (NHL)}
\label{sec:nhl}
We begin by introducing a way to create an RKHS from a \emph{distribution of functions} on $\domX$. 
If $\sigma: \Rbb \to \Rbb$ is the activation function of interest and $\mu \in \Pcal(\Conti)$, we define $\kappa_{\mu}: \domX \times \domX \to \Rbb$ by
\begin{equation}
\label{eq:kappa_mu}
    \kappa_{\mu}(\xb, \xb') \coloneqq \int \sigbig{h(\xb)} \sigbig{h(\xb')} \mu(dh)~,
\end{equation}
which is symmetric and positive semi-definite. Hence, by Lemma~\ref{lem:aron}, there is a unique RKHS on $\domX$ associated with the kernel function
$\Gfun_{\mu}$, which we denote by $\Hilb_{\mu}$. 

Leveraging this observation, we will present a recipe to construct a hierarchy of RKHSs iteratively. At the first level, we define $\Hilb^{(1)} \coloneqq \{ \xb \mapsto \zb^{\intercal} \cdot \xb : \zb \in \Rbb^d \}$ to be the space of linear functions on $\Rbb^d$. Through the canonical isomorphism with $\Rbb^d$, $\Hilb^{(1)}$ inherits an inner product from the Euclidean inner product on $\Rbb^d$, which makes $\Hilb^{(1)}$ the RKHS associated with the kernel function $\Gfun^{(0)}(\xb, \xb') := \xb^{\intercal} \cdot \xb'$. 
Then, for $L \geq 2$, we define an \emph{$L$-level neural Hilbert ladder} (NHL) as follows:
\begin{definition}
\label{def:nhl}
Suppose each of $\Hilb^{(2)}$, ..., $\Hilb^{(L)}$ is an RKHS on $\domX$, 
and $\forall l \in [L-1]$, there exists $\mu^{(l)} \in \Pcal(\Hilb^{(l)})$ such that 
$\Hilb^{(l+1)} = \Hilb_{\mu^{(l)}}$, which is the RKHS associated with $\kappa^{(l)} \coloneqq \kappa_{\mu^{(l)}}$.
Then, we say that $(\Hilb^{(l)})_{l \in [L]}$ is an \emph{$L$-level NHL} induced by the sequence of probability measures,
$(\mu^{(l)})_{l \in [L-1]}$. 
In addition, we say that a function $f$ on $\domX$ belongs to the NHL
if $f \in \Hilb^{(L)}$.
\end{definition}
\noindent Put differently, to define an NHL, at each level $l$ we choose a probability measure supported on $\Hilb^{(l)}$ (which can be viewed as the law of a random field on $\domX$) to generate $\kappa^{(l)}$ via \eqref{eq:kappa_mu}, which then determines $\Hilb^{(l+1)}$. Thus, an NHL is a ladder of RKHSs constructed by interleaving them with random fields and kernel functions, as illustrated in Figure~\ref{fig:nhl}.

This concept allows us to define complexity measures of functions as well as function spaces quantitatively, as we introduce next.

\subsection{Complexity Measures and Function Spaces}
If $\Hilb$, $\Hilb'$ are two RKHSs on $\domX$, then for $p > 0$ or $p = \infty$, we can define
\begin{equation*}
    \mathscr{D}_p \left (\Hilb, \Hilb' \right ) \coloneqq \inf_{\mu \in \Pcal(\Hilb),~ \Hilb_{\mu} = \Hilb'} \| \mu \|_{\Hilb, p}~,
\end{equation*}
 where $\| \mu \|_{\Hilb, p} \coloneqq (\int \| h \|_{\Hilb}^p \mu(dh))^{1/p}$ for $p > 0$ and $\| \mu \|_{\Hilb, \infty}$ denotes the essential supremum of the function $\| \cdot \|_{\Hilb}$ on $\Hilb$ with respect to $\mu$.  Note that the infimum is taken over all $\mu \in \Pcal(\Hilb)$ such that $\kappa_{\mu}$ equals the kernel function of $\Hilb'$.
Then, given an RKHS $\Hilb$ on $\domX$, we define
\begin{equation}
\label{eq:DLp}
    \DompLp{L}{p}(\Hilb) \coloneqq \inf_{
    \Hilb^{(2)},~ ...~,~ \Hilb^{(L-1)}} \prod_{l=1}^{L-1}  \mathscr{D}_p \big (\Hilb^{(l)}, \Hilb^{(l+1)} \big )~,
\end{equation}
 with the infimum taken over all
choices of $\Hilb^{(2)}$, ..., $\Hilb^{(L-1)}$ as RKHSs.
 Heuristically speaking, it quantifies the least \emph{price} (whose meaning regarding approximation and generalization will be discussed later) to be paid for arriving at $\Hilb$ as the $L$th-level of an NHL---note that one is free to choose the RKHSs and the probability measures in the middle of the ladder as long as Definition~\ref{def:nhl} is preserved.
Then, we define the \emph{$(L, p)$-NHL complexity} of a function $f$ as:
\begin{equation}
\label{eq:ladder_norm}
    \CompLp{L}{p}(f) \coloneqq \inf_{\Hilb} \Big ( \| f \|_{\Hilb} \cdot \DompLp{L}{p} \big ( \Hilb \big ) \Big )~, 
\end{equation}
with the infimum taken over all RKHS $\Hilb$.
Finally, we define the \emph{$(L, p)$-NHL space}, $\Ladder{L}_p$, to contain all functions with a finite $(L, p)$-NHL complexity: 
\begin{equation}
\label{eq:inf_union}
    \Ladder{L}_p \coloneqq \{ f: \CompLp{L}{p}(f) < \infty \} = \bigcup_{\DompLp{L}{p}(\Hilb) < \infty} \Hilb~.
\end{equation}
Note that unlike in the kernel theories of NNs (see Section~\ref{sec:related}), the function space $\Ladder{L}_{p}$ is not a \emph{single} RKHS but an infinite union of them. 

Some properties of $\Ladder{L}_{p}$ and $\CompLp{L}{p}$ are in order:
\begin{theorem}
    For $L \geq 2$, it holds that:
\begin{enumerate}[(a)]
    \item  For $p \in (0, \infty]$,  $\ladder^{(L)}_p$ is a vector space and $\CompLp{L}{p}$ is non-decreasing in $p$. \label{prop:vs}
    
    \item If $p \in [2, \infty]$ and $\sigma$ is $J_{\sigma}$-Lipschitz with $\sigma(0)=0$ (e.g. $J_{\sigma} = 1$ and $\sigma$ is ReLU or $\tanh$), then $\forall f \in \Ladder{L}_p$, it holds that $|f(\xb)| \leq (J_{\sigma})^{L-1} \CompLp{L}{p}(f) \| \xb \|$, $\forall \xb \in \domX$, and $f$
    satisfies the Lipschitz condition on $\domX$ with constant $(J_{\sigma})^{L-1} \CompLp{L}{p}(f)$. Hence, $\Ladder{L}_p \subseteq \Conti$.
    \label{prop:infty_norm}
    
    \item If $\sigma$ is $1$-homogeneous, then $\CompLp{L}{p}$ is equal for all $p \in [2, \infty]$. Moreover, for each such $p$, $\Ladder{L}_p$ is a quasi-Banach space with $\CompLp{L}{p}$ as the quasi-norm 
    (see Definition~\ref{def:quasib}).
    \label{prop:p_equiv}
\end{enumerate}
\label{prop:nhl_prop}
\end{theorem}
\noindent These results are proved in Appendix~\ref{app:pf_sec_nhl}. Specially, when $\sigma$ is $1$-homogeneous (e.g. ReLU), Theorem~\ref{prop:nhl_prop}(\ref{prop:p_equiv}) allows us to define $\CompL{L} \coloneqq \CompLp{L}{p}$ and $\Ladder{L} \coloneqq \Ladder{L}_p$ for any $p \in [2, \infty]$, which we will call the \emph{$L$-level NHL complexity} and the \emph{$L$-level NHL space}, respectively. 
 
\begin{remark}
    If $\sigma$ is Lipschitz and $\sigma(0) = 0$, Theorem~\ref{prop:nhl_prop}(\ref{prop:infty_norm}) implies that every $f \in \Ladder{L}$ needs to satisfy $f(0) = 0$. This corresponds to the omission of the bias terms in the NN and can be changed by either accounting for the bias terms in the definitions above (see Appendix~\ref{app:bias}) or redefining the input space to have an extra constant dimension via a map $\xb \mapsto (\xb, 1) \in \Rbb^{d + 1}$.
\end{remark}

\subsubsection{Example: $L=2$}
When $\sigma$ is $1$-homogeneous and $L=2$, we see from \eqref{eq:ladder_norm} and Lemma~\ref{lem:Htilde} in Appendix~\ref{app:pf_prop_p_equiv} that
\begin{equation*}
\begin{split}
    \CompL{2}(f) =&~ \inf_{\mu^{(1)} \in \Pcal(\hat{\Hilb}^{(1)})}~ \| f \|_{\Hilb^{(2)}}~, 
\end{split}
\end{equation*}
where the infimum is taken over all probability measures $\mu^{(1)}$ supported within the unit sphere of $\Hilb^{(1)}$. 
Meanwhile, it is known \citep{wojtowytsch2022representation} that the Barron norm of a function $f$ on $\domX$ can be rewritten as
\begin{equation*}
    \| f \|_{\barron} = \inf_{\xi, \rho}~ \left ( \int |\xi(\zb)|^2 \rho(d \zb) \right )^{1/2}~,
\end{equation*}
where the infimum is taken over all $\rho \in \Pcal(\mathbb{S}^{d-1})$ and all measurable functions $\xi: \mathbb{S}^{d-1} \to \Rbb$ such that
\begin{equation}
    \label{eq:shallow_expe}
    f(\xb) = \int \xi(\zb) \sigbig{\zb^{\intercal} \cdot \xb} \rho(d \zb)~.
\end{equation}
Noticing the analogous roles played by $\mu^{(1)}$ and $\rho$, we see an equivalence between $\CompL{2}$ and the Barron norm in light of 
Lemma~\ref{lem:equiv} below:
\begin{proposition}
\label{prop:C2_barron}
    If $\sigma$ is $1$-homogeneous, then $\| f \|_{\barron} = \CompL{2}(f)$.
\end{proposition}
\begin{lemma}
\label{lem:equiv} Let $\sigma$ be Lipschitz and $\mu \in \Pcal(\Conti)$ with $\int \| h \|_{\infty}^2 \mu(dh) < \infty$. Then, a function $f$ on $\domX$ belongs to $\Hilb_{\mu}$ if and only if $\exists \xi \in L_{2}(\Conti, \mu)$ such that
\begin{equation}
\label{eq:equiv}
    f(\xb) = \int \xi(h) \sigbig{h(\xb)} \mu(dh)~,~ \forall \xb \in \domX~.
\end{equation}
Moreover, 
$\| f \|_{\Hilbmu{\mu}} = \inf_\xi \| \xi \|_{L^2(\Conti, \mu)}$,
with the infimum over all $\xi \in L^{2}(\Conti, \mu)$ satisfying \eqref{eq:equiv}.
\end{lemma}
\noindent Lemma~\ref{lem:equiv} generalizes prior insights on the duality between RKHS and integral transforms \citep{rahimi2008uniform, bach2017breaking, bach2017equivalence}. While those results focus on cases where the basis functions are indexed by a compact set, here the basis functions $\{ \sigma(h(\cdot)) \}_{h \in \Conti}$ are indexed by an infinite-dimensional function space. In Appendix~\ref{app:pf_lem_equiv}, we prove this lemma based on a general result by \cite{saitoh1997integral}.

Proposition~\ref{prop:C2_barron} therefore implies that $\Ladder{2}$ is identical to the Barron space.
In fact, when $L=2$, \eqref{eq:inf_union} reduces to $\Ladder{2} = \bigcup_{\rho \in \Pcal(\Rbb^d)} \Hilb_{\rho}$, where we define $\Hilb_{\rho}$ as the RKHS associated with the kernel function $\kappa_{\rho}(\xb, \xb') \coloneqq \int \sigma(\zb^{\intercal} \cdot \xb) \sigma(\zb^{\intercal} \cdot \xb') \rho(d \zb)$. This agrees with the decomposition of the Barron space as a union of RKHSs \citep{ma2018priori}.

Thus, $\Ladder{L}$ can be seen as generalizing the Barron space to the cases where $L > 2$.

\subsection{Alternative Form via Coupled Random Fields}
\label{sec:entang}
For each $l \in [L-1]$, $\mu^{(l)}$  can be interpreted as the law of a random field, $\Hb^{(l)}$, whose sample paths---as functions on $\domX$---belong to $\Hilb^{(l)}$. In fact, one could define the random fields on a common probability space to derive an alternative formulation of the NHL that will be relevant later:

\begin{proposition}
\label{prop:coupled}
Given an $L$-level NHL as in Definition~\ref{def:nhl}, there exist $L-1$ random fields on $\domX$, $\Hb^{(1)}$, ..., $\Hb^{(L-1)}$, as well as $L-1$ scalar random variables, $\Xib^{(1)}$, ..., $\Xib^{(L-2)}$, and $\Ab$, which are defined on a common probability space and satisfy the following properties:
\begin{itemize}
\item $\Hb^{(1)}$, ..., $\Hb^{(L-1)}$ are mutually independent, and $\forall l \in [L-1], \mu^{(l)} = \text{Law}(\Hb^{(l)})$;

    \item $\forall l \in [L-2]$, $\Xib^{(l)}$ is measurable with respect to the sigma-algebra generated by $\Hb^{(l)}$ and $\Hb^{(l+1)}$. Moreover,
    \begin{equation}
\label{eq:H_entang}
        \Hb^{(l+1)}(\xb) = \EEEb{\Xib^{(l)} \sigbig{\Hb^{(l)}(\xb)} \big | \Hb^{(l+1)}}~,
    \end{equation}
    where $\EEElr{ \cdot | \cdot }$ denotes the conditional expectation, and $\| \Hb^{(l+1)} \|_{\Hilb^{(l+1)}}^2 = \EEEb{(\Xib^{(l)})^2 \big | \Hb^{(l+1)}}$.
    \item  $\Ab$ is measurable with respect to $\Hb^{(L-1)}$. Moreover,
\begin{equation}
\label{eq:f_entang}
    f(\xb) = \EEEb{\Ab \sigbig{\Hb^{(L-1)}(\xb)}}~,
\end{equation}
and $\| f \|_{\Hilb^{(L)}}^2 = \EEEb{\Ab^2}$.
\end{itemize}
\end{proposition}
\noindent The proof of Proposition~\ref{prop:coupled} is given in Appendix~\ref{app:pf_entang_equiv} and also based on Lemma~\ref{lem:equiv}.

\section{Realization and Approximation by Neural Networks}
\label{sec:real_approx}

For $p \in [2, \infty]$, we shall define quantities $M^{(1)}_{m, p}, ..., M^{(L)}_{m, p} \geq 0$ associated with the NN defined in Section~\ref{sec:nn} by
\begin{equation*}
    \begin{split}
        M^{(1)}_{m, p} \coloneqq \bigg ( \frac{1}{m} \summ{i}{m} \| \zb_i \|_{2}^p \bigg )^{\frac{1}{p}}~,~
        M^{(L)}_{m, p} \coloneqq \bigg ( \frac{1}{m} \summ{i}{m} a_i^2 \bigg )^{\frac{1}{2}}~,
    \end{split}
\end{equation*}
\begin{equation*}
    M^{(l+1)}_{m, p} \coloneqq \bigg ( \frac{1}{m} \summ{i}{m} \bigg ( \frac{1}{m} \summ{j}{m} |W^{(l)}_{i, j}|^2 \bigg )^{\frac{p}{2}} \bigg )^{\frac{1}{p}}~,~ \forall l \in [L-2]~,
\end{equation*}
with the usual convention of $(\summ{i}{m} |u_i|^\infty)^{1/\infty} = \sup_{i \in [m]} |u_i|$.
Modulo the $1 / m$ scaling factor, they coincide with the per-layer Frobenius norms of the weight parameters when $p = 2$, and more generally, they appear in the \emph{group norm} of finite-width NNs defined in \citet{neyshabur2015norm}. 
Meanwhile, with the $1 / m$ scaling factor, these quantities admit width-independent upper bounds if each parameter is sampled i.i.d.
\subsection{NN as NHL}
First, we show that every $L$-layer NN represents a function in $\ladder^{(L)}_p$, verifying property \eqref{item:1}.
\begin{theorem}
\label{prop:in_ladder}
$\forall p \in [2, \infty]$, $f_{m} \in \Ladder{L}_p$ with $\CompLp{L}{p}(f_{m}) \leq \prod_{l=1}^L M^{(l)}_{m, p}$. In particular,
$f_{m}$ belongs to the NHL of $(\Hilb^{(l)}_{m})_{l \in [L]}$, where we define $\Hilb^{(1)}_{m} \coloneqq \Hilb^{(1)}$, and 
$\forall l \in [L-1]$, $\Hilb^{(l+1)}_{m} \coloneqq \Hilb_{\mu^{(l)}_{m}}$ with
$\mu_{m}^{(l)} \coloneqq \frac{1}{m} \summ{i}{m} \delta_{h_i^{(l)}} \in \Pcal(\Conti)$ being the empirical measure (on functional space) of the pre-activation functions of the neurons in the $l$th hidden layer.
\end{theorem}
\noindent The proof is given in Appendix~\ref{app:pf_prop_in_ladder} and leverages Lemma~\ref{lem:equiv}. Thus, a multi-layer NN can indeed be viewed as representing an NHL whose complexity is controlled by the layer-wise norms of the NN parameters. In particular, we see that the series of random fields can be constructed based on the pre-activation functions in the respective hidden layers.
\subsection{NHL can be Approximated by NN}
\label{sec:approx}
Conversely,  we can show that any function in $\Ladder{L}_\infty$
can be approximated relatively efficiently in squared error by an $L$-layer NN: 

\begin{theorem}
\label{thm:approx}
Suppose $\sigma$ is $J_{\sigma}$-Lipschitz with $\sigma(0) = 0$ and $\nu \in \Pcal(\domX)$ satisfies $\expenu{\xb \sim \nu}{\| \xb \|^2} = M_{\nu} < \infty$. Given any $f \in \Ladder{L}_\infty$, there exists a function $f_{m}$ represented by an $L$-layer NN with width $m$ such that 
$\expenus{\xb \sim \nu}{| f_{m}(\xb) - f(\xb) |^2} \leq {(L-1)^2} (J_{\sigma})^{2L-2} M_{\nu} (\CompLp{L}{\infty}(f))^2 / m$.
\end{theorem}

\noindent This result is proved in Appendix~\ref{app:thm_approx_pf}, where we use an inductive-in-$L$ argument to show that a randomized approximation strategy based on sampling each $\mu^{(l)}$ independently can achieve a low approximation error in expectation.

Theorem~\ref{thm:approx} implies that a function in $\Ladder{L}_\infty$ can be approximated within a squared error of $\epsilon$ by an $L$-layer NN using $O(L^5 / \epsilon^4)$  number of parameters in total. In comparison, functions in the \emph{neural tree space} defined by \citet{wojtowytsch2020banach} require 
above the order of  $(L/\epsilon)^L$  parameters, or in other words, the approximation error bound decays \emph{exponentially} slowly as the depth increases. The contrast highlights a property of multi-layer NNs---that all neurons in one hidden layer share the same pre-synaptic neurons of the preceding layer---which is captured by the NHL by not by the neural tree space representation, whose branching structure incurs the exponential dependence on the depth.

 We note that for general activation functions, the current proof technique for Theorem~\ref{thm:approx} only succeeds when $p = \infty$.  Nonetheless, when $\sigma$ is $1$-homogeneous, Theorems~\ref{prop:in_ladder} and \ref{thm:approx} are both applicable to the space $\Ladder{L}$ and establish a two-way correspondence between $L$-layer NNs and functions in this space with the approximation cost governed by $\CompL{L}$, thus verifying both criteria \eqref{item:1} and \eqref{item:2}.

\section{Generalization Theory}
\label{sec:gen}
\label{sec:supervised} 
To study the learning aspect of the function space defined above, we consider the task of fitting a target function on the input domain, $f^{*}: \domX \to \mathcal{Y}$ with $\mathcal{Y}$ being a bounded subset of $\Rbb$, using $\Ladder{L}$ as the hypothesis class (and we assume in this section that $\sigma$ is $1$-homogeneous). 
Concretely, we desire for a function $f$ from $\Ladder{L}$ that minimizes the \emph{population risk}, defined as $\Risk(f) \coloneqq \expenu{\xb \sim \nu}{l(f(\xb), f^{*}(\xb))}$, where $\nu \in \Pcal(\domX)$ is an underlying data distribution on $\domX$ and $l$ is a loss function on $\Rbb \times \mathcal{Y}$ (e.g., in least squares regression, $l(\hat{y}, y) = \frac{1}{2}(\hat{y} - y)^2$). In supervised learning, instead of having access to $\nu$ directly, we are typically given a training set of size $n$,  $S = \{\xbtrk{1}, ..., \xbtrk{n} \} \subseteq \domX$, sampled i.i.d. from $\nu$, and we write $\nu_{n} = \frac{1}{n} \summ{k}{n} \delta_{\xbtrk{k}} \in \Pcal(\domX)$. A standard strategy is to find a function from the given function space that achieves a low \emph{empirical risk}, defined as $\Risk_{n}(f) \coloneqq \expenu{\xb}{l(f(\xb), f^{*}(\xb))}$, where we write $\mathcal{E}_{\xb}$ for $\mathcal{E}_{\xb \sim \nu_{n}}$ for simplicity. Then, the question of generalization concerns how fast the discrepancy between $\Risk$ and $\Risk_{n}$ decays as $n$ increases. 

Classical learning theory suggests that one can prove uniform upper bounds on the discrepancy through estimating the \emph{Rademacher complexity} of $\Ladder{L}$. Recall that the \emph{empirical Rademacher complexity} of a function space $\mathcal{F}$ with respect to the set $S$ is defined by $\widehat{\Rad}_{S}(\mathcal{F}) \coloneqq \EE_{\taub} \left [ \frac{1}{n} \sup_{f \in \mathcal{F}} \summ{k}{n} \tau_{k} f(\xb_{k}) \right ]$, where $\taub = [\tau_{1}, ..., \tau_{n}]$ is a vector of i.i.d. Rademacher random variables. Note that while the Rademacher complexity of RKHS is known \citep{mendelson2003performance, bartlett2005local}, $\ladder^{(L)}$ is not \emph{one} RKHS but a union of infinitely many of them (and hence richer), which calls for a different analysis.

Our main result in this section is the following:
\begin{theorem}
\label{prop:rad}
If $\domX \subseteq \Ball(\Rbb^d)$ and $\sigma$ is $1$-homogeneous, then $\forall M > 0$, for any training set $S$ of size $n$, $\widehat{\Rad}_{S}(\Ball(\ladder^{(L)}; M)) \leq M (\sqrt{2 L \log (2 J_{\sigma})} + 1) / \sqrt{n}$, where $J_{\sigma} \coloneqq \max\{|\sigma(1)|, |\sigma(-1)|\}$.
\end{theorem}
\noindent The proof is given in Appendix~\ref{app:pf_rad}. In particular,
we carry out an inductive-in-$L$ argument inspired by \citet{neyshabur2015norm} combined with a strategy proposed by \citet{golowich2018size} for reducing the dependency of the bound on $L$ from exponential to $O(\sqrt{L})$. Note that the same $O(1/\sqrt{n})$ rate is obtained as in the Rademacher complexity bounds for RKHSs.

Combining this result with Theorem~\ref{prop:nhl_prop}(\ref{prop:infty_norm}) and classical generalization bounds based on Rademacher complexity (e.g. \citealp{mohri2018foundations}), we derive the following generalization guarantee for learning in the space of NHLs under the ReLU activation, thus verifying \eqref{item:3}:
\begin{corollary}
    Suppose that $\domX \subseteq \Ball(\Rbb^d)$, $\mathcal{Y} \subseteq [-1, 1]$, $\sigma$ is ReLU, and $\forall y \in \mathcal{Y}$, the function $\hat{y} \mapsto l(\hat{y}, y)$ is $J_{l}$-Lipschitz on $[-1, 1]$. Then, $\forall \delta > 0$, with probability at least $1-\delta$ over the i.i.d. sampling of a sample $S$ of size $n$ in $\domX$, it holds for all functions $f$ with $\CompL{L}(f) \leq 1$ that
    $\Risk(f) \leq \Risk_{n}(f) + 2 J_{l} (\sqrt{2 L \log 2} + 1) / \sqrt{n} + 3 \sqrt{\log(2 / \delta)} / \sqrt{2n}$.
\end{corollary}

\section{Effect of Depth}
\label{sec:depth}
A rich body of work has studied the benefit of depth in NNs in terms of whether certain functions can be approximated much more efficiently by deeper NNs than shallow ones, where the efficiency is usually measured by the width required. Meanwhile, it remains interesting whether the effect of depth can be properly characterized for \emph{width-unlimited} NNs in terms of the respective function spaces. Under the NTK theory, for example, the function spaces corresponding to NNs with different depths are actually equivalent \citep{bietti2021deep, chen2021deep}. By contrast, below we demonstrate an example where the NHL space is strictly larger as we increase the depth.

We consider the case where $\sigma$ is the ReLU function. On one hand, since a single hidden layer is capable of expressing the identity function using a pair of ReLU neurons with opposite weights, it is straightforward to show that $\Ladder{L}$ does not shrink as $L$ increases. In fact, we can obtain a slightly sharper bound on $\CompL{L+1}$ relative to $\CompL{L}$ by further exploiting the properties of ReLU:
\begin{lemma}
\label{lem:relu_grows}
    If $\sigma$ is ReLU, then $\forall L \geq 2$, $\Ladder{L} \subseteq \Ladder{L+1}$ and  $\CompL{L+1}(f) \leq \sqrt{2} \CompL{L}(f)$,  $\forall f \in \Ladder{L}$.
\end{lemma}
\noindent The proof is given in Section~\ref{app:pf_lem_relu_grows}. On the other hand, \cite{ongie2020multivariate} proved that the function $f(\xb) = \max \{ 1 - \| \xb \|_{1}, 0 \}$ on $\Rbb^d$ cannot be represented by any infinite-width shallow ReLU NN with a finite Barron norm, and hence it does not belong to $\Ladder{2}$. Meanwhile, it can be represented by a finite-width $3$-layer NN, and thus, as a consequence of Theorem~\ref{thm:approx}, it belongs to $\Ladder{3}$. Therefore, we establish a strict inclusion of $\Ladder{2}$ within $\Ladder{3}$:
\begin{proposition}
\label{prop:relu_sep}
    When $\sigma$ is ReLU, $\Ladder{2} \subsetneq \Ladder{3}$.
\end{proposition}

Prior results on depth separation (reviewed in Section~\ref{sec:related}) often involve functions whose approximation cost---in terms of the number of parameters---suffers from the curse of dimensionality only when the depth is small. In our framework, with the cost measured by the NHL complexity,
we can show an analogous result via an example studied by \citet{wojtowytsch2024optimal}, as we briefly recount below.
Fixing $\epsilon_{0} \in (0, 1)$, we define $f_{\epsilon_{0}}: \Rbb^d \to \Rbb$ by
\begin{equation*}
    f_{\epsilon_{0}}(\xb) = \begin{cases}
        1~,~ \quad & \text{if } \| \xb \| \leq \epsilon_{0}, \\
        1 - \frac{\| \xb \| - \epsilon_{0}}{1 - \epsilon_{0}}~,~ \quad & \text{if } \epsilon_{0} < \| \xb \| \leq 1, \\
        0~,~ \quad & \text{if } \| \xb \| > 1,
    \end{cases}
\end{equation*}
When the activation function is ReLU and the bias term is considered, it is shown that $f_{\epsilon_{0}}$ can be written as the composition of two functions whose Barron norms depend sublinearly on the input dimension $d$, thus implying a similar bound on $\CompL{3}(f_{\epsilon_{0}})$. Meanwhile, the Barron norm of $f_{\epsilon_{0}}$ itself, $\CompL{2}(f_{\epsilon_{0}})$, is exponentially large in $d$ \citep{wojtowytsch2024optimal}. Hence, we arrive at the following type of depth separation result for NHL spaces:
\begin{proposition}
    When $\sigma$ is ReLU and the bias term is included, there exists a function $f$ on $\Rbb^d$ such that $\CompL{3}(f)$ depends sublinearly on $d$ while $\CompL{2}(f)$ is exponentially large in $d$.
\end{proposition}

\section{Training Dynamics under Gradient Flow (GF)}
\label{sec:training}
\label{sec:gf}
Given the correspondence between NNs and NHLs shown in Section~\ref{sec:approx}, we may regard the training of NNs as a instantiating a particular learning rule for solving empirical risk minimization within $\Ladder{L}$, whose mechanism will be further analyzed in this section. Typically in practice, we initialize an NN by sampling its parameters randomly and train it by performing variants of GD on the parameters with respect to the empirical risk. We assume below that
\begin{assumption}
    \label{ass:sigma}
    $\sigma$ is differentiable and its derivative $\sigma'$ is Lipschitz and bounded. 
\end{assumption} 
\begin{assumption}
    \label{ass:loss}
    $l$ is differentiable with respect to the first argument and the derivative 
    $\partial_{\hat{y}} l(\hat{y}, y)$ is Lipschitz in $\hat{y}$ for $y \in \mathcal{Y}$.
\end{assumption} 
\begin{assumption}
\label{ass:init}
    At $t=0$, each $W_{i, j, 0}$, $a_{i, 0}$ and  $\zb_{i, 0}$ is sampled independently from $\rhoW$, $\rhoa \in \Pcal(\Rbb)$ and $\rhoz \in \Pcal(\Rbb^d)$, respectively. Moreover, $\rhoW$ and $\rhoa$ have zero mean, $\rhoW$ has a finite fourth-moment, $\rhoz$ has a finite covariance, and $\rhoa$ is bounded. 
\end{assumption}
\noindent
Note that Assumption~\ref{ass:sigma} 
is standard in prior works on the mean-field theory of shallow NNs (e.g. \citealp{chizat2018global})
though it is not satisfied when $\sigma$ is ReLU, and Assumption~\ref{ass:loss} is satisfied if $l$ is the squared loss or the logistic loss, for instance. 

For simplicity, we consider GD dynamics in the continuous-time limit---also called the \emph{gradient flow} (GF)---where the parameters evolve over time $t$ (added as a subscript to the quantities in \eqref{eq:f_m} and \eqref{eq:h_i}) according to a system of ordinary differential equations: 
\begin{align*}
    \ddt \zbjt{i}{t} =&~ - \expenuB{\xb}{\zetamt{t}(\xb) \qitl{i}{t}{1} (\xb) \sigpbig{\hitl{i}{t}{1} (\xb)}}, \\
    \ddt a_{i, t} =&~ - \expenuB{\xb}{\zetamt{t}(\xb) \sigbig{\hitl{i}{i}{L-1} (\xb)}}~, 
\end{align*}
and $\forall l \in [L-2]$,
\begin{equation*}
    \ddt \Wijtl{i}{j}{t}{l} = - \expenuB{\xb}{\zetamt{t}(\xb) \qitl{i}{t}{l+1}(\xb) \sigpbig{\hitl{i}{t}{l+1} (\xb)} \sigbig{\hitl{j}{t}{l}(\xb)}}~,\end{equation*}
where $\zetamt{t}(\xb) \coloneqq \partial_{\hat{y}} l(\hat{y}, f^{*}(\xb)) |_{\hat{y} = f_{m, t}(\xb)}$, $\qitl{i}{t}{L-1}(\xb) \coloneqq a_i \sigpbig{\hitl{i}{t}{L-1}(\xb)}$, and $\forall l \in [L-2]$,
\begin{equation*}
    \qitl{j}{t}{l}(\xb) \coloneqq \frac{1}{m} \bigg ( \summ{i}{m} \Wijtl{i}{j}{t}{l} \qitl{i}{t}{l+1}(\xb) \sigpbig{\hitl{i}{t}{l+1}(\xb)}  \bigg )~.
\end{equation*}
Note that GF causes not only the output function $f_{m, t}$ but also the pre-activation functions in the hidden layers---and hence $\mu_{m, t}^{(l)} \coloneqq \frac{1}{m} \summ{i}{m} \delta_{h_{i, t}^{(l)}}$---to evolve, thus leading to a movement of the overall NHL represented by the model.
While the dynamics of $(\mu_{m, t}^{(l)})_{l \in [L-1]}$ is
not closed but depending intricately on the weight matrices, we will show below that their effect can be subsumed by a mean-field description once we consider the infinite-width limit.

\subsection{Mean-Field NHL Dynamics}
\label{sec:mfnhl}
Several prior works have considered the infinite-width limits of multi-layer NNs in the mean-field scaling and derived equations that govern their GF dynamics \citep{araujo2019mean, nguyen2020rigorous, sirignano2022mean_deep}. Following their insights, our goal is to uncover the learning dynamics of the NHL in the mean-field limit. To do so, we first write down the mean-field limit in the form of coupled random fields introduced in Section~\ref{sec:entang}, as follows. Let $\Zb_{0}$ and $\Ab_{0}$ be a $d$-dimensional random vector with law $\rhoz$ and a scalar random variable with law $\rhoa$, respectively, distributed independently from each other. Then, we introduce time-dependent random fields $\Htl{t}{1}$, ..., $\Htl{t}{L-1}$ and $\Qtl{t}{1}$, ..., $\Qtl{t}{L-1}$
on $\domX$. At initial time, they are defined by 
\begin{align*}
    \Htl{0}{1}(\xb) = \Zb_{0}^{\intercal} \cdot \xb~, \qquad \Qtl{0}{L-1}(\xb) = \Ab_{0}~,
\end{align*}
and $\Htl{0}{l+1}(\xb) = \Qtl{0}{l}(\xb) = 0$, $\forall l \in [L-2]$. Their evolution in time is given by, 
$\forall l \in [L-1]$,
\begin{align}
    \Htl{t}{l}(\xb) =&~ \Htl{0}{l}(\xb) - \int_{0}^t \expenuB{\xb'}{\zeta_{s}(\xb') \Gfun^{(l-1)}_{t, s}(\xb, \xb') \Qtl{s}{l}(\xb') \sigpbig{\Htl{s}{l}(\xb')}} ds~, \label{eq:Htl} \\
    \Qtl{t}{l}(\xb) =&~ \Qtl{0}{l}(\xb) - \int_{0}^t \expenuB{\xb'}{\zeta_{s}(\xb') \gamma^{(l+1)}_{t, s}(\xb, \xb') \sigbig{\Htl{s}{l}(\xb')}} ds~, \label{eq:Qtl}
\end{align}
where we define $\Gfun_{t, s}^{(0)}(\xb, \xb') = \gamma^{(L)}_{t, s}(\xb, \xb') = 1$ and, $\forall l \in [L-1]$,
\begin{align*}
    \Gfun_{t, s}^{(l)}(\xb, \xb') \coloneqq&~ \EEElr{\sigbig{\Htl{t}{l}(\xb)} \sigbig{\Htl{s}{l}(\xb')}}~, \\
    \kappaker{t}{s}{l}{\xb}{\xb'} \coloneqq&~ \EEElr{\Qtl{t}{l}(\xb) \Qtl{s}{l}(\xb') \sigpbig{\Htl{t}{l}(\xb)} \sigpbig{\Htl{s}{l}(\xb')}}~.
\end{align*}
Lastly, we set $f_{t}(\xb) = \EEEb{\Qtl{t}{L-1}(\xb) \sig{\Hb_{t}^{(L-1)}(\xb)}}$ (though, from the definition above, it can be seen that $\Qtl{t}{L-1}(\xb)$ is independent of $\xb$) and $\zeta_{t}(\xb) = \partial_{\hat{y}} l(\hat{y}, f^{*}(\xb)) |_{\hat{y} = f_{t}(\xb)}$.

We can show that, by the law of large numbers, the dynamics of $f_{t}$ defined via the above is the infinite-width limit of the GF training of NNs considered in Section~\ref{sec:gf}.
\begin{theorem}
\label{prop:lln} 
 Suppose Assumptions~\ref{ass:sigma} -- \ref{ass:init} hold. Then $\forall t \geq 0$, as $m \to \infty$,
\begin{enumerate}
    \item $f_{m, t}(\xb) \xrightarrow{a.s.} f_{t}(\xb)$,  $\forall \xb \in \domX$;
    \item $\forall l \in [L-1]$, the probability measure $\mu_{m, t}^{(l)}$ converges weakly to $\mu^{(l)}_{t} \coloneqq \Law(\Htl{t}{l})$ in all finite distributions, that is, $\forall N \in \Nbb_+$, $\forall \xbg{1}$, ..., $\xbg{N} \in \domX$,
    \begin{align*}
        \sup_{g \in \text{Lip}(\Rbb^{N})} &~ \bigg | \int g(h(\xbg{1}), ..., h(\xbg{N})) \mu^{(l)}_{m, t}(dh) - \int g(h(\xbg{1}), ..., h(\xbg{N})) \mu^{(l)}_{t}(dh) \bigg | \xrightarrow{a.s.} 0~.
    \end{align*}
\end{enumerate}
\end{theorem}
\begin{remark}
\label{rmk:degen}
If $L \geq 4$, then for $2 \leq l \leq L-2$, the random field $\Htl{t}{l}$ defined through the above is actually deterministic for all $t \geq 0$, indicating a type of degeneracy in deep NNs under the mean-field scaling \citep{araujo2019mean, nguyen2020rigorous}. Randomness can be restored if we add a bias term to each layer that is randomly initialized (see Appendix~\ref{app:mf_bias}).
\end{remark}
\noindent Theorem~\ref{prop:lln} is proved in Appendix~\ref{app:pf_lln}, including in the more general case where the bias terms are added. The proof consists of two steps. First, we rewrite the limiting dynamics in a differential form using the formulation introduced in Section~\ref{sec:entang}.
Second, we prove the convergence of the dynamics as $m \to \infty$
via a propagation-of-chaos-type argument \citep{mckean1967propagation, braun1977vlasov}, which also appears in several prior works on the mean-field limit of multi-layer NNs \citep{araujo2019mean, pham2020iclr, sirignano2022mean_deep}. 
Note that our result here is meant not to improve upon these prior works in techniques, but rather to demonstrate that the training dynamics in the mean-field limit can be interpreted as a learning dynamics of the NHL.
Specifically, as a corollary of Lemma~\ref{lem:equiv}, we see that the limiting dynamics indeed defines an evolving NHL (hence named the \emph{mean-field NHL dynamics}), with each $\Hilb^{(l+1)}_{t}$ being the RKHS associated with the kernel function $\kappa^{(l)}_{t} \coloneqq \kappa^{(l)}_{t, t}$:
\begin{proposition}
\label{prop:mf_NHL}
    Suppose Assumptions~\ref{ass:sigma} -- \ref{ass:init} hold. $\forall t \geq 0$, $f_{t}$ belongs to the NHL of 
    $(\Hilb^{(l)}_{t})_{l \in [L]}$,
    where we define $\Hilb^{(1)}_{t} \coloneqq \Hilb^{(1)}$, and $\forall l \in [L-1]$, $\Hilb^{(l+1)}_{t} \coloneqq \Hilb_{\mu^{(l)}_{t}}$ as the RKHS associated with 
    $\kappa_{t}^{(l)}$. 
    Moreover, $\forall l \in [L-1]$, as $m \to \infty$,  $\Gfun_{m, t}^{(l)}$ converges to $\Gfun_{t}^{(l)}$ almost surely.
\end{proposition}

\begin{remark}
Unlike the mean-field limit of shallow NNs or a usual gradient descent dynamics, the dynamics of each $\Hb^{(l)}_{t}$ is notably no longer Markovian when $L > 2$. The memory effect emerges because we track the evolution of the neurons' pre-activation functions rather than that of their weight parameters. Similar phenomena are also seen in an alternative infinite-width limit studied by \citet{yang2021feature, bordelon2022selfconsistent}, 
\end{remark}

\noindent From the mean-field NHL dynamics, we derive that the output function satisfies
\begin{equation}
\label{eq:kgf}
    \ddt f_{t}(\xb) = \expenu{\xb'}{\zeta_{t}(\xb') \theta_{t}(\xb, \xb')}~,
\end{equation}
where we define
\begin{equation*}
    \theta_{t}(\xb, \xb') \coloneqq \summ{l}{L} \kappa^{(l-1)}_{t}(\xb, \xb') \gamma^{(l)}_{t}(\xb, \xb')~,
\end{equation*}
and we set $\kappa^{(0)}_{t}(\xb, \xb') \coloneqq \xb^{\intercal} \cdot \xb'$, $\gamma^{(L)}_{t}(\xb, \xb') \coloneqq 1$ and  $\gamma^{(l)}_{t} \coloneqq \gamma^{(l)}_{t, t}$, $\forall l \in [L-1]$. Hence, one can view $f_{t}$ as evolving according to a \emph{functional gradient flow}---also called the residual dynamics in the mean-field theory of shallow NNs \citep{rotskoff2018parameters}---under a \emph{time-varying} and \emph{data-dependent} kernel function $\theta_{t}$. It is analogous to the Neural Tangent Kernel (NTK) that governs the training dynamics of infinite-width NNs under a difference choice of scaling (\citet{jacot2018neural}; see the discussions in Section~\ref{sec:related}). However, a crucial difference is that the NTK remains \emph{fixed} during training and agnostic to the training data---and hence the name ``lazy training'' \citep{chizat2018global}---whereas the model here exhibits feature learning as $\theta_{t}$ evolves during training based on the data, as summarized in criterion \eqref{item:4}.

\subsubsection{Example $1$: $L=2$}
\label{sec:shallow_wgf}
When $L=2$, the mean-field NHL dynamics is Markovian and reduces to the following ODEs of the random variables $\Ab_{t}$ and $\Zb_{t}$:
\begin{align*}
    \ddt \Ab_{t} = &~ \expenuB{\xb}{\zeta_{t}(\xb) \sigbig{\Zb_{t}^{\intercal} \cdot \xb}}~, \\
    \ddt \Zb_{t} = &~ \Ab_{t} \expenuB{\xb}{\zeta_{t}(\xb) \sigpbig{\Zb_{t}^{\intercal} \cdot \xb} \xb}~.
\end{align*}
In this case, one can fully characterize the NHL dynamics by the evolution of the joint law of $\Ab_{t}$ and $\Zb_{t}$, which follows a Wasserstein gradient flow (WGF) in $\Pcal(\Rbb^{d+1})$, \emph{without} an underlying probability space for defining the random variables. This allows us to recover the mean-field theory of shallow NNs, which we review in Section~\ref{sec:related}. In particular, under the same assumptions, the global convergence guarantees of the WGF
\citep{nitanda2017stochastic, chizat2018global, mei2018mean, rotskoff2018parameters, wojtowytsch2020convergence} 
also apply to the mean-field NHL dynamics at $L=2$.

Note that when $L > 2$, the probability-space-free representation becomes less convenient. On one hand, the random fields $\Htl{t}{l}$ can no longer be reduced to finite-dimensional random variables when $l > 1$. On the other hand, the mean-field NHL dynamics \eqref{eq:Htl} becomes non-Markovian. For these reasons, inspired by the approaches taken by \citet{nguyen2020rigorous, fang2021modeling}, we choose to formulate the general mean-field dynamics through the random field representation.

\subsubsection{Example $2$: Deep Linear NN}
\label{sec:linear}
When $\sigma$ is the identity function, the model becomes a linear NN, whose output function can be written as $f_{t}(\xb) = \vb_{t}^{\intercal} \cdot \xb$. Moreover, for all $l \in \{2, ..., L\}$, $\Hilb^{(l)}_{t}$ always contains the same set of functions, namely, the linear functions on $\Rbb^d$, except that their norms in $\Hilb^{(l)}_{t}$, which are governed by the kernel function $\kappa^{(l-1)}_{t}$, can differ with $l$ and evolve over time. 

Below, we show the mean-field NHL dynamics reduces in this case to a finite-dimensional system. 
We consider the setting of fitting a linear target function $f^{*}(\xb) = (\vb^{*})^{\intercal} \cdot \xb$ with least squares regression, and we define $\Sigma \coloneqq \frac{1}{n} \summ{k}{n} \xb_{k} \cdot \xb_{k}^{\intercal}$ and $\zetab_{t} \coloneqq \Sigma \cdot (\vb_{t} - \vb^{*})$. We assume for simplicity that $\rhoa$ has zero mean and unit variance while $\rhoz$ has the zero vector as its mean and the identity matrix as its covariance.

Thanks to the linearity, each $\kappa^{(l)}_{t, s}(\xb, \xb')$ is bilinear in $\xb$ and $\xb'$ while $\kappaker{l}{t}{s}{\xb}{\xb'}$ does not depend on $\xb$ or $\xb'$. In other words, $\exists K^{(l)}_{t, s} \in \Rbb^{d \times d}$ and $c^{(l)}_{t, s} \in \mathbb{R}$ such that $\kappa^{(l)}_{t, s}(\xb, \xb') = \xb^{\intercal} \cdot K^{(l)}_{t, s} \cdot \xb'$ and $\kappaker{t}{s}{l}{\xb}{\xb'} = c^{(l)}_{t, s}$. Then, \eqref{eq:kgf} reduces to
\begin{equation*}
    \ddt \vb_{t} = \bigg ( \summ{l}{L} c^{(l)}_{t} K^{(l-1)}_{t}  \bigg )  \cdot \Sigma \cdot (\vb_{t} - \vb^{*})~,
\end{equation*}
where we set $K^{(0)}_{t} \coloneqq I_d$, $c^{(L)}_{t} \coloneqq 1$, and each $K^{(l-1)}_{t} \coloneqq K^{(l-1)}_{t, t}$, $c^{(l)}_{t} \coloneqq c^{(l)}_{t, t}$.
Furthermore, the mean-field NHL dynamics reduces to closed equations in $K^{(l)}_{t, s}$ and $c^{(l)}_{t, s}$. For example,
$\forall l \in [L-2]$, it holds that
\begin{align}
    K^{(l+1)}_{t, s} =&~ \int_{0}^t \int_{0}^s c^{(l+1)}_{r, p} K^{(l)}_{t, r} \cdot \zetab_r \cdot \zetab_p^{\intercal} \cdot K^{(l)}_{p, s} ~ dp ~dr~, \label{eq:K}\\
    c^{(l)}_{t, s} =&~ \int_{0}^t \int_{0}^s c^{(l+1)}_{t, r} c^{(l+1)}_{s, p} \zetab_r^{\intercal} \cdot K^{(l)}_{r, p} \cdot \zeta_p ~dp ~dr~ \label{eq:c},
\end{align}
The full system of equations is derived in Appendix~\ref{app:linear}.
We see that although $f_{t}$ is always a linear function (of its input), its training dynamics is nonlinear and non-Markovian, which is in contrast with, for example, the GF dynamics of plain linear regression:
\begin{equation*}
    \ddt \vb_{t} = \Sigma \cdot (\vb_{t} - \vb^{*})~.
\end{equation*}

\subsection{Numerical Illustrations}
\label{sec:numerical}
\subsubsection{Experiment $1$: Linear NN}
\begin{figure}
    \centering
    \includegraphics[scale=0.24]{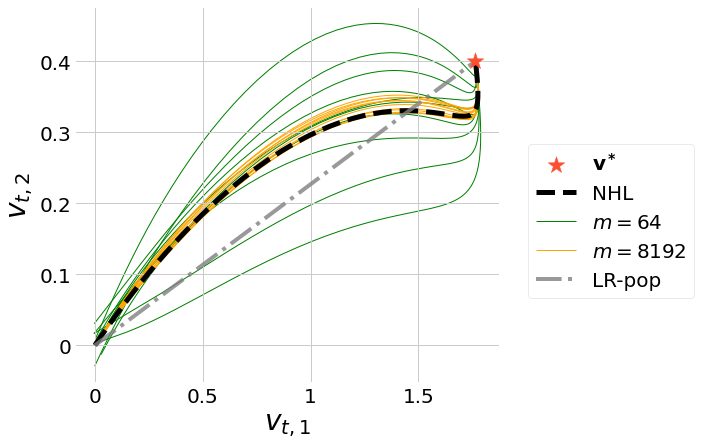}
    \caption{\textbf{Learning trajectories of linear $3$-layer NN versus the NHL dynamics}. \emph{Solid}: $3$-layer linear NNs trained by GD with width $64$ and $8192$. \emph{Dashed}: numerical integration of the NHL dynamics derived in Section~\ref{sec:linear}. \emph{Dot-dashed}: linear regression (LR) under population loss. }
    \label{fig:lin_traj}
\end{figure}
To validate the NHL dynamics derived above for linear NNs, we compare its numerical solution with the GD training of an actual $3$-layer linear NN on a least squares regression task of learning a linear target $\vb^{*}$, as described in Section~\ref{sec:linear}. We choose $d=10$, $n = 50$ and $\nu = \mathcal{N}(0, I_d)$. In Figure~\ref{fig:lin_traj}, we plot the learning trajectories in the linear model space projected into the first two dimensions, i.e., $v_{t, 1}$ and $v_{t, 2}$. We see that the NHL dynamics solved by numerical integration closely predicts the actual GD dynamics when the width is large. Moreover, the NHL dynamics presents a \emph{nonlinear} learning trajectory in the space of \emph{linear} models, which is in contrast with the linear learning trajectory of performing linear regression under the population loss.

\subsubsection{Experiment $2$: ReLU NN}
\begin{figure*}[h]
    \centering
    \begin{minipage}{.42\linewidth}
        \centering
        \includegraphics[scale=0.22]{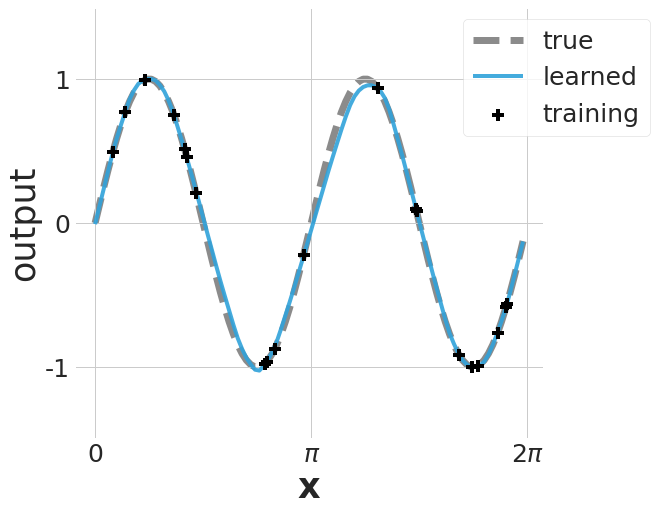}
    \end{minipage}
    \centering
    \begin{minipage}{.48\linewidth}
        \centering
        \includegraphics[scale=0.24]{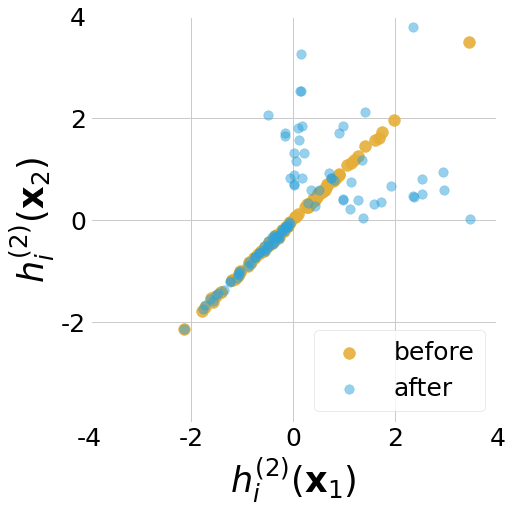}
    \end{minipage}
    \begin{minipage}{.44\linewidth}
        \centering
        (a) {Learned function}.
    \end{minipage}
    \begin{minipage}{.51\linewidth}
        \centering
        (b) {Pre-activation values.}
    \end{minipage}
    \\
    \begin{minipage}{\linewidth}
        
    \end{minipage}
    \\
    \centering
    \begin{minipage}{.44\linewidth}
        \centering
        \includegraphics[scale=0.23]{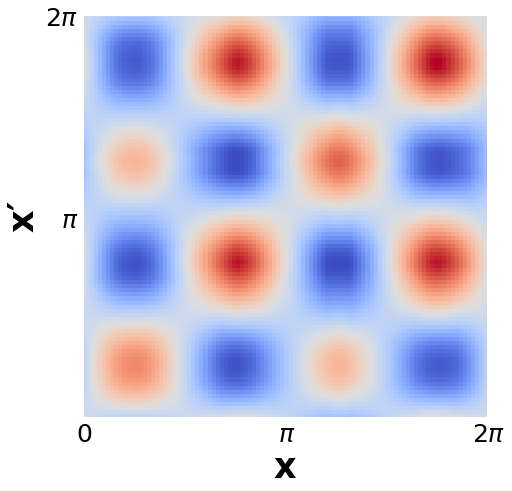}
    \end{minipage}
    \centering
    \begin{minipage}{.54\linewidth}
        \centering
        \includegraphics[scale=0.20]{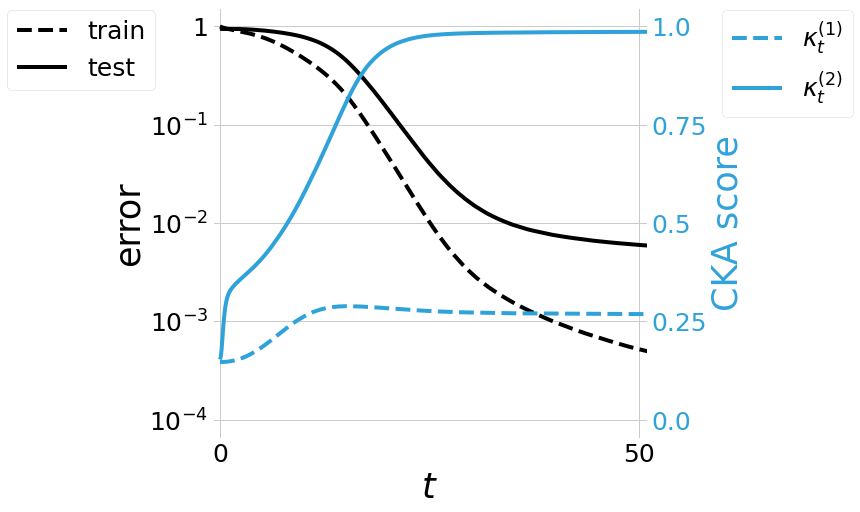}
    \end{minipage}
    \\
    \begin{minipage}{.48\linewidth}
        \centering
        (c) {$\kappa^{(2)}_{t}$ after training.}
    \end{minipage}
    \begin{minipage}{.50\linewidth}
        \centering
        (d) {Kernel-target alignment.}
    \end{minipage}
    \caption{\textbf{Results of GD training of $3$-layer NN with ReLU activation}. \textbf{(a)}: Target versus learned function. \textbf{(b)}: Pre-activation values across neurons in the second hidden layer on two training data points, $\xb_{1}$ and $\xb_{2}$, before and after training. \textbf{(c)}: The kernel function of the second hidden layer, $\kappa^{(2)}_{t}(\xb, \xb')$, after training (red means a higher value). \textbf{(d)}: Training and test errors and the CKA scores of $\kappa^{(1)}_{t}$ and $\kappa^{(2)}_{t}$ with respect to the target function over time, averaged over $10$ runs.}
    \label{fig:nonlin}
\end{figure*}
\noindent To gain insights into feature learning and the evolution of the NHL through training, we perform GD on $3$-layer NNs with the ReLU activation on a least squares regression task. We choose $d=1$, $n = 20$, $m=512$, the target function being $f^{*}(\xb) = \sin(2 \xb)$, and $\nu$ being the uniform distribution on $[0, 2 \pi]$. All parameters in the model, including untrained bias terms, are sampled i.i.d. from $\mathcal{N}(0, 1)$ at initialization. 

We see from Figure~\ref{fig:nonlin}(b) that the pre-activation values across all neurons in the second hidden layer---which correspond to $\Hb^{(2)}_{m, t}$ and approximate $\Hb^{(2)}_{t}$---move substantially through training, suggesting the occurrence of feature learning. In fact, as shown in Figure~\ref{fig:nonlin}(c), the movement results in a learned kernel function $\kappa^{(2)}_{t}$ that bears the same periodicity as the target function, showing that the kernel function is \emph{adaptive} through training. In particular, as measured by the Centered Kernel Alignment (CKA) score \citep{cortes2012algorithms},
$\kappa^{(2)}_{t}$ becomes more \emph{aligned} with the target function during training---an important notion in the literature of learning kernels \citep{christianini2001kernel, canatar2022kernel}---and more so than $\kappa^{(1)}_{t}$. It suggests that the space $\Hilb^{(L)}_{t}$ can move \emph{closer} to the target function via training, though a theoretical explanation for the alignment phenomenon is still lacking.

\section{Related Works}
\label{sec:related}
Below, we further discuss the novelty and significance of our work relative to prior literature.

\paragraph{Function space of width-unlimited NNs} 
As discussed in Section~\ref{sec:shallow_mf}, 
prior works have proposed the function space of shallow NNs
based on a type of variation norm, which is known to control the generalization and approximation errors
\citep{bach2017breaking, ma2018priori, siegel2020approximation, chen2020dynamical}. Other works have also established the regularity, spectral properties and representer theorems \citep{savarese2019infinite, ongie2020multivariate, parhi2021banach, siegel2021characterization, bartolucci2021understanding, 
unser2023ridges, sanford2023intrinsic, boursier2023penalising} of this space. Hence, for shallow NNs, a relatively complete picture has been established that covers approximation, generalization and optimization. 

For multi-layer NNs, however, a satisfactory theory for the function space and the complexity measure is missing. 
The neural tree space \citep{wojtowytsch2020banach} is an interesting proposal, but it does not correspond directly with the training of NNs since the neurons in this model do \emph{not} have shared preceding layers, which leads to approximation error bounds that grow exponentially in the depth, as discussed in Section~\ref{sec:approx}. Another line of studies including \citet{lee2017ability, sonoda2017double, bartlett2018gradient, zou2020gradient, lu2020mean, zou2020resnets, sonoda2021ghosts, dutordoir2021deep, littwin2021implicit, ma2022barron,ding2022deepresnet, hayou2022on, parhi2022dnnspline} has focused on 
NNs with ``bottleneck'' layers 
 that have fixed width, which can be viewed as expressing certain compositions or accumulations of functions in the Barron space. Though these models are interesting in their own right, the bottleneck assumption is not often obeyed in practice. In contrast, our work allows all layers of the NN to have unlimited width, resulting in a fundamentally different way of generalizing the Barron space---not by merely composing functions from the Barron space, but rather requiring a hierarchical generation of the RKHS.
A thorough comparison between them is left for future work.

\paragraph{Mean-field theory of NNs} In the mean-field scaling, shallow NNs under training are analogous to an interacting particles system \citep{rotskoff2018parameters, rotskoff2022cpam}. Hence, as described in Section~\ref{sec:shallow_wgf}, its infinite-width limit can be modeled as an integral over a probability measure on the parameter space (or, via an alternative form, a signed Radon measure), which evolves according to a Wasserstein GF during training \citep{mei2018mean, chizat2018global, sirignano2020mean_lln}. Notably, under suitable conditions, the Wasserstein GF can be proved to converge to global minimizers of the loss \citep{nitanda2017stochastic, mei2018mean, chizat2018global, rotskoff2018parameters, wojtowytsch2020convergence, chen2022on, boursier2022gradient}. 

Several works have then proposed mean-field-type models for multi-layer NNs via probability measures defined in different ways \citep{nguyen2019mean, araujo2019mean, nguyen2020rigorous, pham2020iclr, fang2021modeling, sirignano2022mean_deep}. In particular, \citet{araujo2019mean, nguyen2020rigorous} proved law-of-large-numbers results similar to Theorem~\ref{prop:lln} for the convergence of finite-width NNs to the mean-field limit, while \citet{nguyen2020rigorous, fang2021modeling} adopted a distributional view of the hidden-layer neurons that inspires our model. Nonetheless, these works have not addressed the function space and complexity measure associated with these models or proved error bounds. \citet{chen2022functional} proposed a mean-field theory on functional space to study the training dynamics of a type of partially-trained three-layer NNs in the infinite-width limit and defined a complexity measure based on optimal transport distances for distributions on functional space. The point of view therein sets the path for the current work.

\paragraph{NTK theory} If we replace the $1/m$ factor by $1/\sqrt{m}$ in \eqref{eq:f_m}, we arrive at what is commonly called the \emph{NTK scaling} of NNs. As shown by \citet{jacot2018neural}, if we initialize the NN randomly and take $m \to \infty$ under this scaling, then the pre-activation functions in the hidden layers barely move throughout training, and thus, the GF dynamics can be well-approximated by its linearization around the initialization, which is described by a kernel GF with a \emph{fixed} kernel (that is the NTK). In other words, the evolution of the output function can also be written as \eqref{eq:kgf} except that the kernel function $\theta_{t}$ is now independent of $t$. Thanks to this simplification, gradient descent is proved to converge to global minimum at a linear rate for over-parameterized NNs in the NTK regime \citep{du2019provably, du2019deep, allen2019convergence, zou2020gradient, oymak2020moderate}. Furthermore, generalization guarantees can be proved for such models through the learning theory of RKHS \citep{arora2019fine, cao2019generalization, e2020comparative}. 

However, the fact that the hidden layer neurons and hence the kernel function remain fixed to their initialization indicates a lack of \emph{feature learning}. For this reason, the NTK limit is described as a regime of ``\emph{lazy training}'' \citep{chizat2019lazy, woodworth2020kernel}, and the NTK theory does not satisfy criterion \eqref{item:4}. Several works have studied the limitations of the NTK regime compared to feature-learning regimes, both theoretically \citep{ghorbani2019limitations, ghorbani2020neural, wei2019regularization, woodworth2020kernel, liu2020linearity, luo2021phase} and empirically \citep{geiger2019disentangling, lee2020infinite}.

\paragraph{NNs as kernels} Besides the NTK theory, a number of prior works have also explored the connection between NNs and kernels, by either proposing new kernel methods inspired by NNs \citep{cho2009kernel, mairal2014convolutional, wilson2016deep, bietti2017invariance, shankar2020without, radhakrishnan2024mechanism} or by modeling NNs as kernels \citep{rahimi2008weighted, montavon2011kernel, hazan2015steps, anselmi2015deep, domingos2020every, aitchinson2021dkkp, amid2022nrpk, seroussi2023separation, yang2023kernel}, or both. As an exemplar of the latter category, \citet{daniely2016toward} proposed the conjugate kernel model of multi-layer NNs together with a random feature scheme for approximating the kernel \citep{daniely2017random} and a theoretical guarantee for learning with stochastic gradient descent (SGD) in polynomial time \citep{daniely2017sgd}. However, like other efforts to understand NNs as kernel methods, it does not satisfy properties \eqref{item:1} or \eqref{item:4} as a theory for the function space of multi-layer NNs. In fact, under our framework, the conjugate kernel space can be seen as a particular \emph{fixed} NHL determined by the random initialization of the weights, which is therefore quite restrictive. In contrast, the function space $\Ladder{L}$ is not \emph{one} RKHS but an infinite collection of them and hence richer. 

In the NTK scaling, a randomly-initialized NN in the infinite-width limit can also be viewed representing a function sampled from a Gaussian process whose covariance function is connected to the NTK, thus leading to a Bayesian interpretation \citep{neal1996bayesian, williams1996computing, lee2017deep, matthews2018gaussian, garriga-alonso2018deep, borovykh2018gaussian, novak2019bayesian, agrawal2020wide}. In particular, \citet{lee2019wide} showed that SGD training corresponds to a linear dynamics of the Gaussian process and mimics Bayesian inference. However, like the NTK theory (and in contrast with ours), this analysis relies on a linear approximation of the training dynamics close to initialization and hence does not exhibit feature learning, either. Though our NHL framework also involves random fields, the fundamental difference can be seen through the fact that, in our case, while the \emph{hidden layers} are modeled as random fields, the \emph{output function} is always deterministic.

\paragraph{Complexity measures of NNs}
With large numbers of parameters, NNs in practice often have enough capacity to fit data with even random labels \citep{zhang2016understanding}. Hence, to derive meaningful generalization bounds, complexity measures of NNs that do no depend on the network size have been proposed.
For example, several complexity measures based on certain norms of the parameters have been proposed, for both shallow \citep{bartlett1998sample, koltchinskii2002empirical, bartlett2002rademacher, rosset2007l1, cho2009kernel} and multi-layer ones \citep{neyshabur2015norm, bartlett2017spectrally}, which lead to generalization bounds that are independent of the number of parameters. In particular, the group norm in \citet{neyshabur2015norm} is closely related to the NHL norm proposed in our work, as the NHL norm of the function represented by an NN can be bounded by the group norm of the NN. Thus, the NHL norm can be viewed as a generalization of the group norm to the continuous, width-unlimited setup via the NHL model. In practice, regularizing the parameter norms through weight decay has long been proposed as a strategy to improve the performance of NNs (e.g., \citealp{hinton1987learning, krogh1991simple, lee2020infinite}).

\paragraph{Beyond lazy training} There have been efforts on extending the NTK analysis beyond the lazy training regime by considering higher-order Taylor expansions of the GD dynamics or corrections to the NTK due to a finite width or a large depth \citep{allenzhu2019going, huang2019nth, bai2020beyond, hanin2020finite, yaida2020nongaussian, ryh2022principles, hanin2022correlation}, though the function space implication of these proposals is not clear. Meanwhile, several works have studied the effect of different scaling choices on the behavior of the infinite-width limit \citep{golikov2020towards, luo2021phase, zhou2022empirical}. In particular, \citet{yang2021feature} proposed another choice called the maximum-update scaling, which exhibits feature learning while avoiding the degeneracy of the mean-field scaling mentioned in Remark~\ref{rmk:degen}. A few works have studied the training dynamics in the infinite-width limit under this scaling \citep{yang2021feature, golikov2022nongaussian, hajjar2021training, ba2022high, bordelon2022selfconsistent, chizat2022infinite}, but the function space induced by this model is unclear except when only the penultimate layer is trained \citep{chen2022functional}.

\paragraph{Training dynamics of deep linear NNs} Many prior studies have examined the GD or GF dynamics of deep linear NNs \citep{saxe2014linear, jacot2021saddle}, including deriving their global convergence guarantees \citep{kawaguchi2016deep, du2019width, eftekhari2020training, bah2022learning} and implicit bias \citep{gunasekar2017implicit, ji2018gradient, arora2019implicit, gidel2019implicit, li2021towards}. The infinite-width limit of deep linear NNs under the maximum-update scaling have been studied in \citet{bordelon2022selfconsistent, chizat2022infinite}. In particular, the latter work derived the limiting dynamics and implicit bias rigorously, and our Figure~\ref{fig:lin_traj} is inspired by the figures therein. Meanwhile, we are not aware of prior studies on deep linear NNs in the mean-field limit.

\paragraph{Depth separation} A number of works have studied the benefit of depth in NN approximation, with examples including \citet{telgarsky2016benefits, eldan2016power, daniely2017depth, safran2017depth, yarotsky2017error, vardi2020barriers, venturi2022depth}. While most of these studies quantifies the price of approximation by the width required in the NN, a few have also considered NNs in the infinite-width limit. For example, \citet{ongie_function_2019} constructs a function that does not belong to the Barron space under ReLU activation but can be represented by a finite-width three-layer NN, which is directly used in our proof of Proposition~\ref{prop:relu_sep}. In addition, \citet{ren2023depth} shows a function that cannot be approximated by shallow NNs but can be learned by an infinite-width three-layer NN, albeit of the type mentioned above with a bottleneck layer.

\section{Conclusions} In this work, we propose to model multi-layer NNs as ladders of RKHSs, thereby defining the function space of multi-layer NNs as an infinite union of  hierarchically-generated RKHSs. It is associated with a complexity measure of functions which depends on the number of layers and governs the approximation and generalization errors for functions in this space. Moreover, the training of multi-layer NNs in a feature-learning regime translates to a particular learning dynamics of the NHLs.  
Hence, our proposal emerges as a desirable candidate for the function space of multi-layer NNs and establishes a rigorous
foundation for their further studies.

There are several limitations of our work, including the various assumptions on the activation function $\sigma$---the approximation and generalization guarantees in Sections~\ref{sec:real_approx} and \ref{sec:gen} are most tailored for $1$-homogeneous activation functions (such as ReLU), whereas the analysis of the training dynamics in Section~\ref{sec:training} assumes that the activation function is differentiable (thus excluding ReLU)---and that we ignore the effects of finite step sizes and stochasticity in practical NN training. Meanwhile, our work opens up interesting directions for future research, including additional properties of the NHL space, approximation theory in other metrics and formulations (e.g. \citealp{klusowski2018approximation, e2020towards}), generalization guarantees beyond the analysis of Rademacher complexity, 
and long-time behaviors of the mean-field NHL dynamics when $L>2$. Importantly, we hope it sets the stage for quantitative and rigorous probes into the popular idea that deep NNs perform hierarchical learning \citep{poggio2003mathematics, poggio2020pnas, allen2020backward, chen2020towards}.

\paragraph{Acknowledgements}
The author is indebted to Joan Bruna and Eric Vanden-Eijnden for earlier discussions and thankful to Pengning Chao and the anonymous reviewers for feedbacks on the manuscript.

\bibliography{ref}
\newpage
\appendix

\section{Supplementary Materials for Section~\ref{sec:nhl}}
\label{app:pf_sec_nhl}

\subsection{Proof of Theorem~\ref{prop:nhl_prop}(\ref{prop:vs})}
\label{app:pf_vs}
    For any function $f$, it is obvious by the property of $L^p$ norms that $\CompLp{L}{p}(f) \geq \CompLp{L}{p'}(f)$ if $p \geq p'$. 

     Let $p \in (0, \infty]$.  Suppose $f$ is a function in $\Ladder{L}_p$. By definition, $\exists \mu^{(1)}, ..., \mu^{(L-1)}$ such that $\| f \|_{\Hilb^{(L)}} < \infty$ and
    $\forall l \in [L-1]$, 
    $\| \mu^{(l)} \|_{\Hilb^{(l)}, p} < \infty$
    and $\Hilb^{(l+1)} \coloneqq \Hilb_{\mu^{(l)}}$.  Given any $c \in \Rbb$, the function $cf \in \Hilbl{L}$ with $\| cf \|_{\Hilbl{L}} = |c| \| f \|_{\Hilb^{(L)}}$, which implies that $c f$ belongs to the same NHL as $f$ and  $\CompLp{L}{p}(c f)\leq |c| \CompLp{L}{p}(f) < \infty$. This shows that $\Ladder{L}_p$ is closed under scalar multiplication. Meanwhile, $\CompLp{L}{p}(f) = \CompLp{L}{p}(c^{-1} cf) \leq |c|^{-1} \CompLp{L}{p}(c f)$. As a result, $\CompLp{L}{p}(c f) = |c| \CompLp{L}{p}(f)$. This proves the absolute homogeneity of $\CompLp{L}{p}$.
    
    Let $f'$ be another function in $\Ladder{L}_p$. Similarly, by definition, $\exists \mu^{(1)'}, ..., \mu^{(L-1)'}$ such that
     $\| f \|_{\Hilb^{(L)'}} < \infty$, $\Hilb^{(1)'} = \Hilb^{(1)}$ and $\forall l \in [L-1]$, 
     $\| \mu^{(l)} \|_{\Hilb^{(l)}, p} < \infty$
     and $\Hilb^{(l+1)'} \coloneqq \Hilb_{\mu^{(l)'}}$.
    Then, to show that $\Ladder{L}_p$ is a vector space, we need an upper bound on $\CompLp{L}{p}(f + f')$.

    For $l \in [L-1]$, we define 
    $\tilde{\mu}^{(l)} \coloneqq \frac{1}{2} \mu^{(l)} + \frac{1}{2} \mu^{(l)'}$.
    Thus, $\tilde{\mu}^{(l)}$ is supported within $\Hilbl{l} \cup \Hilb^{(l)'}$.
    We define $\tilde{\Hilb}^{(1)} \coloneqq \Hilbl{1}$ and $\forall l \in [L-1]$, $\tilde{\Hilb}^{(l+1)} \coloneqq \Hilb_{\tilde{\mu}^{(l)}}$, and we will show that $f + f'$ belongs to the NHL formed by $(\Tilde{\Hilb}^{(l)})_{l \in [L]}$. To do so, we need the following lemma:
    \begin{lemma}
    \label{lem:measure_{t}ransform}
        Let $\mu_{1}, \mu_{2} \in \Pcal(\Conti)$. Suppose $\mu_{1}$ is absolutely continuous with respect to $\mu_{2}$ and the Radon-Nikodym derivative $\frac{d \mu_{1}}{d \mu_{2}}$ is upper-bounded by $M > 0$ on the support of $\mu_{1}$. Then it holds that $\Hilb_{\mu_{1}} \subseteq \Hilb_{\mu_{2}}$ with $\| \cdot \|_{\Hilb_{\mu_{2}}} \leq \sqrt{M} \| \cdot \|_{\Hilb_{\mu_{1}}}$.
    \end{lemma}
    \noindent This lemma is proved in Appendix~\ref{app:pf_lem_measure_{t}ransform}. For each $l \in [L-1]$, noticing that $\mu^{(l)}$ (respectively, $\mu^{(l)'}$) are absolutely continuous with respect to $\tilde{\mu}^{(l)}$ with $\frac{d \mu^{(l)}}{d \tilde{\mu}^{(l)}} \leq 2$ on $\Hilb^{(l)}$ (respectively, $\frac{d \mu^{(l)'}}{d \tilde{\mu}^{(l)}} \leq 2$ on $\Hilb^{(l)'}$ ), we can apply Lemma~\ref{lem:measure_{t}ransform} to obtain that,
    \begin{equation*}
        \begin{split}
            \| \tilde{\mu}^{(l)} \|_{\tilde{\Hilb}^{(l)}, p}^p =&~ \int \| h \|_{\tilde{\Hilb}^{(l)}}^p \tilde{\mu}^{(l)} (dh) \\
            =&~ \frac{1}{2} \int \| h \|_{\tilde{\Hilb}^{(l)}}^p \mu^{(l)} (dh) + \frac{1}{2} \int \| h \|_{\tilde{\Hilb}^{(l)}}^p \mu^{(l)'} (dh) \\
            \leq &~ 2^{p/2-1} \int \| h \|_{\Hilb^{(l)}}^p \mu^{(l)} (dh) + 2^{p/2-1} \int \| h \|_{\Hilb^{(l)}}^p \mu^{(l)'} (dh)~,
        \end{split}
    \end{equation*}
    and in addition, 
    \begin{align*}
        \| \tilde{\mu}^{(l)} \|_{\tilde{\Hilb}^{(l)}, \infty} =&~ \max \left \{\| \mu^{(l)} \|_{\tilde{\Hilb}^{(l)}, \infty}, \|\mu^{(l)'} \|_{\tilde{\Hilb}^{(l)}, \infty} \right \} \\
        \leq &~ \sqrt{2} \max \left ( \| \mu^{(l)} \|_{\Hilb^{(l)}, \infty} + \| \mu^{(l)'} \|_{\tilde{\Hilb}^{(l)}, \infty} \right )~.
    \end{align*}
    Moreover, $\| f + f' \|_{\Tilde{\Hilb}^{(L)}} \leq \| f \|_{\Tilde{\Hilb}^{(L)}} + \| f' \|_{\Tilde{\Hilb}^{(L)}} \leq \sqrt{2} (\| f \|_{\Hilb^{(L)}} + \| f' \|_{\Hilb^{(L)'}})$.
    Therefore, $\forall p > 0$,
    \begin{equation}
    \label{eq:CompLp_{t}ilde_bdd}
        \begin{split}
            \big ( \CompLp{L}{p}(f+f') \big )^p \leq &~ \Big ( \prod_{l=1}^{L-1}  \| \tilde{\mu}^{(l)} \|_{\tilde{\Hilb}^{(l)}, p}^p \Big ) \| f + f' \|_{\tilde{\Hilb}^{(L)}}^p \\
            \leq &~ 2^{pL/2 - L + 1} \left ( \prod_{l=1}^{L-1} \Big (  \| \tilde{\mu}^{(l)} \|_{\Hilb^{(l)}, p}^p +  \| \mu^{(l)'} \|_{\tilde{\Hilb}^{(l)}, p}^p \Big ) \right ) \left (\| f \|_{\Hilb^{(L)}} + \| f' \|_{\Hilb^{(L)'}} \right )^p \\
            < & \infty~,
        \end{split}
    \end{equation}
and in addition,
\begin{align*}
    \CompLp{L}{\infty}(f+f') \leq &~ \Big ( \prod_{l=1}^{L-1}  \| \tilde{\mu}^{(l)} \|_{\tilde{\Hilb}^{(l)}, \infty} \Big ) \| f + f' \|_{\tilde{\Hilb}^{(L)}} \\
    \leq &~ 2^{L/2} \Big ( \prod_{l=1}^{L-1}  \| \mu^{(l)} \|_{\Hilb^{(l)}, \infty} + \| \mu^{(l)'} \|_{\tilde{\Hilb}^{(l)}, \infty} \Big ) \| f \|_{\tilde{\Hilb}^{(L)}} + \| f' \|_{\tilde{\Hilb}^{(L)}} \\
    < &~ \infty~.
\end{align*}
Hence, $f + f' \in \Ladder{L}_p$, and this concludes the proof that $\Ladder{L}_p$ is a vector space, $\forall p \in (0, \infty]$.

\subsubsection{Proof of Lemma~\ref{lem:measure_{t}ransform}}
\label{app:pf_lem_measure_{t}ransform}
Suppose $f \in \Hilb_{\mu_{1}}$. By Lemma~\ref{lem:equiv}, $\exists \xi \in L^2(\Conti, \mu_{1})$ such that $f = \int \xi(h) \sigma(h(\cdot)) \mu_{1}(dh)$ and $\| f \|_{\Hilb_{\mu_{1}}} = \| \xi \|_{L^2(\Conti, \mu_{1})}$. By the definition of Radon-Nikodym derivative, it then holds for all $\xb \in \domX$ that
\begin{equation*}
    \begin{split}
        f(\xb) =&~ \int  \xi(h) \sigbig{h(\xb)} \frac{d \mu_{1}}{d \mu_{2}}(h) \mu_{2}(dh) \\
        =&~ \int \left ( \xi(h) \frac{d \mu_{1}}{d \mu_{2}}(h) \right ) \sigbig{h(\xb)} \mu_{2}(dh)~.
    \end{split}
\end{equation*}
Hence, there is
\begin{equation*}
    \begin{split}
        \| f \|_{\Hilb_{\mu_{2}}}^2 \leq &~ \int \left | \xi(h) \frac{d \mu_{1}}{d \mu_{2}}(h) \right |^2 \mu_{2}(dh) \\
        =&~ \int \left | \xi(h) \right |^2 \frac{d \mu_{1}}{d \mu_{2}}(h) \mu_{1}(dh) \\
        \leq &~ M \| \xi \|_{L^2(\Conti, \mu_{1})}^2 \\
        = &~ M \| f \|_{\Hilb_{\mu_{1}}}^2~.
    \end{split}
\end{equation*}

\subsection{Proof of Theorem~\ref{prop:nhl_prop}(\ref{prop:infty_norm})}
\label{app:pf_prop_infty_norm}
In light of Theorem~\ref{prop:nhl_prop}(\ref{prop:vs}), it suffices to consider the case $p = 2$. The basic regularity properties of functions in RKHS given below is simple to show using the reproducing property and the Cauchy-Schwartz inequality:
\begin{lemma}
    \label{lem:rkhs_reg}
        Let $\Hilb$ be an RKHS on $\domX$ associated with the kernel function $\kappa$. Then $\forall f \in \Hilb$, 
        \begin{align*}
            |f(\xb) | \leq &~ \| f \|_{\Hilb} (\kappa(\xb, \xb))^{1/2}~, \quad \forall \xb \in \domX~, \\
            |f(\xb) - f(\xb') | \leq &~ \| f \|_{\Hilb}  d_{\kappa}(\xb, \xb')~, \quad \forall \xb, \xb' \in \domX~, 
        \end{align*}
        where we define $d_{\kappa}(\xb, \xb') \coloneqq (\kappa(\xb, \xb) + \kappa(\xb' , \xb') - 2 \kappa(\xb, \xb'))^{1/2}$.
\end{lemma}
\noindent Another central observation is the following lemma, which is proved in Appendix~\ref{app:pf_lem_infty_induction}:

\begin{lemma}
\label{lem:infty_induction}
    If $\sigma$ is $J_{\sigma}$-Lipschitz and $\sigma(0) = 0$, then $\forall L \in \Nbb$,
    \begin{align*}
        \left ( \kappa^{(L)}(\xb, \xb) \right )^{1/2} \leq &~ (J_{\sigma})^L \| \xb \| \prod_{l=1}^{L} \| \mu^{(l)} \|_{\Hilb^{(l)}, 2}~, \quad \forall \xb \in \domX~, \\
        d_{\kappa^{(L)}}(\xb, \xb') \leq &~ (J_{\sigma})^L \| \xb - \xb' \|  \prod_{l=1}^{L} \| \mu^{(l)} \|_{\Hilb^{(l)}, 2}~, \quad \forall \xb, \xb' \in \domX~,
    \end{align*}
\end{lemma}

\noindent Suppose $f \in \Ladder{L}$, and let $(\Hilb^{(l)})_{l \in [L]}$ be an NHL to which it belongs. Then, Lemmas~\ref{lem:rkhs_reg} and \ref{lem:infty_induction} allow us to derive that

\begin{align}
    |f(\xb)| \leq&~ (J_{\sigma})^{L-1} \| \xb \| \| f \|_{\Hilb^{(L)}} \prod_{l=1}^{L-1} \| \mu^{(l)} \|_{\Hilb^{(l)}, 2}~, \quad \forall \xb \in \domX~, \label{eq:f_bounded}\\
    |f(\xb) - f(\xb')| \leq&~ (J_{\sigma})^{L-1} \| \xb - \xb' \| \| f \|_{\Hilb^{(L)}} \prod_{l=1}^{L-1} \| \mu^{(l)} \|_{\Hilb^{(l)}, 2}~, \quad \forall \xb, \xb' \in \domX~. \label{eq:f_lipschitz}
\end{align}
\noindent Theorem~\ref{prop:nhl_prop}(\ref{prop:infty_norm}) thus follows if we minimize the right-hand side of \eqref{eq:f_bounded} and \eqref{eq:f_lipschitz} over all NHLs.

\subsubsection{Proof of Lemma~\ref{lem:infty_induction}}
\label{app:pf_lem_infty_induction}
We can prove Lemma~\ref{lem:infty_induction} inductively in $L$. 
By the definition of $\kappa^{(0)}$, there is

\begin{align*}
    \kappa^{(0)}(\xb, \xb) =&~ \| \xb \|^2~, \quad \forall \xb \in \domX~, \\
    d_{\kappa^{(0)}}(\xb, \xb') =&~ \left ( \| \xb \|^2 + \| \xb' \|^2 - 2 \xb^{\intercal} \cdot \xb' \right )^{1/2} \\
    =&~ \| \xb - \xb' \|~,  \quad \forall \xb, \xb' \in \domX~.
\end{align*}

Next, suppose that the statements of Lemma~\ref{lem:infty_induction} hold for a certain $L \in \Nbb$. Then, for $L+1$, Lemma~\ref{lem:rkhs_reg} together with the assumptions on $\sigma$ implies that,
$\forall \xb \in \domX$,

\begin{equation*}
    \begin{split}
        \kappa^{(L+1)}(\xb, \xb) \leq &~ \int \big | \sigbig{h(\xb)} \big |^2 \mu^{(L+1)}(dh) \\
        \leq &~ (J_{\sigma})^2 \int \big | h(\xb) \big |^2 \mu^{(L+1)}(dh) \\
        \leq &~ (J_{\sigma})^2 \left ( \int \| h \|_{\Hilb^{(L+1)}}^2 \mu^{(L+1)}(dh) \right ) \kappa^{(L)}(\xb, \xb) \\
        \leq &~ (J_{\sigma})^2 \left ( \int \| h \|_{\Hilb^{(L+1)}}^2 \mu^{(L+1)}(dh) \right ) (J_{\sigma})^{2L} \| \xb \|^2 \prod_{l=1}^L \| \mu^{(l)} \|_{\Hilb^{(l)}, 2}^2 \\
        =&~ (J_{\sigma})^{2L+2} \| \xb \|^2 \prod_{l=1}^{L+1} \| \mu^{(l)} \|_{\Hilb^{(l)}, 2}^2~,
    \end{split}
\end{equation*}

and moreover, $\forall \xb, \xb' \in \domX$,

\begin{equation*}
    \begin{split}
        d_{\kappa^{(L+1)}}(\xb, \xb')^2 \leq&~ \int \left | \sigbig{h(\xb)} - \sigbig{h(\xb')} \right |^2 \mu^{(L+1)}(dh) \\
        \leq&~ (J_{\sigma})^2 \int \left | h(\xb) - h(\xb') \right |^2 \mu^{(L+1)}(dh) \\
        \leq &~ (J_{\sigma})^2 \left ( \int \| h \|_{\Hilb^{(L+1)}}^2 \mu^{(L+1)}(dh) \right ) \big ( d_{\kappa^{(L)}}(\xb, \xb') \big )^2 \\
        \leq &~ (J_{\sigma})^2 \left ( \int \| h \|_{\Hilb^{(L+1)}}^2 \mu^{(L+1)}(dh) \right ) (J_{\sigma})^{2L} \| \xb - \xb' \|^2 \prod_{l=1}^L \| \mu^{(l)} \|_{\Hilb^{(l)}, 2}^2 \\
        \leq&~ (J_{\sigma})^{2L+2} \| \xb - \xb' \|^2 \prod_{l=1}^{L+1} \| \mu^{(l)} \|_{\Hilb^{(l)}, 2}^2~.
    \end{split}
\end{equation*}

which proves the statements for $L+1$.

\subsection{Proof of Theorem~\ref{prop:nhl_prop}(\ref{prop:p_equiv})}
\label{app:pf_prop_p_equiv}
First, we show the equivalence for different $p \in [2, \infty]$. For any functions $f$, it is obvious that $\CompLp{L}{p}(f) \geq \CompLp{L}{p'}(f)$ if $p \geq p'$. Then, to prove Theorem~\ref{prop:nhl_prop}(\ref{prop:p_equiv}), it suffices to show that for all $f$ such that $\CompLp{L}{2}(f) < \infty$, we can find an NHL $(\Hilb^{(l)})_{l \in [L]}$ induced by $(\mu^{(l)})_{l \in [L-1]}$ such that  $\|f \|_{\Hilb^{(L)}} = \CompLp{L}{2}(f)$ and $\forall l \in [L-1]$, $\mu^{(l)}$ is supported on $\hat{\Hilb}^{(l)}$, the unit-norm sphere of $\Hilb^{(l)}$
(in which case, $\| \mu^{(l)} \|_{\Hilb^{(l)}, p} = 1$, $\forall p \in [2, \infty]$).

To this end, we introduce the following lemma, which is proved in Appendix~\ref{app:pf_lem_Htilde}:
\begin{lemma}
\label{lem:Htilde}
    Suppose that $\sigma$ is $1$-homogeneous. Let $\Hilb$ be a Hilbert space of functions on $\domX$, and let $\hat{\Hilb}$ denote the unit-norm sphere of $\Hilb$. Given any $\mu \in \Pcal(\Hilb)$ such that $\| \mu \|_{\Hilb, 2} = 1$, there exists $\Tilde{\mu} \in \Pcal(\hat{\Hilb})$ such that 
    $\| \cdot \|_{\Hilb_{\Tilde{\mu}}} \leq \| \cdot \|_{\Hilb_{\mu}}$.
\end{lemma}
\noindent Suppose $\CompLp{L}{2}(f) < \infty$. By definition, we can find $\mu^{(1)}$, ..., $\mu^{(L-1)}$ such that $\CompLp{L}{2}(f) = \| f \|_{\Hilb^{(L)}} \prod_{l=1}^{L-1}  \| \mu^{(l)} \|_{\Hilb^{(l)}, 2}$ and $\forall l \in [L-1]$, $\Hilb^{(l+1)} = \Hilb_{\mu^{(l)}}$. Thanks to the homogeneity of $\sigma$, we may assume without loss of generality that $\| f \|_{\Hilb^{(L)}} = \CompLp{L}{2}(f)$ while $\forall l \in [L-1]$, $\| \mu^{(l)} \|_{\Hilb^{(l)}, 2} = 1$. Then, assume for contradiction that for some $l \in [L-1]$, $\mu^{(l)}$ is not supported entirely within $\hat{\Hilb}^{(l)}$.
But Lemma~\ref{lem:Htilde} implies that there exists a probability measure
$\tilde{\mu}^{(l)}$ supported within $\hat{\Hilb}^{(l)}$
such that if we define $\tilde{\Hilb}^{(l+1)} \coloneqq \Hilb_{\Tilde{\mu}^{(l)}}$, then $\forall f' \in \Hilb^{(l+1)}$, $\| f' \|_{\tilde{\Hilb}^{(l+1)}} \leq \| f' \|_{\Hilb^{(l+1)}}$. In particular, this implies that $\| \mu^{(l+1)} \|_{\tilde{\Hilb}^{(l+1)}, 2} \leq \| \mu^{(l+1)} \|_{\Hilb^{(l+1)}, 2}$. Hence, by replacing $\Hilb^{(l+1)}$ with $\tilde{\Hilb}^{(l+1)}$, we obtain another NHL
which contains $f$ and also realizes the minimization problem in $\CompLp{L}{2}(f)$.

Applying this argument to each $l$, we see that $\mu^{(1)}$, ..., $\mu^{(L-1)}$ can be chosen such that $\forall l \in [L-1]$, $\mu^{(l)}$ is supported within the unit-norm sphere of $\Hilb^{(l)}$, and hence $\| \mu^{(l)} \|_{\Hilb^{(l)}, p} = \| \mu^{(l)} \|_{\Hilb^{(l)}, 2} = 1$, $\forall p \geq (2, \infty]$. This proves that $\forall p \in (2, \infty]$, $\CompLp{L}{p} = \CompLp{L}{2}$.

Having shown the equivalence of $\CompLp{L}{p}$ for $p \in [2, \infty]$ if $\sigma$ is $1$-homogeneous, below we only need to consider the case $p = 2$.  Let us first recall the definition of a quasi-Banach space (e.g., \citealp{kalton2003quasi}):
\begin{definition}
\label{def:quasib}
    Let $\Uilb$ be a vector space. A functional $\| \cdot \|_{\Uilb}: \Uilb \to [0, \infty)$ is called a quasi-norm on $\Uilb$ with a constant $K \geq 1$ if the following properties are satisfied:
    \begin{itemize}
        \item Positive-definiteness: $\forall f \in \Uilb$, $\| f \|_\Uilb > 0$ if $f \neq 0$;
        \item Absolute homogeneity: $\forall f \in \Uilb$, $\forall c \in \Rbb$, $\| c f \|_\Uilb = | c | \| f \|_\Uilb$;
        \item Modified triangle inequality: $\forall f, g \in \Uilb$, $\| f + g \|_\Uilb \leq K (\| f \|_\Uilb + \| g \|_\Uilb)$.  
    \end{itemize}
    Then, $\Uilb$ is called a quasi-Banach space if it is equipped with a quasi-norm $\| \cdot \|_{\Uilb}$ and is complete with respect to the topology induced by $\| \cdot \|_{\Uilb}$.
\end{definition}

The positive-definiteness of $\CompLp{L}{2}$ is implied by Theorem~\ref{prop:nhl_prop}\eqref{prop:infty_norm} and its absolute homogeneity has been shown in Appendix~\ref{app:pf_vs}. To show that $\CompLp{L}{2}$ satisfies the modified triangle inequality, we follow the construction in Appendix~\ref{app:pf_vs}. Given $f$ and $f' \in \Ladder{L}_{2}$, since $\sigma$ is $1$-homogeneous, we may assume without loss of generality that $\| f \|_{\Hilb^{(L)}} = \CompLp{L}{2}(f)$, $\| f \|_{\Hilb^{(L)'}} = \CompLp{L}{2}(f')$, and $\forall l \in [L-1]$, $\| \mu^{(l)} \|_{\Hilb^{(l)}, 2} = \| \mu^{(l)'} \|_{\Hilb^{(l)'}, 2} = 1$. Then, \eqref{eq:CompLp_{t}ilde_bdd} can be tightened to yield
\begin{equation*}
    \begin{split}
        \CompLp{L}{2}(f+f') \leq 2^{\frac{L}{2}} \left ( \| f \|_{\Hilb^{(L)}} + \| f' \|_{\Hilb^{(L)'}} \right ) = 2^{\frac{L}{2}} \left ( \CompLp{L}{2}(f) + \CompLp{L}{2}(f')  \right )~,
    \end{split}
\end{equation*}
which means that $\CompLp{L}{2}$ is a quasi-norm on $\Ladder{L}_{2}$  (with a constant $2^{\frac{L}{2}}$). 

Hence, to prove that $\Ladder{L}_{2}$ is a quasi-Banach space, it remains to show that it is complete under $\CompLp{L}{2}$ as the quasi-norm. Let $(f_{k})_{k=1}^\infty$ be a Cauchy sequence in $\Ladder{L}_{2}$. By definition, $\forall \epsilon > 0$, $\exists N(\epsilon) \in \Nbb_+$ such that $\forall n_{1}, n_{2} > N(\epsilon)$, $\CompLp{L}{2}(f_{n_{1}} - f_{n_{2}}) < \epsilon$. For each $k \in \Nbb_+$, we define $\epsilon_{k} \coloneqq 2^{-(k+1)L}$ and $n_{k} \coloneqq \ceil[]{N(\epsilon_{k})}$. We then define $g_{0} \coloneqq f_{n_{1}}$ and $g_{k} \coloneqq f_{n_{k+1}} - f_{n_{k}}$ for each $k \in \Nbb_+$. Thus, by construction, it holds that $\CompLp{L}{2}(g_{k}) < \epsilon_{k}$, $\forall k \in \Nbb_+$. By the argument above, $\forall k \in \Nbb$, there exists an NHL $(\Hilb^{(l)}_{k})_{l \in [L]}$ induced by $(\mu^{(l)}_{k})_{l \in [L-1]}$ such that $\| g_{k} \|_{\Hilb^{(L)}_{k}} = \CompLp{L}{2}(g_{k})$ and $\mu^{(l)}_{k}$ is supported within $\hat{\Hilb}^{(l)}_{k}$, $\forall l \in [L-1]$.

By Theorem~\ref{prop:nhl_prop}\eqref{prop:infty_norm}, when evaluated at any $\xb \in \domX$, the sequence $(f_{k}(\xb))_{k=1}^\infty$ is a Cauchy sequence in $\Rbb$ and hence admits a limit, which allows us to define a function $f_\infty$ on $\domX$ by $f_\infty(\xb) \coloneqq \lim_{k \to \infty} f_{k}(\xb)$, $\forall \xb \in \domX$.
Thus, it remains to show that $f_\infty \in \Ladder{L}_{2}$. 

$\forall l \in [L-1]$, we define
\[
\mu^{(l)} \coloneqq (2^l - 1) \sum_{k=0}^\infty 2^{-(k+1)l} \mu^{(l)}_{k}~.
\]
It is straightforward to show that the series converges in total variation as a measure on $\Conti$. In addition, as $\sum_{k=0}^\infty 2^{-(k+1)l} = \frac{1}{2^l - 1}$, it further holds that 
\begin{equation*}
    \mu^{(l)}(\Conti) = (2^l - 1) \sum_{k=0}^\infty 2^{-(k+1)l} \mu^{(l)}_{k}(\Hilb^{(l)}_{k}) = 1.
\end{equation*}
This implies that $\mu^{(l)} \in \Pcal(\Conti)$ and allows us to define $\Hilb^{(l+1)} \coloneqq \Hilb_{\mu^{(l)}}$ for each $l$. Thus, to prove that $f_\infty \in \Ladder{L}_{2}$, it suffices to show that $\| f_\infty \|_{\Hilb^{(L)}} < \infty$ and $\forall l \in [L-1]$, $\| \mu^{(l)} \|_{\Hilb^{(l)}, 2} < \infty$.

Notice that each $\mu^{(l)}_{k}$ is absolutely continuous with respect to $\mu^{(l)}$ with the Radon-Nikodym derivative upper-bounded by $2^{(k+1)l} / (2^l-1)$ on $\Hilb^{(l)}_{k}$. Thus, Lemma~\ref{lem:measure_{t}ransform} implies that $\| \cdot \|_{\Hilb^{(l+1)}}^2 \leq 2^{(k+1)l} \| \cdot \|_{\Hilb^{(l+1)}_{k}}^2  / (2^l-1)$. Hence, it holds for each $l \in [L-2]$ that 
\begin{equation*}
    \begin{split}
        \| \mu^{(l+1)} \|_{\Hilb^{(l+1)}, 2}^2 =&~ \int \| h \|_{\Hilb^{(l+1)}}^2 \mu^{(l+1)}(dh) \\
        =&~ (2^{l+1} - 1) \sum_{k=0}^\infty 2^{-(k+1)(l+1)} \int \| h \|_{\Hilb^{(l+1)}}^2 \mu^{(l+1)}_{k}(dh) \\
        \leq &~ \frac{2^{l+1} - 1}{2^l - 1} \sum_{k=0}^\infty 2^{-(k+1)} \| \mu^{(l+1)}_{k} \|_{\Hilb^{(l+1)}_{k}}^2 \\
        \leq &~ 3~.
    \end{split}
\end{equation*}
Meanwhile, 
\begin{equation*}
    \begin{split}
        \| \mu^{(1)} \|_{\Hilb^{(1)}, 2}^2 =&~ \int \| h \|_{\Hilb^{(1)}}^2 \mu^{(1)}(dh) \\
        =&~ \sum_{k=0}^\infty 2^{-(k+1)} \int \| h \|_{\Hilb^{(1)}}^2 \mu^{(1)}_{k}(dh) \\
        =&~ 1~.
    \end{split}
\end{equation*}
Therefore, $\forall l \in [L-1]$, it holds that $\| \mu^{(l)} \|_{\Hilb^{(l)}, 2} < \infty$. Meanwhile, $\forall k \in \Nbb$, Lemma~\ref{lem:measure_{t}ransform} implies that $\| \cdot \|_{\Hilb^{(L)}}^2 \leq 2^{(k+1)(L-1)} \| \cdot \|_{\Hilb^{(L)}_{k}}^2  / (2^{L-1}-1)$, and hence
\begin{equation*}
    \begin{split}
        \| f_\infty \|_{\Hilb^{(L)}} =&~ \left \| \sum_{k=0}^\infty g_{k} \right \|_{\Hilb^{(L)}} \\
        \leq &~ \sum_{k=0}^\infty \| g_{k} \|_{\Hilb^{(L)}} \\
        \leq &~ \left ( 2^{(L-1)/2} \CompLp{L}{2}(f_{n_{1}}) + 2^{(k+1)(L-1)/2} \sum_{k=1}^\infty \CompLp{L}{2}( g_{k} ) \right ) / \sqrt{2^{L-1}-1} \\
        \leq &~ \left ( 2^{(L-1)/2} \CompLp{L}{2}(f_{n_{1}}) + \sum_{k=1}^\infty 2^{-(k+1)} \right ) / \sqrt{2^{L-1}-1} \\
        <&~ \infty~.
    \end{split}
\end{equation*}
This finishes the proof that $f_\infty \in \Ladder{L}_{2}$.

\subsubsection{Proof of Lemma~\ref{lem:Htilde}}
\label{app:pf_lem_Htilde}
Without loss of generality, we assume that $\mu(\{0\}) = 0$ (since otherwise we can replace $\mu$ with an $\mu' \in \Pcal(\Hilb)$ such that $\mu'(\{0\}) = 0$, $\| \mu' \|_{\Hilb, 2} \leq 1$ and $\| \cdot \|_{\Hilb_{\mu'}} \leq \| \cdot \|_{\Hilb_{\mu}})$. Let $\Hb$ be any $\Hilb$-valued random variable with law $\mu$. Note that we can define a bijection between $\Rbb_+ \times \hat{\Hilb}$ and $\Hilb \setminus \{0\}$ via the map $(c, \hat{h}) \mapsto c \hat{h}$, and we let $(\Cb, \hat\Hb)$ denote the image of $\Hb$ under the inverse of this map, which is a pair of random variables supported on $\Rbb_+ \times \hat{\Hilb}$. We first see that $\EEEs{\Cb^2} = \EEEs{\| \Hb \|_{\Hilb}^2} = 1$.

We choose a $\hat{\Hilb}$-valued random variable, $\Tilde{\Hb}$, whose law has a Radon-Nikodym derivative of $\EEEs{\Cb^2 | \hat \Hb }$ with respect to the law of $\hat{\Hb}$, i.e., $\forall \hat h \in \hat \Hilb$, $[\Law(\Tilde{\Hb})](d \hat h) = \EEEs{\Cb^2 | \hat \Hb = \hat{h}} [\Law(\hat \Hb)](d \hat h)$. We can verify that $\Law(\Tilde{\Hb})$ defined as such is indeed a probability measure on $\hat \Hilb$ since $\EEEs{\Cb^2 | \hat \Hb = \hat{h}} \geq 0$, and moreover,
\begin{equation*}
    \begin{split}
        [\Law(\Tilde{\Hb})](\hat \Hilb) = \int_{\hat \Hb} \EEElr{\Cb^2 | \hat \Hb = \hat{h}} [\Law(\hat \Hb)](d \hat h) 
 = \EEElr{ \EEElr{\Cb^2 | \hat \Hb} } = \EEElr{\Cb^2  } = 1~.
    \end{split}
\end{equation*}
Consider any function $f \in \Hilb_{\mu}$. By Lemma~\ref{lem:equiv} and the bijection between $\Rbb_+ \times \hat{\Hilb}$ and $\Hilb \setminus \{0\}$, there exists a function $\xi : \Rbb_+ \times \hat{\Hilb} \to \Rbb$ such that
\begin{align*}
    f(\xb) =&~ \EEElr{\xi(\Cb, \hat \Hb) \sigbig{\Hb(\xb)}} = \EEElr{\Cb \xi(\Cb, \hat \Hb) \sigbig{\hat \Hb(\xb)}}~,
\end{align*}
and $\EEElr{| \xi(\Cb, \hat \Hb) |^2} = \| f \|_{\Hilb_{\mu}}^2$. Then, defining a function $\Tilde{\xi}: \hat \Hilb \to \Rbb$ by $\forall \hat h \in \hat \Hilb$,
\begin{equation*}
    \Tilde{\xi}(\hat h) \coloneqq \frac{\EEElr{\Cb \xi(\Cb, \hat \Hb) \big | \hat \Hb = \hat h}}{\EEElr{ \Cb^2 \big | \hat \Hb = \hat h}}~,
\end{equation*}
we see that
\begin{equation*}
    \begin{split}
        f(\xb) =&~ \EEElr{ \EEElr{\Cb \xi(\Cb, \hat \Hb) \sigbig{\hat \Hb(\xb)} \big | \hat \Hb} }\\
        =&~ \EEElr{ \frac{\EEElr{\Cb \xi(\Cb, \hat \Hb) \big | \hat \Hb }}{\EEElr{ \Cb^2 \big | \hat \Hb}} \sigbig{\hat \Hb(\xb)} \EEElr{ \Cb^2 \big | \hat \Hb}} \\
        =&~ \int_{\hat \Hilb} \Tilde{\xi}(\hat h) \sigbig{\hat h(\xb)} \EEElr{ \Cb^2 \big | \hat \Hb = \hat h} [\Law(\hat \Hb)](d \hat h)  \\
        =&~ \int_{\hat \Hilb} \Tilde{\xi}(\hat h) \sigbig{\hat h(\xb)}  [\Law(\tilde \Hb)](d \hat h) \\
        =&~ \EEElr{ \Tilde{\xi}(\Tilde{\Hb}) \sigbig{\Tilde{\Hb}(\xb)} }~.
    \end{split}
\end{equation*}
Thus, there is
\begin{equation*}
    \begin{split}
        \| f \|_{\Hilb_{\Law(\Tilde{\Hb})}}^2 \leq&~ \EEElr{ |\Tilde{\xi}(\Tilde{\Hb})|^2 } \\
        =&~ \int_{\hat \Hilb} \left | \frac{\EEElr{\Cb \xi(\Cb, \hat \Hb) | \hat \Hb = \hat h}}{\EEElr{ \Cb^2 \big | \hat \Hb = \hat h}} \right |^2 [\Law(\tilde \Hb)](d \hat h) \\
        =&~ \int_{\hat \Hilb} \left | \frac{\EEElr{\Cb \xi(\Cb, \hat \Hb) | \hat \Hb = \hat h}}{\EEElr{ \Cb^2 \big | \hat \Hb = \hat h}} \right |^2 \EEEs{\Cb^2 \big | \hat \Hb = \hat{h}} [\Law(\hat \Hb)](d \hat h) \\
        =&~ \int_{\hat \Hilb} \frac{\left | \EEElr{ \Cb \xi(\Cb, \hat \Hb) \big | \hat \Hb = \hat h} \right |^2}{\EEElr{ \Cb^2 \big | \hat \Hb = \hat h}} [\Law(\hat \Hb)](d \hat h) \\
        =&~ \EEElr{\frac{\left | \EEElr{ \Cb \xi(\Cb, \hat \Hb) \big | \hat \Hb} \right |^2}{\EEElr{ \Cb^2 \big | \hat \Hb}}}
    \end{split}
\end{equation*}
By the Cauchy-Schwartz inequality, $\left | \EEElr{ \Cb \xi(\Cb, \hat \Hb) \big | \hat \Hb} \right |^2 \leq \EEElr{ \Cb^2 | \hat \Hb} \EEElr{ |\xi(\Cb, \hat \Hb)|^2 | \hat \Hb}$. Hence, 
\begin{equation*}
    \| f \|_{\Hilb_{\Law(\Tilde{\Hb})}}^2 \leq \EEElr{\EEElr{ |\xi(\Cb, \hat \Hb)|^2 \big | \hat \Hb}} = \EEElr{ |\xi(\Cb, \hat \Hb)|^2} = \| f \|_{\Hilb_{\mu}}^2~.
\end{equation*}

\subsection{Proof of Proposition~\ref{prop:coupled}}
\label{app:pf_entang_equiv}
For each $l \in [L-1]$, let $\Hb^{(l)}$ be a $\Hilb^{(l)}$-valued random variable with law $\mu^{(l)}$, and let $\Hb^{(1)}, ..., \Hb^{(L-1)}$ be distributed independently on a common probability space. 

First, we define $\Ab$ as follows. By Lemma~\ref{lem:equiv}, there exists $\xi \in L^2(\Hilb^{(L-1)}, \mu^{(L-1)})$ such that 
$f(\xb) = \int \xi(h) \sigbig{h(\xb)} \mu^{(L-1)}(dh)$
and 
$\| f \|_{\Hilb^{(L)}} = \| \xi \|_{L^2(\Hilb^{(L-1)}, \mu^{(L-1)})}$.
Then, we define $\Ab \coloneqq \xi(\Hb^{(L-1)})$, which is measurable with respect to $\Hb^{(L-1)}$. Hence, \eqref{eq:f_entang} as well as the equality $\| f \|_{\Hilb^{(L)}}^2 = \EEEb{\Ab^2}$ are implied.

Next, $\forall l \in [L-2]$, we define $\Xib^{(l)}$ as follows. By Lemma~\ref{lem:equiv}, $\forall h \in \Hilb^{(l+1)}$, $\exists \xi_h \in L^2(\Hilb^{(l)}, \mu^{(l)})$ such that 
\begin{equation}
\label{eq:h_entang}
    h(\xb) = \int \xi_h(h') \sigbig{h'(\xb)} \mu^{(l)}(dh')~,
\end{equation}
and
\begin{equation}
    \label{eq:h_norm_entang}
    \| h \|_{\Hilb^{(l+1)}} = \| \xi_h \|_{L^2(\Hilb^{(l)}, \mu^{(l)})}~.
\end{equation}
We denote the map $h \mapsto \xi_h$ by $\Xi^{(l)}$, i.e., $[\Xi^{(l)}(h)](h') \coloneqq \xi_h(h')$ for $h \in \Hilb^{(l+1)}$ and $h' \in \Hilb^{(l)}$, and finally define $\Xib^{(l)} \coloneqq [\Xi^{(l)}(\Hb^{(l+1)})](\Hb^{(l)})$, which is, by definition, measurable with respect to the sigma-algebra generated by $\Hb^{(l)}$ and $\Hb^{(l+1)}$. Then, \eqref{eq:H_entang} and the relation $\| \Hb^{(l+1)} \|_{\Hilb^{(l+1)}}^2 = \EEEb{(\Xib^{(l)})^2 \big | \Hb^{(l+1)}}$ are implied by \eqref{eq:h_entang} and \eqref{eq:h_norm_entang}.

\subsection{Proof of Lemma~\ref{lem:equiv}}
\label{app:pf_lem_equiv}
We will take advantage of the following general result on integral transforms and RKHS \citep{saitoh1997integral, saitoh1999integral}:
\begin{lemma}[Saitoh]
\label{lem:saitoh}
Given a Hilbert space $\Hilb_{0}$ and a map $\varphi: \domX \to \Hilb_{0}$, we define the kernel function $\kappa: \domX \times \domX \to \Rbb$ by $\kappa(\xb, \xb') = \langle \varphi(\xb), \varphi(\xb') \rangle_{\Hilb_{0}}$, and let $\Hilb$ denote the RKHS associated with $\kappa$. Then, a function $f$ on $\domX$ belongs to $\Hilb$ if and only if $\exists \xi \in \Hilb_{0}$ such that 
\begin{equation}
\label{eq:equiv_saitoh}
    f(\xb) = \langle \xi, \varphi(\xb) \rangle_{\Hilb_{0}}~,~ \quad \forall \xb \in \domX~,
\end{equation}and moreover, 
\begin{equation*}
    \| f \|_{\Hilb} = \inf \big \{ \| \xi \|_{\Hilb_{0}}: \xi \in \Hilb_{0} \text{ such that }  \eqref{eq:equiv_saitoh} \text{ holds} \big \}~,
\end{equation*}
with the infimum achieved at a unique $\xi^{*} \in \Hilb_{0}$.
\end{lemma}
\noindent For our purpose, we would like to apply Lemma~\ref{lem:saitoh} with $\Hilb_{0} = L^2(\Conti, \mu)$ and $\varphi: \domX \to L^2(\Conti, \mu)$ defined such that $\varphi(\xb)$ is the function on $\Conti$ defined by $h \mapsto \sigma(h(\xb))$. Indeed, $L^2(\Conti, \mu)$ is equipped with the inner product $\langle \xi_{1}, \xi_{2} \rangle_{L^2(\Conti, \mu)} \coloneqq \int \xi_{1}(h) \xi_{2}(h) \mu(dh)$ and complete (\citealt{brezis2011functional}, Theorem 4.8), and hence it is a Hilbert space. Moreover, $\forall \xb \in \domX$, as $\sigma$ is Lipschitz, there exists $J_{\sigma} > 0$ such that $| \sigma(h(\xb)) | \leq |\sigma(0)| + J_{\sigma} |h(\xb)| \leq |\sigma(0)| + J_{\sigma} \| h \|_\infty$, and hence $\varphi(\xb) \in L^2(\Conti, \mu)$ by the assumption on $\mu$ (note that $\varphi(\xb)$ is a measurable function on $\Conti$ since the evaluation functionals are measurable with respect to the Borel sigma-algebra on $\Conti$). Thus, we are able to prove Lemma~\ref{lem:equiv} by applying Lemma~\ref{lem:saitoh} with $\Hilb_{0}$ and $\varphi$ defined as above.

\section{Supplementary Materials for Section~\ref{sec:real_approx}}
\subsection{Including the Bias Terms}
\label{app:bias}
Below, we provide a way to include the bias term into the framework proposed in the main text.  We note that alternative ways to do so are also possible, such as by generalizing the homogeneous Barron space construction proposed by \citet{park2023minimum}. 
\subsubsection{Multi-layer NN}
By including the bias term, we mean replacing \eqref{eq:h_i} in the definition of the multi-layer NN by
\begin{equation*}
\label{eq:h_i_bias}
    h_i^{(l+1)}(\xb) \coloneqq b^{(l+1)}_i + \frac{1}{m} \summ{j}{m} W_{i, j}^{(l)} \sigbig{h_j^{(l)}(\xb)}~.
\end{equation*}
In the GF dynamics, the bias terms evolve according to the following ODE:
\begin{equation*}
    \ddt b^{(l)}_{i, t} = - \beta \expenu{\xb}{\zeta_{m, t} q^{(l)}_{i, t}(\xb)}~,
\end{equation*}
where $\beta$ denotes the learning rate of the bias parameters relative to the weight parameters. If $\beta = 0$, for example, it corresponds to having untrained bias terms.
\subsubsection{NHL}
\label{app:mf_bias}
To incorporate the bias term, we replace Definition~\ref{def:nhl} of the NHL in the following way. We consider the map
\begin{equation}
\label{eq:hb_map}
    \begin{split}
       \Conti \times \Rbb \to&~ \Conti \\
        (h, b) \mapsto&~ h(\cdot) + b~,
    \end{split}
\end{equation}
If $\mu \in \Pcal(\Con \times \Rbb)$, we let $\mu_+ \in \Pcal(\Con)$ denote its push-forward under this map.
\begin{definition}
\label{def:nhl_bias}
Suppose each of $\Hilb^{(2)}$, ..., $\Hilb^{(L)}$ is an RKHS on $\domX$, 
and $\forall l \in [L-1]$, there exists $\mu^{(l)} \in \Pcal(\Hilb^{(l)} \times \Rbb)$ such that $\Hilb^{(l+1)}$ is the RKHS associated with the kernel function
\begin{equation*}
    \kappa^{(l)}(\xb, \xb') \coloneqq \int \sigbig{h(\xb) + b} \sigbig{h(\xb') + b} \mu(dh, db)~.
\end{equation*}
In other words, we can write 
$\Hilb^{(l+1)} = \Hilb_{\mu^{(l)}_+}$.
Then, we say that $(\Hilb^{(l)})_{l \in [L]}$ is an \emph{$L$-level NHL} (with the bias terms included) induced by the sequence of probability measures,
$(\mu^{(l)})_{l \in [L-1]}$; in addition, a function $f$ on $\domX$ belongs to the NHL
if $f \in \Hilb^{(L)}$.
\end{definition}
\noindent If $\Hilb$ is a Hilbert space and $\mu \in \Pcal(\Hilb \times \Rbb)$, we write $\| \mu \|_{\Hilb, p, +} \coloneqq ( \int \| h \|_{\Hilb}^p + |b|^p \mu
(dh, db))^{1/p}$ for $p \in (0, \infty)$, and analogously for $p = \infty$. Given an RKHS $\Hilb$, we then define
\begin{equation*}
    \mathscr{D}_p \left (\Hilb, \Hilb' \right ) \coloneqq \inf_{\mu \in \Pcal(\Hilb \times \Rbb),~ \Hilb_{\mu} = \Hilb'} \| \mu \|_{\Hilb, p, +}~.
\end{equation*}
Next, we can define the $(L, p)$-NHL complexity of a function similarly as through \eqref{eq:DLp} and \eqref{eq:ladder_norm}.

In addition, in place of Proposition~\ref{prop:coupled}, the representation of an NHL through coupled random fields can be defined in the following way.
\begin{proposition}
Given an $L$-level NHL as in Definition~\ref{def:nhl_bias}, there exist $L-1$ random fields on $\domX$, $\Hb^{(1)}$, ..., $\Hb^{(L-1)}$, as well as $2L-2$ scalar random variables, $\Bb^{(1)}$, ..., $\Bb^{(L-1)}$, $\Xib^{(1)}$, ..., $\Xib^{(L-2)}$, and $\Ab$, which are defined on a common probability space and satisfy the following properties:
\begin{itemize}
\item The pairs $(\Hb^{(1)}, \Bb^{(1)})$, ..., $(\Hb^{(L-1)}, \Bb^{(L-1)})$ are mutually independent, and $\forall l \in [L-1]$, $\mu^{(l)} = \Law(\Hb^{(l)}, \Bb^{(l)})$;
    \item $\forall l \in [L-2]$, $\Xib^{(l)}$ is measurable with respect to the sigma-algebra generated by $\Hb^{(l)}$, $\Hb^{(l+1)}$, $\Bb^{(l)}$, and $\Bb^{(l+1)}$. Moreover,
    \begin{equation}
\label{eq:H_entang_bias}
        \Hb^{(l+1)}(\xb) = \EEEb{\Xib^{(l)} \sigbig{\Hb^{(l)}(\xb) + \Bb^{(l)}} \big | \Hb^{(l+1)}}~,
    \end{equation}
    where $\EEEb{ \cdot \big | \cdot }$ denotes the conditional expectation, and $\| \Hb^{(l+1)} \|_{\Hilb^{(l+1)}}^2 = \EEEb{(\Xib^{(l)})^2 \big | \Hb^{(l+1)}}$;
    \item $\Ab$ is measurable with respect to the sigma-algebra generated by $\Hb^{(L-1)}$ and $\Bb^{(L-1)}$. Moreover,
\begin{equation}
\label{eq:f_entang_bias}
    f(\xb) = \EEEb{\Ab \sigbig{\Hb^{(L-1)}(\xb) + \Bb^{(L-1)}}}~,
\end{equation}
and $\| f \|_{\Hilb^{(L)}}^2 = \EEEb{\Ab^2}$.
\end{itemize}
\end{proposition}
Furthermore, in the mean-field dynamics, the evolution of the bias term is governed by
\[
\ddt \Bb^{(l)}_{t} = - \beta \expenu{\xb}{\zeta_{t}(\xb) \Qtl{t}{l}(\xb) \sigpbig{\Htl{t}{l}(\xb) + \Btl{t}{l}}}
\]

\subsection{Proof of Theorem~\ref{prop:in_ladder}}
\label{app:pf_prop_in_ladder}
First, by the definition of $\Hilb^{(1)}_{m}$, there is $\| \mu^{(1)}_{m} \|_{\Hilbl{1}_{m}, p} = M^{(1)}_{m, p}$. For $l \in [L-2]$, Lemma~\ref{lem:equiv} implies $\| h_i^{(l+1)} \|_{\Hilbl{l+1}_{m}}^2 \leq \frac{1}{m} \summ{j}{m} |W^{(l)}_{i, j}|^2$, and so $\| \mu^{(l+1)}_{m} \|_{\Hilbl{l+1}_{m}, p}^p = \frac{1}{m} \summ{i}{m} \| h_i^{(l+1)} \|_{\Hilbl{l+1}_{m}}^p \leq (M^{(l+1)}_{m, p})^p$. Finally, Lemma~\ref{lem:equiv} implies $\| f \|_{\Hilbl{L}_{m}, p} \leq M^{(L)}_{m, p}$. Together, they prove the theorem.

\subsection{Proof of Theorem~\ref{thm:approx}}
\label{app:thm_approx_pf}
Let $f$ be a function on $\domX$ with $\CompLp{L}{\infty}(f) < \infty$. By definition, there exist probability measures $\mu^{(1)}$, ..., $\mu^{(L-1)}$ and deterministic functions $\Xi^{(1)}$, ..., $\Xi^{(L-1)}$ satisfying the conditions specified in Appendix~\ref{app:pf_entang_equiv}. Our strategy will be to consider a random approximation of $f$ using an $L$-layer NN with width $m$ that achieves a low approximation error in expectation.

$\forall l \in [L-1]$, we let $\{ \Hb^{(l)}_i \}_{i \in [m]}$ be $m$ independent samples from $\mu^{(l)}$ in $\Hilb^{(l)}$. For $l=1$, we let $\bar{\Hb}^{(1)}_i \coloneqq \Hb^{(1)}_i$. For $l \in [L-2]$, writing $\bar{\Wb}^{(l)}_{i,j} \coloneqq \Xi^{(l)}(\Hb^{(l+1)}_i, \Hb^{(l)}_j)$, we iteratively define
\begin{equation*}
    \bar{\Hb}^{(l+1)}_i(\xb) \coloneqq \frac{1}{m} \summ{j}{m} \bar{\Wb}^{(l)}_{i,j} \sigbig{\bar{\Hb}^{(l)}_j(\xb)}~,
\end{equation*}
and finally, writing $\bar{\Ab}_i \coloneqq \Xi^{(L-1)}(\Hb^{(L-1)}_i)$, we define
\begin{equation*}
    \Fb_{m}(\xb) \coloneqq \frac{1}{m} \summ{i}{m} \bar{\Ab} \sigbig{\bar{\Hb}^{(L-1)}_i(\xb)}~,
\end{equation*}
and will bound its difference with $f$ in expectation. We will use the two following lemmas:

\begin{lemma}
\label{lem:approx_ind}
    $\forall l \in [L-1]$, $\forall i \in [m]$, almost surely, 
    \begin{equation}
    \label{eq:inductive_approx}
        \expenu{\xb \sim \nu}{\EEElr{ \big (\bar{\Hb}^{(l)}_i(\xb) - \Hb^{(l)}_i(\xb) \big )^2 \big | \Hb^{(l)}_i } } \leq  \frac{(l-1)^2}{m} (J_{\sigma})^{2l-2} M_{\nu} \prod_{l'=1}^{l} \| \mu^{(l')} \|_{\Hilb^{(l')}, \infty}^2~.
    \end{equation}
\end{lemma}

\begin{lemma}
\label{lem:square_sum}
    $\forall a, b \in \Rbb$, $\forall \lambda > 0$, there is $(a + b)^2 \leq (1 + 1/\lambda) a^2 + (1 + \lambda) b^2$.
\end{lemma}

\noindent Lemma~\ref{lem:approx_ind} is proved in Appendix~\ref{app:pf_lem_approx_ind}. Lemma~\ref{lem:square_sum} is easy to see since $2ab = 2(a/\sqrt{\lambda})(b \sqrt{\lambda}) \leq a^2 / \lambda + \lambda b^2$. Compared to the more naive bound $(a + b)^2 \leq 2 (a^2 + b^2)$, this lemma allows us to obtain a sharper bound when $a$ and $b$ are imbalanced in magnitude by adjusting $\lambda$.

$\forall \xb \in \domX$, we can write
\begin{equation*}
    \begin{split}
        \Fb_{m}(\xb) - f(\xb) = \bigg ( \Fb_{m}(\xb) - \frac{1}{m} \summ{i}{m} \bar{\Ab}_i \sigbig{\Hb^{(L-1)}_i(\xb)} \bigg ) + \bigg ( \frac{1}{m} \summ{j}{m} \bar{\Ab}_{i} \sigbig{\Hb^{(L-1)}_i(\xb)} - f(\xb) \bigg )
    \end{split}
\end{equation*}
Thus, using Lemma~\ref{lem:square_sum}, we see that, $\forall \xb \in \domX$, $\forall \lambda > 0$,
\begin{equation*}
    \begin{split}
        \expenu{\xb \sim \nu}{\EEElr{| \Fb_{m}(\xb) - f(\xb) |^2}} \leq (1 + 1/\lambda) \cdot \expenu{\xb \sim \nu}{(\text{I})} + (1 + \lambda) \cdot \expenu{\xb \sim \nu}{(\text{II})}~,
    \end{split}
\end{equation*}
where we define
\begin{align*}
    (\text{I}) \coloneqq &~ \EEElr{ \Big ( \Fb_{m}(\xb) - \frac{1}{m} \summ{i}{m} \bar{\Ab}_i \sigbig{\Hb^{(L-1)}_i(\xb)} \Big )^2}~, \\
     (\text{II}) \coloneqq &~ \EEElr{ \bigg ( \frac{1}{m} \summ{j}{m} \bar{\Ab}_{i} \sigbig{\Hb^{(L-1)}_i(\xb)} - f(\xb) \bigg )^2 }~,
\end{align*}
and we will bound these two terms separately. For $(\text{I})$, there is
 
\begin{equation*}
    \begin{split}
        (\text{I}) 
         =&~ \EEEBB{ \bigg ( \frac{1}{m} \summ{i}{m} \bar{\Ab}_i \big ( \sigbig{\bar{\Hb}^{(L-1)}_i(\xb)} - \sigbig{\Hb^{(L-1)}_i(\xb)} \big ) \bigg )^2} \\
         \leq &~ \EEElr{ \bigg ( \frac{1}{m} \summ{i}{m} (\bar{\Ab}_i)^2 \bigg ) \bigg ( \frac{1}{m} \summ{i}{m} \Big ( \sigbig{\bar{\Hb}^{(L-1)}_i(\xb)} - \sigbig{\Hb^{(L-1)}_i(\xb)} \Big )^2 \bigg)} \\
         =&~ \EEEBB{ \EEEbb{ \bigg ( \frac{1}{m} \summ{i}{m} (\bar{\Ab}_i)^2 \bigg ) \bigg ( \frac{1}{m} \summ{i}{m} \Big ( \sigbig{\bar{\Hb}^{(L-1)}_i(\xb)} - \sigbig{\Hb^{(L-1)}_i(\xb)} \Big )^2 \bigg) \Big | \{ \Hb^{(L-1)}_{i'} \}_{i' \in [m]} } }  \\
         \leq &~ (J_{\sigma})^2 \EEElr{ \bigg ( \frac{1}{m} \summ{i}{m} (\bar{\Ab}_i)^2 \bigg ) \bigg ( \frac{1}{m} \summ{i}{m} \EEElr{ \big ( \bar{\Hb}^{(L-1)}_i(\xb) - \Hb^{(L-1)}_i(\xb) \big )^2 \big |  \Hb^{(L-1)}_i} \bigg)}
    \end{split}
\end{equation*}
where on the second line, we use the Cauchy-Schwartz inequality; on the third line, we use the tower property of conditional expectation; and on the fourth line, we use that 1) each $\bar{\Ab}_i$ is measurable with respect to $\Hb^{(L-1)}_i$, 2) each $\bar{\Hb}^{(L-1)}_i$ is independent from $\Hb^{(L-1)}_{i'}$ when $i \neq i'$, and 3) $\sigma$ is $J_{\sigma}$-Lipschitz. 
Then, using Lemma~\ref{lem:approx_ind}, we derive that

\begin{equation}
\label{eq:approx_I}
    \begin{split}
       &~ \expenu{\xb \sim \nu}{(\text{I})} \\
       \leq&~ (J_{\sigma})^2 \EEEBB{ \bigg ( \frac{1}{m} \summ{i}{m} (\bar{\Ab}_i)^2 \bigg ) \bigg ( \frac{1}{m} \summ{i}{m} \expenu{\xb \sim \nu}{\EEElr{ \big ( \bar{\Hb}^{(L-1)}_i(\xb) - \Hb^{(L-1)}_i(\xb) \big )^2 \big |  \Hb^{(L-1)}_i}} \bigg)} \\
        \leq & \frac{(L-2)^2}{m} (J_{\sigma})^{2L-2} M_{\nu} \EEElr{ \frac{1}{m} \summ{i}{m} \big ( \Xi^{(L-1)} \big (\Hb^{(L-1)}_i \big ) \big )^2}  \left ( \prod_{l=1}^{L-1} \| \mu^{(l)} \|_{\Hilb^{(l)}, \infty}^2 \right ) \\
         \leq & \frac{(L-2)^2}{m} (J_{\sigma})^{2L-2} M_{\nu}  \| f \|_{\Hilb^{(L)}}^2 \prod_{l=1}^{L-1} \| \mu^{(l)} \|_{\Hilb^{(l)}, \infty}^2~.
    \end{split}
\end{equation}
 
For $(\text{II})$, there is
\begin{equation*}
    \begin{split}
        (\text{II}) \coloneqq &~ \EEEBB{ \bigg ( \frac{1}{m} \summ{j}{m} \bar{\Ab}_{i} \sigbig{\Hb^{(L-1)}_i(\xb)} - f(\xb) \bigg )^2 } \\
        =&~ \frac{1}{m} \summ{i}{m} \EEElr{ \left ( \Xi^{(L-1)}(\Hb^{(L-1)}_i) \sigbig{\Hb^{(L-1)}_i(\xb)} - \EEElr{\Xi^{(L-1)}(\Hb_i^{(L-1)}) \sigbig{\Hb^{(L-1)}_i(\xb)}} \right )^2} \\
        \leq & ~ \frac{1}{m} \EEElr{ \left ( \Xi^{(L-1)}(\Hb^{(L-1)}) \right )^2 \left ( \sigbig{\Hb^{(L-1)}(\xb)} \right )^2 }~,
    \end{split}
\end{equation*}
where we use the independence among $\{ \Hb^{(L-1)}_i \}_{i \in [m]}$ and their equivalence in law to $\Hb^{(L-1)}$.  By Theorem~\ref{prop:nhl_prop}(\ref{prop:infty_norm}), $\forall \xb \in \domX$, there is 
\begin{equation*}
    |\Hb^{(L-1)}(\xb)| \leq (J_{\sigma})^{L-2} \| \xb \| \CompLp{L-1}{2}(\Hb^{(l)}_i) \leq (J_{\sigma})^{L-2} \| \xb \| \| \Hb^{(L-1)} \|_{\Hilb^{(L-1)}} \prod_{l'=1}^{L-2} \| \mu^{(l')} \|_{\Hilb^{(l')}, 2}~,
\end{equation*}
and hence, by the assumptions on $\sigma$, it holds that 
\begin{equation*}
    \big ( \sigbig{\Hb^{(L-1)}(\xb)} \big )^2 \leq (J_{\sigma})^{2} \big ( \Hb^{(L-1)}(\xb) \big )^2 \leq (J_{\sigma})^{2L-2} \| \xb \|^2 \| \Hb^{(L-1)} \|_{\Hilb^{(L-1)}}^2 \prod_{l'=1}^{L-2} \| \mu^{(l')} \|_{\Hilb^{(l')}, 2}^2~,
\end{equation*}
which is almost surely bounded by $(J_{\sigma})^{2L-2} \| \xb \|^2 \prod_{l'=1}^{L-1} \| \mu^{(l')} \|_{\Hilb^{(l')}, \infty}^2$. Thus,
\begin{equation}
\label{eq:approx_II}
    \begin{split}
        \expenu{\xb \sim \nu}{(\text{II})} \leq & ~ \frac{1}{m} (J_{\sigma})^{2L-2} M_{\nu} \left ( \prod_{l'=1}^{L-1} \| \mu^{(l')} \|_{\Hilb^{(l')}, \infty}^2 \right ) \EEElr{ \left ( \Xi^{(L-1)}(\Hb^{(L-1)}) \right )^2 } \\
        \leq &~ \frac{1}{m} (J_{\sigma})^{2L-2} M_{\nu}  \| f \|_{\Hilb^{(L)}}^2 \prod_{l'=1}^{L-1} \| \mu^{(l')} \|_{\Hilb^{(l')}, \infty}^2~.
    \end{split}
\end{equation}
Together, combining \eqref{eq:approx_I} and \eqref{eq:approx_II} with Lemma~\ref{lem:square_sum} and choosing $\lambda = L-2$, we derive that, 
\begin{equation*}
\begin{split}
    &~ \expenu{\xb \sim \nu}{\EEElr{| \Fb_{m}(\xb) - f(\xb) |^2}} \\
    \leq &~ \big ( (1 + 1 / \lambda) (L-2)^2 + (1 + \lambda) \big ) (J_{\sigma})^{2L-2} M_{\nu} \| f \|_{\Hilb^{(L)}}^2 \prod_{l'=1}^{L-1} \| \mu^{(l')} \|_{\Hilb^{(l')}, \infty}^2 \\
    =&~  \frac{(L-1)^2}{m} (J_{\sigma})^{2L-2} M_{\nu} \| f \|_{\Hilb^{(L)}}^2 \prod_{l'=1}^{L-1} \| \mu^{(l')} \|_{\Hilb^{(l')}, \infty}^2 \\
    =&~  \frac{(L-1)^2}{m}(J_{\sigma})^{2L-2} M_{\nu} \CompLp{L}{\infty}(f)^2~.
\end{split}
\end{equation*}
Interchanging the expectations, we get
\begin{equation*}
\begin{split}
    \EEElr{\expenu{\xb \sim \nu}{| \Fb_{m}(\xb) - f(\xb) |^2}} =&~ \expenu{\xb \sim \nu}{\EEElr{| \Fb_{m}(\xb) - f(\xb) |^2}} \\
    \leq &~  \frac{(L-1)^2}{m} (J_{\sigma})^{2L-2} M_{\nu} \CompLp{L}{\infty}(f)^2~.
\end{split}
\end{equation*}
Thus, as a consequence of Markov's inequality, there exists a realization of $(\Hb^{(l)}_i)_{l \in [L-1], i \in [m]}$ under which
\begin{equation*}
    \expenu{\xb \sim \nu}{| \Fb_{m}(\xb) - f(\xb) |^2} \leq  \frac{(L-1)^2}{m} (J_{\sigma})^{2L-2} M_{\nu} \CompLp{L}{\infty}(f)^2~.
\end{equation*}

\subsubsection{Proof of Lemma~\ref{lem:approx_ind}}
\label{app:pf_lem_approx_ind}
For $l = 1$, there is $\bar{\Hb}^{(1)}_i = \Hb^{(1)}_i$, and hence $\bar{\Hb}^{(1)}_i(\xb) - \Hb^{(1)}_i(\xb) = 0$, $\forall \xb \in \domX$, thus validating \eqref{eq:inductive_approx} when $l = 1$. Next, we suppose that the statement holds for some $l \in [L-2]$ and prove that it holds for $l+1$ as well. 

For level $l+1$, $\forall i \in [m]$ and $\xb \in \domX$, we can write
\begin{equation*}
\begin{split}
    \bar{\Hb}^{(l+1)}_i(\xb) - \Hb^{(l+1)}_i(\xb) =&~ \Big( \bar{\Hb}^{(l+1)}_i(\xb) - \frac{1}{m} \summ{j}{m} \bar{\Wb}^{(l)}_{i,j} \sigbig{\Hb^{(l)}_j(\xb)} \Big ) \\
    +&~ \bigg ( \frac{1}{m} \summ{j}{m} \bar{\Wb}^{(l)}_{i,j} \sigbig{\Hb^{(l)}_j(\xb)} - \Hb^{(l+1)}_i(\xb) \bigg ) 
\end{split}
\end{equation*}
Hence, by Lemma~\ref{lem:square_sum}, it holds for any $\lambda > 0$ that
\begin{equation*}
    \begin{split}
        &~ \expenu{\xb \sim \nu}{\EEElr{ \big ( \bar{\Hb}^{(l+1)}_i(\xb) - \Hb^{(l+1)}_i(\xb) \big )^2 \Big | \Hb^{(l+1)}_i }} \\
        \leq &~ (1 + 1/\lambda) \cdot \expenu{\xb \sim \nu}{(\text{I})} + (1 + \lambda) \cdot \expenu{\xb \sim \nu}{(\text{II})}~,
    \end{split}
\end{equation*}

where we define
\begin{align*}
(\text{I}) \coloneqq &~ \EEEBB{ \Big ( \bar{\Hb}^{(l+1)}_i(\xb) - \frac{1}{m} \summ{j}{m} \bar{\Wb}^{(l)}_{i,j} \sigbig{\Hb^{(l)}_j(\xb)} \Big )^2 \Big | \Hb^{(l+1)}_i }~, \\
 (\text{II}) \coloneqq &~ \EEEBB{ \bigg ( \frac{1}{m} \summ{j}{m} \bar{\Wb}^{(l)}_{i,j} \sigbig{\Hb^{(l)}_j(\xb)} - \Hb^{(l+1)}_i(\xb) \bigg )^2 \Big | \Hb^{(l+1)}_i }~,    
\end{align*}
and we will bound these two terms separately. For $(\text{I})$, there is

\begin{equation*}
    \begin{split}
        (\text{I})
        =&~ \EEEBB{ \bigg ( \frac{1}{m} \summ{j}{m} \bar{\Wb}^{(l)}_{i,j} \big ( \sigbig{\bar{\Hb}^{(l)}_j(\xb)} - \sigbig{\Hb^{(l)}_j(\xb)} \big ) \bigg )^2 \Big | \Hb^{(l+1)}_i }  \\
        \leq &~ \EEEBB{ \bigg ( \frac{1}{m} \summ{j}{m} \big ( \bar{\Wb}^{(l)}_{i,j} \big )^2 \bigg ) \bigg( \frac{1}{m} \summ{j}{m} \big ( \sigbig{\bar{\Hb}^{(l)}_j(\xb)} - \sigbig{\Hb^{(l)}_j(\xb)} \big )^2 \bigg ) \Big | \Hb^{(l+1)}_i } \\
        = &~ \mathbb{E} \Bigg [ \mathbb{E} \bigg [ \bigg ( \frac{1}{m} \summ{j}{m} \big ( \bar{\Wb}^{(l)}_{i,j} \big )^2 \bigg ) \bigg( \frac{1}{m} \summ{j}{m} \big ( \sigbig{\bar{\Hb}^{(l)}_j(\xb)} - \sigbig{\Hb^{(l)}_j(\xb)} \big )^2 \bigg ) \\
        & \hspace{190pt} \Big | \Hb^{(l+1)}_i, \{ \Hb^{(l)}_j \}_{j \in [m]} \bigg ] ~ \Big | \Hb^{(l+1)}_i \Bigg ] \\
        \leq &~ (J_{\sigma})^2 \EEEBB{ \bigg ( \frac{1}{m} \summ{j}{m} \big ( \bar{\Wb}^{(l)}_{i,j} \big )^2 \bigg ) \bigg ( \frac{1}{m} \summ{j}{m} \EEEbb{\big ( \bar{\Hb}^{(l)}_j(\xb) - \Hb^{(l)}_j(\xb) \big )^2 \Big | \Hb^{(l)}_j } \bigg ) \Big | \Hb^{(l+1)}_i }
    \end{split}
\end{equation*}
\noindent where on the second line, we use the Cauchy-Schwartz inequality; on the third line, we the tower property of conditional expectations; and
on the last line, we use that 1) $\bar{\Wb}^{(l)}_{i,j}$ is measurable with respect to (the sigma-algebra generated by) $\Hb^{(l+1)}_i$ and $\Hb^{(l)}_j$, 2) $\bar{\Hb}^{(l)}_j$ and $\Hb^{(l)}_j$ are independent from $\Hb^{(l+1)}_i$, 3) each $\bar{\Hb}^{(l)}_j$ is independent from $\Hb^{(l)}_{j'}$ when $j \neq j'$, and 4) $\sigma$ is $J_{\sigma}$-Lipschitz.  
Then, using the inductive hypothesis, we derive that

\begin{equation*}
    \begin{split}
        &~ \expenu{\xb \sim \nu}{(\text{I})} \\
        \leq &~ (J_{\sigma})^2 \EEEBB{ \bigg ( \frac{1}{m} \summ{j}{m} \big ( \bar{\Wb}^{(l)}_{i,j} \big )^2 \bigg ) \bigg ( \frac{1}{m} \summ{j}{m} \expenu{\xb \sim \nu}{\EEEbb{\big ( \bar{\Hb}^{(l)}_j(\xb) - \Hb^{(l)}_j(\xb) \big )^2 \Big | \Hb^{(l)}_j }} \bigg ) \Big | \Hb^{(l+1)}_i } \\
        \leq &~ \frac{(l-1)^2}{m} (J_{\sigma})^{2l} M_{\nu} \left ( \prod_{l'=1}^{l} \| \mu^{(l')} \|_{\Hilb^{(l')}, \infty}^2 \right ) \EEEBB{ \frac{1}{m} \summ{j}{m} \big ( \bar{\Wb}^{(l)}_{i,j} \big )^2 \Big | \Hb^{(l+1)}_i } \\
        \leq &~ \frac{(l-1)^2}{m} (J_{\sigma})^{2l}M_{\nu} \left ( \prod_{l'=1}^{l} \| \mu^{(l')} \|_{\Hilb^{(l')}, \infty}^2 \right ) \| \Hb^{(l+1)}_i \|_{\Hilb^{(l+1)}}^2 \\
        \leq &~ \frac{(l-1)^2}{m} (J_{\sigma})^{2l} M_{\nu} \left ( \prod_{l'=1}^{l+1} \| \mu^{(l')} \|_{\Hilb^{(l')}, \infty}^2 \right )~,
    \end{split}
\end{equation*}

For $(\text{II})$, there is
\begin{equation*}
    \begin{split}
        (\text{II}) \coloneqq &~ \EEElr{ \bigg ( \frac{1}{m} \summ{j}{m} \bar{\Wb}^{(l)}_{i,j} \sigbig{\Hb^{(l)}_j(\xb)} - \Hb^{(l+1)}_i(\xb) \bigg )^2 \Big | \Hb^{(l+1)}_i } \\
        =&~ \mathbb{E} \Bigg [ \bigg ( \frac{1}{m} \summ{j}{m} \Xi^{(l)}(\Hb^{(l+1)}_i, \Hb^{(l)}_j) \sigbig{\Hb^{(l)}_j(\xb)} \\
        & \hspace{25pt} - \EEElr{\Xi^{(l)}(\Hb^{(l+1)}_i, \Hb^{(l)}) \sigbig{\Hb^{(l)}(\xb)} \big |  \Hb^{(l+1)}_i} \bigg )^2 \Big | \Hb^{(l+1)}_i \Bigg ] \\
        \leq & ~ \frac{1}{m} \EEElr{ \left ( \Xi^{(l)}(\Hb^{(l+1)}_i, \Hb^{(l)}) \right )^2 \left ( \sigbig{\Hb^{(l)}(\xb)} \right )^2  \Big | \Hb^{(l+1)}_i }~,
    \end{split}
\end{equation*}
where we use the independence among $\{ \Hb^{(l)}_j \}_{j \in [m]}$ and their equivalence in law to $\Hb^{(l)}$.
 By Theorem~\ref{prop:nhl_prop}(\ref{prop:infty_norm}), $\forall \xb \in \domX$, there is
\begin{equation*}
    |\Hb^{(l)}_j(\xb)| \leq (J_{\sigma})^{l-1} \| \xb \| \CompLp{l}{2}(\Hb^{(l)}_j) = \| \xb \| \| \Hb^{(l)}_j \|_{\Hilb^{(l)}} \prod_{l'=1}^{l-1} \| \mu^{(l')} \|_{\Hilb^{(l')}, 2}~,
\end{equation*}
and hence, as $\sigma$ is $J_{\sigma}$-Lipschitz with $\sigma(0) = 0$, it holds that
\begin{equation*}
    \big ( \sigbig{\Hb^{(l)}_j(\xb)} \big )^2 \leq (J_{\sigma})^2 \big ( \Hb^{(l)}_j(\xb) \big )^2 \leq (J_{\sigma})^{2l} \| \xb \|^2 \| \Hb^{(l)}_j \|_{\Hilb^{(l)}}^2 \prod_{l'=1}^{l-1} \| \mu^{(l')} \|_{\Hilb^{(l')}, 2}^2~,
\end{equation*}
which is bounded almost surely by $(J_{\sigma})^{2l} \| \xb \|^2 \prod_{l'=1}^{l} \| \mu^{(l')} \|_{\Hilb^{(l')}, \infty}^2$. Thus,
\begin{equation*}
    \begin{split}
        \expenu{\xb \sim \nu}{(\text{II})} \leq & ~ \frac{1}{m} (J_{\sigma})^{2l} M_{\nu} \EEElr{ \left ( \Xi^{(l)}(\Hb^{(l+1)}_i, \Hb^{(l)}) \right )^2  \Big | \Hb^{(l+1)}_i } \prod_{l'=1}^{l} \| \mu^{(l')} \|_{\Hilb^{(l')}, \infty}^2 \\
        \leq &~ \frac{1}{m} (J_{\sigma})^{2l} M_{\nu} \| \Hb^{(l+1)}_i \|_{\Hilb^{(l+1)}}^2 \prod_{l'=1}^{l} \| \mu^{(l')} \|_{\Hilb^{(l')}, \infty}^2~,
    \end{split}
\end{equation*}
and the right-hand side is bounded almost surely by $\frac{1}{m} (J_{\sigma})^{2l} M_{\nu} \prod_{l'=1}^{l+1} \| \mu^{(l')} \|_{\Hilb^{(l')}, \infty}^2$.

 Therefore, combining the bounds for $(\text{I})$ and $(\text{II})$ with Lemma~\ref{lem:square_sum}, and further choosing $\lambda = l - 1$, we get
\begin{equation*}
\begin{split}
    &~ \expenu{\xb \sim \nu}{\EEElr{ \Big ( \bar{\Hb}^{(l+1)}_i(\xb) - \Hb^{(l+1)}_i(\xb) \Big )^2 \Big | \Hb^{(l+1)}_i }} \\
    \leq &~ ((1 + 1 / \lambda) (l-1)^2 + (1 + \lambda)) (J_{\sigma})^{2l} M_{\nu} \prod_{l'=1}^{l+1} \| \mu^{(l')} \|_{\Hilb^{(l')}, \infty}^2 \\
    \leq & ~ \frac{l^2}{m} (J_{\sigma})^{2l} M_{\nu} \prod_{l'=1}^{l+1} \| \mu^{(l')} \|_{\Hilb^{(l')}, \infty}^2
\end{split}
\end{equation*}

which proves the inductive hypothesis at level $l+1$.

\begin{remark}
    Note that using Lemma~\ref{lem:square_sum} and optimizing for $\lambda$ as a function of $l$---compared to just using the basic inequality $(a+b)^2 \leq 2(a^2 + b^2)$---reduces the explicit dependency of the right-hand side of \eqref{eq:inductive_approx} on $l$ from exponential to quadratic.
\end{remark}

\section{Supplementary Materials for Section~\ref{sec:gen}}
\subsection{Proof of Theorem~\ref{prop:rad}}
\label{app:pf_rad}
When $\sigma$ is $1$-homogeneous, we see that $\CompLp{L}{}$ can be alternatively expressed as
\begin{equation*}
\begin{split}
    \CompLp{L}{}(f) =&~ \inf_{\mu^{(1)}, ..., \mu^{(L-1)}} \| f \|_{\Hilb^{(L)}} \\
    & \hspace{25pt} \text{s.t.   } \hspace{5pt} \| \mu^{(l)} \|_{\Hilbl{l}, 2} = 1~,~ \forall l \in [L-1]~.
\end{split}
\end{equation*}
In the following, for brevity, we will write $\sup_{\mu^{(l)}}$ and $\sup_{\xi}$ for 
\begin{equation*}
    \sup_{\substack{\mu^{(l)} \in \Pcal(\Hilbl{l}) \\ \| \mu^{(l)} \|_{\Hilb^{(l)}, 2} \leq 1}}~ \text{ and } \quad \sup_{\substack{\xi \in L^2(\Hilbl{L-1}, \mu^{(L-1)}) \\ \| \xi \|_{L^2(\Hilbl{L-1}, \mu^{(L-1)})} \leq 1}}~,
\end{equation*}
respectively.
By the definition of the empirical Rademacher complexity,
\begin{equation*}
     \widehat{\Rad}_{S}(\Ball(\Ladder{L})) \coloneqq \EE_{\taub} \left [\frac{1}{n} \sup_{\CompL{L}(f) \leq 1} \summ{k}{n} \tau_{k} f(\xb_{k}) \right ]~.
\end{equation*}
For any $\lambda > 0$, we consider the function $g_\lambda: \Rbb \to \Rbb$ defined by $g_\lambda(u) = \exp(\lambda u)$, which is positive, monotonically increasing and convex. Thus, using Jensen's inequality,  we can write
\begin{equation*}
    \begin{split}
        n~ \widehat{\Rad}_{S}(\Ball(\Ladder{L})) = &~ \frac{1}{\lambda} \log \left ( g_\lambda \left ( \EE_{\taub} \left [ \sup_{\CompL{L}(f) \leq 1} \summ{k}{n} \tau_{k} f(\xb_{k})  \right ] \right ) \right ) \\
        \leq &~ \frac{1}{\lambda} \log \left ( \EE_{\taub} \left [ g_\lambda \left ( \sup_{\CompL{L}(f) \leq 1} \summ{k}{n} \tau_{k} f(\xb_{k})  \right ) \right ]  \right ) \\
        \leq &~ \frac{1}{\lambda} \log \mathcal{M}^{(L)}_\lambda ~,
    \end{split}
\end{equation*}
where we define, $\forall l \in [L]$, $\mathcal{M}^{(l)}_\lambda \coloneqq \EE_{\taub} \left [   g_\lambda \left ( \sup_{\CompL{l}(f) \leq 1}  \Big | \summ{k}{n} \tau_{k} f(\xb_{k}) \Big | \right ) \right ]$.
\begin{lemma} Under the assumptions on $\sigma$ in Theorem~\ref{prop:rad},
\label{lem:M_lambda}
    \begin{equation*}
        \mathcal{M}^{(L)}_\lambda \leq (2 J_{\sigma})^{L-1} \EE_{\taub} \left [  g_\lambda \left ( \left \| \summ{k}{n} \tau_{k} \xb_{k} \right \| \right ) \right ]~.
    \end{equation*}
\end{lemma}
This lemma is proved in Appendix~\ref{app:pf_lem_M_lambda}. Then, if we choose
$\lambda = \frac{\sqrt{2 (L-1) \log (2 J_{\sigma})}}{\sqrt{\summ{k}{n} \| \xb_{k} \|_{2}^2}}$, 
we can generalize the argument used by \citet{golowich2018size} slightly to show that
\begin{equation*}
    \frac{1}{\lambda} \log \left ( 2^{L-1} \EE_{\taub} \left [  g_\lambda \left ( \left \| \summ{k}{n} \tau_{k} \xb_{k} \right \| \right ) \right ] \right ) \leq \big (\sqrt{2 L \log (2 J_{\sigma})} + 1 \big ) \sqrt{\summ{k}{n} \| \xb_{k} \|_{2}^2}~,
\end{equation*}
which yields the desired result for the case $M=1$. The extension to general $M > 0$ is easy to see because of the linearity of Hilbert space norms.

\subsubsection{Proof of Lemma~\ref{lem:M_lambda}}
\label{app:pf_lem_M_lambda}
We see that
\begin{equation*}
    \begin{split}
        \sup_{\CompL{L}(f) \leq 1} \left | \summ{k}{n} \tau_{k} f(\xb_{k}) \right | \leq &~ \sup_{\mu^{(1)}, ..., \mu^{(L-1)}, \xi} \left | \summ{k}{n} \tau_{k} \int \xi(h) \sigbig{h(\xb_{k})} \mu^{(L-1)}(dh) \right | \\
        \leq &~ \sup_{\mu^{(1)}, ..., \mu^{(L-1)}, \xi} \left | \int \summ{k}{n} \tau_{k} \frac{\sigbig{h(\xb_{k})}}{\|h\|_{\Hilbl{L-1}}} \xi(h) \|h\|_{\Hilbl{L-1}} \mu^{(L-1)}(dh) \right |~.
    \end{split}
\end{equation*}
By H\"older's inequality and the homogeneity of $\sigma$, there is 
\begin{equation*}
    \begin{split}
        &~ \left | \int \summ{k}{n} \tau_{k} \frac{\sigbig{h(\xb_{k})}}{\|h\|_{\Hilbl{L-1}}} \xi(h) \|h\|_{\Hilbl{L-1}} \mu^{(L-1)}(dh) \right | \\
        \leq &~ \left ( \sup_{h \in \Hilbl{L-1}} \left | \summ{k}{n} \tau_{k}  \frac{\sigbig{h(\xb_{k})}}{\|h\|_{\Hilbl{L-1}}}\right | \right ) \int |\xi(h)| \|h\|_{\Hilbl{L-1}} \mu^{(L-1)}(dh) \\
        \leq &~ \left ( \sup_{\| \hat{h} \|_{\Hilbl{L-1}} \leq 1} \left | \summ{k}{n} \tau_{k} \sigbig{\hat{h}(\xb_{k})} \right | \right ) \left (\int |\xi(h)|^2 \mu^{(L-1)}(dh) \right )^{1/2} \left (\int\|h\|_{\Hilbl{L-1}}^2 \mu^{(L-1)}(dh) \right )^{1/2} ~,
    \end{split}
\end{equation*}
and hence
\begin{equation*}
\begin{split}
    \sup_{\CompL{L}(f) \leq 1} \left | \summ{k}{n} \tau_{k} f(\xb_{k}) \right | 
    \leq &~ \sup_{\substack{\| \hat{h} \|_{\Hilbl{L-1}} \leq 1 \\ \mu^{(1)}, ..., \mu^{(L-2)}}} \left | \summ{k}{n} \tau_{k} \sigbig{\hat{h}(\xb_{k})} \right | \\
    =&~
    \begin{cases} 
    \sup_{\CompL{L-1}(\hat{h}) \leq 1} \left | \summ{k}{n} \tau_{k} \sigbig{\hat{h}(\xb_{k})} \right |~, & \text{ if } L \geq 3~, \\
    \sup_{\| \hat{h} \|_{\Hilbl{1}} \leq 1} \left | \summ{k}{n} \tau_{k} \sigbig{\hat{h}(\xb_{k})} \right |~, & \text{ if } L = 2~.
    \end{cases}
\end{split}
\end{equation*}
Notice that, since $g_\lambda$ is positive, there is $g_\lambda(|u|) \leq g_\lambda(u) + g_\lambda(-u)$. In addition, the assumptions on $\sigma$ imply that it is $J_{\sigma}$-Lipschitz. Therefore, when $L \geq 3$,
\begin{equation*}
    \begin{split}
        \mathcal{M}^{(L)}_\lambda 
        \leq &~ \EE_{\taub} \Bigg [  g_\lambda \Bigg ( \sup_{\CompL{L}(f) \leq 1}  \Bigg | \summ{k}{n} \tau_{k} f(\xb_{k}) \Bigg | \Bigg ) \Bigg ] \\
        = &~ \EE_{\taub} \Bigg [  \sup_{\CompL{L-1}(\hat{h}) \leq 1} g_\lambda \Bigg ( \Bigg | \summ{k}{n} \tau_{k} \sigbig{\hat{h}(\xb_{k})} \Bigg | \Bigg ) \Bigg ] \\
        \leq &~ \EE_{\taub} \Bigg [  \sup_{\CompL{L-1}(\hat{h}) \leq 1}  g_\lambda \Bigg (\summ{k}{n} \tau_{k} \sigbig{\hat{h}(\xb_{k})} \Bigg ) \Bigg ]  + \EE_{\taub} \Bigg [  \sup_{\CompL{L-1}(\hat{h}) \leq 1}  g_\lambda \Bigg (- \summ{k}{n} \tau_{k} \sigbig{\hat{h}(\xb_{k})} \Bigg ) \Bigg ] \\
        = &~ \EE_{\taub} \Bigg [  g_\lambda \Bigg (\sup_{\CompL{L-1}(\hat{h}) \leq 1} \summ{k}{n} \tau_{k} \sigbig{\hat{h}(\xb_{k})} \Bigg ) \Bigg ]  + \EE_{\taub} \Bigg [ g_\lambda \Bigg (\sup_{\CompL{L-1}(\hat{h}) \leq 1} \summ{k}{n} (-\tau_{k}) \sigbig{\hat{h}(\xb_{k})} \Bigg ) \Bigg ] \\
        = &~ 2 \EE_{\taub} \Bigg [  g_\lambda \Bigg (\sup_{\CompL{L-1}(\hat{h}) \leq 1} \summ{k}{n} \tau_{k} \sigbig{\hat{h}(\xb_{k})} \Bigg ) \Bigg ] \\
        \leq &~ 2 J_{\sigma} \EE_{\taub} \Bigg [  g_\lambda \Bigg (\sup_{\CompL{L-1}(\hat{h}) \leq 1} \summ{k}{n} \tau_{k} \hat{h}(\xb_{k}) \Bigg ) \Bigg ]~,
    \end{split}
\end{equation*}
where for the second and the fourth lines we use the monotonicity of $g_\lambda$; for the fifth line we use the symmetry of the Rademacher distribution; and for the sixth line we apply the \emph{contraction lemma} (Equation 4.20 in \citealp{ledoux1991probability}) by using the Lipschitzness of $\sigma$ and the monotonicity and convexity of $g_\lambda$. Hence, we derive that
\begin{equation*}
    \mathcal{M}^{(L)}_\lambda \leq 2 J_{\sigma} \mathcal{M}^{(L-1)}_\lambda~.
\end{equation*}
Thus, by induction, it holds that
\begin{equation*}
    \begin{split}
        \mathcal{M}^{(L)}_\lambda \leq&~ (2 J_{\sigma})^{L-1} \mathcal{M}^{(1)}_\lambda \\
        =&~ (2 J_{\sigma})^{L-1} \EE_{\taub} \left [   g_\lambda \left ( \sup_{\|f\|_{\Hilbl{1}} \leq 1}  \left | \summ{k}{n} \tau_{k} f(\xb_{k}) \right | \right ) \right ] \\
        =&~ (2 J_{\sigma})^{L-1} \EE_{\taub} \left [   g_\lambda \left ( \sup_{\|\zb\|_{2} \leq 1}  \left | \summ{k}{n} \tau_{k} \zb^{\intercal} \cdot \xb_{k} \right | \right ) \right ] \\
        \leq &~ (2 J_{\sigma})^{L-1} \EE_{\taub} \left [  g_\lambda \left ( \left \| \summ{k}{n} \tau_{k} \xb_{k} \right \| \right ) \right ] ~,
    \end{split}
\end{equation*}
which proves the lemma.

\section{Supplementary Materials for Section~\ref{sec:training}}
\subsection{Proof of Theorem~\ref{prop:lln}}
\label{app:pf_lln}
\paragraph{Notations}
If $(u_{m})_{m \in \mathbb{N}_+}$ and $(u'_{m})_{m \in \mathbb{N}_+}$ are two sequences of non-negative random variables, we write $u_{m} = o_{\mathbb{P}}(u_{m}')$ if it holds almost surely that $\forall \epsilon > 0$, $\exists M > 0$ such that $\forall m > M$, $u_{m} \leq \epsilon u'_{m}$.

\paragraph{Preliminaries} 
We start by defining the limiting dynamics in an alternative way in terms of flow maps and will show its equivalence to the definition given in Section~\ref{sec:mfnhl}. We will also generalize the analysis by taking into account the bias terms as introduced in Appendix~\ref{app:bias}. For $t \geq 0$, let $\Ab_{t}$, $\Xib^{(1)}_{t}$, ..., $\Xib^{(L-2)}_{t}$, $\Bb^{(1)}_{t}$, ..., and $\Bb^{(L-1)}_{t}$ be random variables, let $\Zb_{t}$ be a $d$-dimensional random vector, and let $\Htl{t}{1}$, ..., $\Htl{t}{L-1}$ and $\Qtl{t}{1}$, ..., $\Qtl{t}{L-1}$ be random fields on $\domX$, which all depend on time and are defined as follows. At initial time, we let $\Ab_{0}$, $\Zb_{0}$, $\Bb^{(1)}_{0}$, ..., $\Bb^{(L-1)}_{0}$ be distributed independently with laws $\rhoa$, $\rhoz$, $\rho_{b^{(1)}}$, ..., $\rho_{b^{(L-1)}}$, respectively.
Then, we define $\Htl{t}{l}$, $\Qtl{t}{l}$, $\Xitl{t}{l}$, $\Btl{t}{l}$ and $\Zb_{t}$ alternatively as:
\begin{equation}
\label{eq:flow}
    \begin{split}
        \Htl{t}{l}(\xb) =&~ \begin{cases}
    \Hdtl{t}{1}(\xb, \Zb_{0}, \Btl{0}{1})~, &\quad l = 1 \\
    \Hdtl{t}{l}(\xb, \Btl{0}{l})~, & \quad l \in \{2, ..., L-2\} \\
    \Hdtl{t}{L-1}(\xb, \Ab_{0}, \Btl{0}{L-1})~, &\quad l = L-1
    \end{cases} \\
    \Qtl{t}{l}(\xb) =&~ \begin{cases}
    \Qdtl{t}{1}(\xb, \Zb_{0}, \Btl{0}{1})~, &\quad l = 1 \\
    \Qdtl{t}{l}(\xb, \Btl{0}{l})~, & \quad l \in \{2, ..., L-2\} \\
    \Qdtl{t}{L-1}(\xb, \Ab_{0}, \Btl{0}{L-1})~, &\quad l = L-1
    \end{cases} \\
    \Xitl{t}{l} =&~ \begin{cases}
    \Xidtl{t}{1}(\Zb_{0}, \Btl{0}{1}, \Btl{0}{2})~, &\quad l = 1 \\
    \Xidtl{t}{l}(\Btl{0}{l}, \Btl{0}{l+1})~, & \quad l \in \{2, ..., L-2\} \\
    \Xidtl{t}{L-2}(\Ab_{0}, \Btl{0}{L-2}, \Btl{0}{L-1})~, &\quad l = L-2~,
    \end{cases} \\
    \Btl{t}{l} =&~ \begin{cases}
    \Bdtl{t}{1}(\Zb_{0}, \Btl{0}{1})~, &\quad l = 1 \\
    \Bdtl{t}{l}(\Btl{0}{l})~, & \quad l \in \{2, ..., L-2\} \\
    \Bdtl{t}{L-1}(\Ab_{0}, \Btl{0}{L-1})~, &\quad l = L-1
    \end{cases} \\
    \Zb_{t} =&~ Z_{t}(\Zb_{0}, \Btl{0}{1})~, \\
    \Ab_{t} =&~ A_{t}(\Ab_{0}, \Btl{0}{L-1})~.
    \end{split}
\end{equation}
by introducing the following (deterministic) functions:
\begin{equation*}
    \begin{split}
        \Hdtl{t}{l}, \Qdtl{t}{l} :&~ \begin{cases}
        \domX \times \Rbb^d \times \Rbb \to \Rbb~,& \quad l = 1, \\
        \domX \times \Rbb \to \Rbb~,& \quad l \in \{2, ..., L-2\}~,\\
        \domX \times \Rbb \times \Rbb \to \Rbb ~,& \quad l = L-1~,
    \end{cases} \\
    \Xidtl{t}{l} :&~ \begin{cases}
        \Rbb^d \times \Rbb \times \Rbb \to \Rbb~,& \quad l = 1~, \\
        \Rbb \times \Rbb \to \Rbb~,& \quad l \in \{2, ..., L-3 \}~,\\
        \Rbb \times \Rbb \times \Rbb \to \Rbb ~,& \quad l = L-2~,
    \end{cases}
    \end{split}
\end{equation*}
\begin{align*}
    \Bdtl{t}{l} :&~ \begin{cases}
        \Rbb^d \times \Rbb \to \Rbb~,& \quad \quad \quad l = 1~, \\
        \Rbb \to \Rbb~,& \quad \quad \quad l \in \{2, ..., L-2 \}~,\\
        \Rbb \times \Rbb \to \Rbb ~,& \quad \quad \quad l = L-1~,
    \end{cases} \\
    Z_{t} :&~ \Rbb^d \times \Rbb \to \Rbb^d~, \\
    A_{t} :&~ \Rbb \times \Rbb \to \Rbb~,
\end{align*}
They can be interpreted as flow maps and are defined as follows: $\forall t \geq 0$,
\begin{itemize}
    \item For $l \in [L-1]$, $\Hdtl{t}{l}$ is defined by, $\forall \xb \in \domX$, $\forall \zb \in \Rbb^d$, $\forall a, b \in \Rbb$,
\begin{align*}
    \Hdtl{t}{1}(\xb, \zb, b) =&~ Z_{t}(\zb, b)^{\intercal} \cdot \xb~, \\
    \Hdtl{t}{2}(\xb, b) =&~ \EEElr{\Xidtl{t}{1}(\Zb_{0}, \Btl{0}{1}, b) \sigbig{\Hdtl{t}{1}(\xb, \Zb_{0}, \Btl{0}{1}) + \Bdtl{t}{1}(\Zb_{0}, \Bdtl{0}{1})} }~, \\
    \Hdtl{t}{l+1}(\xb, b) =&~ \EEElr{\Xidtl{t}{l}(\Btl{0}{l}, b) \sigbig{\Hdtl{t}{l}(\xb, \Btl{0}{l}) + \Bdtl{t}{l}(\Btl{0}{l})} }~, \quad \forall l \in \{2, ..., L-3\}~, \\
    \Hdtl{t}{L-1}(\xb, a, b) =&~ \EEElr{\Xidtl{t}{L-2}(a, \Btl{0}{L-2}, b) \sigbig{\Hdtl{t}{L-2}(\xb, \Btl{0}{L-2}) + \Bdtl{t}{L-2}(\Btl{0}{L-2})} }~.
\end{align*}

\item For $l \in [L-1]$, $\Qdtl{t}{l}$ is defined by, $\forall \xb \in \domX$, $\forall \zb \in \Rbb^d$, $\forall a, b \in \Rbb$,
\begin{align*}
    \Qdtl{t}{L-1}(\xb, a, b) =&~ A_{t}(a, b)~, \\
    \Qdtl{t}{L-2}(\xb, b) =&~ \EEEB{\Xidtl{t}{L-2}(\Ab_{0}, b, \Btl{0}{L-1}) \Qdtl{t}{L-1}(\xb, \Btl{0}{L-1}) \sigpbig{\Hdtl{t}{L-1}(\xb, \Ab_{0},\Btl{0}{L-1}) \\
    &~ \hspace{180pt}+ \Bdtl{t}{L-1}(\Ab_{0},\Btl{0}{L-1})}}~, \\
    \Qdtl{t}{l-1}(\xb, b) =&~ \EEElr{\Xidtl{t}{l-1}(b, \Btl{0}{l}) \Qdtl{t}{l}(\xb, \Btl{0}{l}) \sigpbig{\Hdtl{t}{l}(\xb, \Btl{0}{l}) + \Bdtl{t}{l}(\Btl{0}{l})}}, \\
    &\hspace{230pt} \forall l \in \{1, ..., L-2\}~, \\
    \Qdtl{t}{1}(\xb, \zb, b) =&~ \EEElr{\Xidtl{t}{1}(\zb, b, \Btl{0}{2}) \Qdtl{t}{2}(\xb, \Btl{0}{2}) \sigpbig{\Hdtl{t}{2}(\xb, \Btl{0}{2}) + \Bdtl{t}{2}(\Btl{0}{2})}}~.
\end{align*}

\item For $l \in [L-2]$, $\Xidtl{t}{l}$ is defined by, $\forall \zb \in \Rbb^d$, $\forall a, b \in \Rbb$~,
\begin{equation*}
    \begin{split}
    \ddt \Xidtl{t}{1}(\zb, b, b') =&~ - \expenuB{\xb}{\zeta_{t}(\xb) \Qdtl{t}{2}(\xb, b') \sigpbig{\Hdtl{t}{2}(\xb, b') + \Bdtl{t}{2}(b')} \\
    &~ \hspace{90pt} \sigbig{\Hdtl{t}{1}(\xb, \zb, b) + \Bdtl{t}{1}(\zb, b)}}~, \\
    \ddt \Xidtl{t}{l}(b, b') =&~ - \expenuB{\xb}{\zeta_{t}(\xb) \Qdtl{t}{l+1}(\xb, b') \sigpbig{\Hdtl{t}{l+1}(\xb, b') + \Bdtl{t}{l+1}(b')} \\
    &~ \hspace{90pt} \sigbig{\Hdtl{t}{l}(\xb, b) + \Bdtl{t}{l}(b)}} ~, \quad \forall l \in \{2, ..., L-2\}~, \\    
    \ddt \Xidtl{t}{L-2}(a', b, b') =&~ - \expenuB{\xb}{\zeta_{t}(\xb) \Qdtl{t}{L-1}(\xb, a', b') \sigpbig{\Hdtl{t}{L-1}(\xb, a', b') + \Bdtl{t}{L-1}(a', b')} \\
    &~ \hspace{90pt} \sigbig{\Hdtl{t}{l}(\xb, b) + \Bdtl{t}{l}(b)}} ~,
    \end{split}
\end{equation*}

together with the initial conditions
\begin{align*}
    \Xidtl{0}{1}(\zb, b, b') =&~ 0~, \\
    \Xidtl{0}{l}(b, b') =&~ 0 ~, \quad \forall l \in \{2, ..., L-3\}~, \\    
    \Xidtl{0}{L-2}(a', b, b') =&~ 0~.
    \end{align*}

\item For $l \in [L-1]$, $\Bdtl{t}{l}$ is defined by,  $\forall \zb \in \Rbb^d$, $\forall a, b \in \Rbb$,
\begin{align*}
    \ddt \Bdtl{t}{1}(\zb, b) =&~ -\beta \expenu{\xb}{\zeta_{t}(\xb) \Qdtl{t}{1}(\xb, \zb, b) \sigpbig{\Hdtl{t}{1}(\xb, \zb, b) + \Bdtl{t}{1}(\zb, b)}}~,  \\
    \ddt \Bdtl{t}{l}(b) =&~ -\beta \expenu{\xb}{\zeta_{t}(\xb) \Qdtl{t}{l}(\xb, b) \sigpbig{\Hdtl{t}{l}(\xb, b) + \Bdtl{t}{l}(b)}}~, \\
    &~ \hspace{200pt} \forall l \in \{2, ..., L-2 \}~, \\
    \ddt \Bdtl{t}{L-1}(a, b) =&~ -\beta \expenu{\xb}{\zeta_{t}(\xb) \Qdtl{t}{L-1}(\xb, a, b) \sigpbig{\Hdtl{t}{L-1}(\xb, a, b) + \Bdtl{t}{L-1}(a, b)}}~,
\end{align*}
together with the initial conditions
\begin{align*}
    \Bdtl{0}{1}(\zb, b) =&~ b~,  \\
    \Bdtl{0}{l}(b) =&~ b~, \quad \forall l \in \{2, ..., L-2 \}~, \\
    \Bdtl{0}{L-1}(a, b) =&~ b~.
\end{align*}

\item $Z_{t}$ is defined by, $\forall \zb \in \Rbb^d$, $\forall b \in \Rbb$,
\begin{equation*}
    \ddt Z_{t}(\zb, b) = - \expenu{\xb}{\zeta_{t}(\xb) \Qdtl{t}{1}(\xb, \zb, b) \sigpbig{\Hdtl{t}{1}(\xb, \zb, b) + \Bdtl{t}{1}(\zb, b)} \hspace{1pt} \xb }~,
\end{equation*}
together with the initial condition
\begin{equation*}
    Z_{0}(\zb, b) = \zb~.
\end{equation*}

\item $A_{t}$ is defined by, $\forall a, b \in \Rbb$,
\begin{equation*}
    \ddt A_{t}(a, b) = - \expenu{\xb}{\zeta_{t}(\xb) \sigbig{\Hdtl{t}{l}(\xb, a, b) + \Bdtl{t}{l}(a, b)}  }~,
\end{equation*}
together with the initial condition
\begin{equation*}
    A_{0}(a, b) = a~.
\end{equation*}
\end{itemize}
Lastly, $f_{t}$ and $\zeta_{t}$ are defined in the same way as in Section~\ref{sec:mfnhl}.

Defined in this way, one can verify that $\Xib^{(l)}_{0} \coloneqq 0$, $\forall l \in [L-2]$, and furthermore, for $t \geq 0$, the dynamics of $\Ab_{t}$, $\Zb_{t}$, $\Xib^{(1)}_{t}$, ..., $\Xib^{(L-2)}_{t}$, $\Bb^{(1)}_{t}$, ..., $\Bb^{(L-1)}_{t}$ satisfy the following equations:
\begin{align*}
    \ddt \Ab_{t} =&~ - \expenuB{\xb}{\zeta_{t}(\xb) \sigbig{\Htl{t}{L-1}(\xb) + \Btl{t}{L-1}} }~, \\
    \ddt \Zb_{t} =&~ - \expenuB{\xb}{\zeta_{t}(\xb) \Qtl{t}{1}(\xb) \sigpbig{\Htl{t}{1}(\xb) + \Btl{t}{1}} \hspace{1pt}\xb }~, \\
    \ddt \Bb^{(l)}_{t} =&~ - \beta \expenu{\xb}{\zeta_{t}(\xb) \Qtl{t}{l}(\xb) \sigpbig{\Htl{t}{l}(\xb) + \Btl{t}{l}}}~, \quad \forall l \in [L-1],
\end{align*}
and $\forall l \in [L-2]$,
\begin{equation}
    \ddt \Xitl{t}{l} =- \mathcal{E}_{\xb} \Big \{ \zeta_{t}(\xb) \Qtl{t}{l+1}(\xb) \sigpbig{\Htl{t}{l+1}(\xb) + \Btl{t}{l+1}} \sigbig{\Htl{t}{l}(\xb) + \Btl{t}{l}} \Big \}~, \label{eq:ddt_Xitl} 
\end{equation}
and moreover, the random fields satisfy $\Htl{t}{1}(\xb) = \Zb_{t}^{\intercal} \cdot \xb$, $\Qtl{t}{L-1}(\xb) = \Ab_{t}$,
and $\forall l \in [L-2]$,
\begin{align}
        \Hb^{(l+1)}_{t}(\xb) =&~ \EEElr{\Xib^{(l)}_{t} \sigbig{\Htl{t}{l}(\xb) + \Btl{t}{l}} \Big | \Hb^{(l+1)}_{t}}~. \label{eq:Ht_entang} \\
    \Qtl{t}{l}(\xb) =&~ \EEElr{\Xib^{(l)}_{t} \Qtl{t}{l+1}(\xb) \Big | \Htl{t}{l}}~. \label{eq:Qtl_inte}
\end{align}
From \eqref{eq:ddt_Xitl}, we derive that, $\forall l \in [L-2]$,
\begin{equation}
\label{eq:Xitl_inte}
     \Xitl{t}{l} = - \int_{0}^t \expenub{\xb}{\zeta_{s}(\xb) \Qtl{s}{l+1}(\xb) \sigpbig{\Htl{s}{l+1}(\xb) + \Btl{s}{l+1}} \sigbig{\Htl{s}{l}(\xb) + \Btl{s}{l}}} ds~.
\end{equation}
When $\rho_{b^{(1)}} = ... = \rho_{b^{(L-1)}} = \delta_{0}$ and $\beta = 0$, there is $\Btl{t}{l} = 0$, $\forall t \geq 0$, $\forall l \in [L-1]$. Then, substituting \eqref{eq:Xitl_inte} into \eqref{eq:Ht_entang} and \eqref{eq:Qtl_inte}, we obtain \eqref{eq:Htl} and \eqref{eq:Qtl}. Hence, the definitions given by \eqref{eq:flow} are indeed consistent with the mean-field dynamics described in Section~\ref{sec:mfnhl}.

To simpilify the propagation-of-chaos argument in the remaining proof, we also define
\begin{align*}
    \tilzbjt{i}{t} =&~ Z_{t}(\zbjt{i}{0}, \bitl{i}{0}{1}) \\
    \tilait{i}{t} =&~ A_{t}(a_{i, 0}, \bitl{i}{0}{L-1}) \\
    \tilbitl{i}{t}{l} =&~ \begin{cases}
        \Bdtl{t}{1}(\zbjt{i}{0}, \bitl{i}{0}{1})~,& \quad l = 1 \\
        \Bdtl{t}{l}(\bitl{i}{0}{l})~,& \quad l \in \{2, ..., L-2 \} \\
        \Bdtl{t}{L-1}(a_i, \bitl{i}{0}{L-1})~,& \quad l = L-1
    \end{cases} \\
    \tilhitl{i}{t}{l}(\xb) =&~ \begin{cases}
        \Hdtl{t}{1}(\xb, \zbjt{i}{0}, \bitl{i}{0}{1})~,& \quad l = 1\\
        \Hdtl{t}{l}(\xb, \bitl{i}{0}{l})~,& \quad l \in \{2, ..., L-2\} \\
        \Hdtl{t}{L-1}(\xb, a_i, \bitl{i}{0}{L-1})~,& \quad l = L-1
    \end{cases} \\
    \tilqitl{i}{t}{l}(\xb) =&~ \begin{cases}
        \Qdtl{t}{1}(\xb, \zbjt{i}{0}, \bitl{i}{0}{1})~,& \quad l = 1\\
        \Qdtl{t}{l}(\xb, \bitl{i}{0}{l})~,& \quad l \in \{2, ..., L-2\} \\
        \Qdtl{t}{L-1}(\xb, a_i, \bitl{i}{0}{L-1})~,& \quad l = L-1
    \end{cases} \\
    \tilWijtl{i}{j}{t}{l} =&~ \begin{cases}
        \Xidtl{t}{1}(\zbjt{j}{0}, \bitl{j}{0}{1}, \bitl{i}{0}{2})~,& \quad l = 1 \\
        \Xidtl{t}{l}(\bitl{j}{0}{l}, \bitl{i}{0}{l+1})~,&\quad l \in \{2, ..., L-3 \} \\
        \Xidtl{t}{L-2}(a_i, \bitl{j}{0}{L-2}, \bitl{i}{0}{L-1})~,& \quad l = L-2~.
    \end{cases}  
\end{align*}

\paragraph{Main proof}
Given a function $g: \Rbb^N \to \Rbb$, by 
the triangle inequality, there is
\begin{equation*}
    \begin{split}
        &~  \left | \int g(h(\xbg{1}), ..., h(\xbg{N})) \mu^{(l)}_{m, t}(dh) - \EEElr{ g(\Htl{t}{l}(\xbg{1}), ..., \Htl{t}{l}(\xbg{N})) } \right | \\
        = &~  \left | \frac{1}{m} \summ{i}{m} g(\hitl{i}{t}{l}(\xbg{1}), ..., \hitl{i}{t}{l}(\xbg{N})) - \EEElr{ g(\Htl{t}{l}(\xbg{1}), ..., \Htl{t}{l}(\xbg{N})) } \right | \\
        \leq &~ \text{(I)} + \text{(II)}~,
    \end{split}
\end{equation*}
where 
\begin{align*}
\text{(I)} \coloneqq &~ \left | \frac{1}{m} \summ{i}{m} g(\tilhitl{i}{t}{l}(\xbg{1}), ..., \tilhitl{i}{t}{l}(\xbg{N})) - \EEElr{ g(\Htl{t}{l}(\xbg{1}), ..., \Htl{t}{l}(\xbg{N})) } \right | \\
\text{(II)} \coloneqq &~ \left | \frac{1}{m} \summ{i}{m} g(\hitl{i}{t}{l}(\xbg{1}), ..., \hitl{i}{t}{l}(\xbg{N})) - \frac{1}{m} \summ{i}{m} g(\tilhitl{i}{t}{l}(\xbg{1}), ..., \tilhitl{i}{t}{l}(\xbg{N})) \right |
\end{align*}
For the first term, there is
\begin{equation*}
    \begin{split}
        \text{(I)} = & ~ \begin{cases}
            \Big | \frac{1}{m} \summ{i}{m} g(
            \Hdtl{t}{1}(\xbg{1}, \zb_{i, 0}, \bitl{i}{0}{1}), ..., \Hdtl{t}{1}(\xbg{N}, \zb_{i, 0}, \bitl{i}{0}{1})) \\
            \hspace{20pt} - \EEElr{ g(
            \Hdtl{t}{1}(\xbg{1}, \Zb_{0}, \Btl{0}{1}), ..., \Hdtl{t}{1}(\xbg{N}, \Zb_{0}, \Btl{0}{1}) } \Big |~,& \quad l = 1 \\
            \Big | \frac{1}{m} \summ{i}{m} g(
            \Hdtl{t}{l}(\xbg{1}, \bitl{i}{0}{l}), ..., \Hdtl{t}{l}(\xbg{N},  \bitl{i}{0}{l})) \\
            \hspace{20pt} - \EEElr{ g(
            \Hdtl{t}{l}(\xbg{1}, \Btl{0}{l}), ..., \Hdtl{t}{l}(\xbg{N}, \Btl{0}{l}) } \Big |~,& \quad l \in \{2, ..., L-2 \} \\
            \Big | \frac{1}{m} \summ{i}{m} g(
            \Hdtl{t}{L-1}(\xbg{1}, a_i, \bitl{i}{0}{L-1}), ..., \Hdtl{t}{L-1}(\xbg{N}, a_i, \bitl{i}{0}{L-1})) \\
            \hspace{20pt} - \EEElr{ g(
            \Hdtl{t}{L-1}(\xbg{1}, \Ab_{0}, \Btl{0}{1}), ..., \Hdtl{t}{L-1}(\xbg{N}, \Ab_{0}, \Btl{0}{L-1}) } \Big |~,& \quad l = L-1
        \end{cases}
    \end{split}
\end{equation*}
Since for each $i \in [m]$, $\bitl{i}{0}{l}$, $\zb_{i, 0}$ and $a_i$ are independent realizations of $\Btl{0}{l}$, $\Zb_{0}$ and $\Ab$, and moreover, each $\Hdtl{t}{l}$ is a Lipschitz function at any finite $t \geq 0$, 
we know from the law of large numbers that $\text{(II)} = o_{\PP}(1)$.

For the second term, if $g \in \text{Lip}(\Rbb^N)$, then 
\begin{equation*}
\begin{split}
    \text{(II)} =&~ \left | \frac{1}{m} \summ{i}{m} g(\hitl{i}{t}{l}(\xbg{1}), ..., \hitl{i}{t}{l}(\xbg{N})) - \frac{1}{m} \summ{i}{m} g(\tilhitl{i}{t}{l}(\xbg{1}), ..., \tilhitl{i}{t}{l}(\xbg{N})) \right | \\
    \leq &~ \left ( \frac{1}{N} \summ{k}{N} \left | \Delta \hitl{m}{t}{l} (\xbg{k}) \right |^2 \right )^{1/2}~,
\end{split}
\end{equation*}
where we define, $\forall l \in [L-1]$, $\forall \xb \in \domX$,
\begin{equation*}
    \Delta \hitl{m}{t}{l}(\xb) \coloneqq \left ( \frac{1}{m} \summ{j}{m} \big | \hitl{i}{t}{l}(\xb) - \tilhitl{i}{t}{l}(\xb) \big |^2 \right )^{\frac{1}{2}}~.
\end{equation*}
\begin{lemma} 
\label{lem:Delta_h_mt_N}
$\frac{1}{N} \summ{k}{N} \left | \Delta \hitl{m}{t}{l} (\xbg{k}) \right |^2 = o_{\PP}(1)$.
\end{lemma}
\noindent This lemma is proved in Appendix~\ref{app:pf_lem_Delta_h_mt_N} using a propagation-or-chaos argument \citep{braun1977vlasov}, and it implies that $\text{(II)} = o_{\PP}(1)$. This concludes this proof of Theorem~\ref{prop:lln}.

\subsubsection{Proof of Lemma~\ref{lem:Delta_h_mt_N}}
\label{app:pf_lem_Delta_h_mt_N}
We additionally define
\begin{align*}
\Delta \zeta_{m, t}(\xb) \coloneqq &~ | \zeta_{m, t}(\xb) - \zeta_{t}(\xb)| \\
    \Delta \zb_{m, t} \coloneqq &~ \left ( \frac{1}{m} \summ{j}{m} | \zbjt{j}{t} - \tilzbjt{j}{t}|^2 \right )^{\frac{1}{2}}~, \\
    \Delta a_{m, t} \coloneqq &~ \left ( \frac{1}{m} \summ{i}{m} | a_{i, t} - \tilde{a}_{i, t}|^2 \right )^{\frac{1}{2}}~, \\
     \Delta \bitl{m}{t}{l} \coloneqq &~ \left ( \frac{1}{m} \summ{i}{m} | \bitl{i}{t}{l} - \tilbitl{i}{t}{l}|^2 \right )^{\frac{1}{2}}~, \quad \forall l \in [L-1]~, \\
     \Delta \qitl{m}{t}{l}(\xb) \coloneqq &~ \left ( \frac{1}{m} \summ{j}{m} \left | \qitl{i}{t}{l}(\xb) \sigpbig{\hitl{i}{t}{l}(\xb) + \bitl{i}{t}{l}} - \tilqitl{i}{t}{l}(\xb) \sigpbig{\tilhitl{i}{t}{l}(\xb) + \tilbitl{i}{t}{l}} \right |^2 \right )^{\frac{1}{2}}~, \quad \forall l \in [L-1]~, \\
     \Delta W_{m, t}^{(l)} \coloneqq &~ \left ( \frac{1}{m^2} \summ{i, j}{m} | \Wijtl{i}{j}{t}{l} - \Wijtl{i}{j}{0}{l} - \tilde{W}^{(l)}_{i, j, t})|^2 \right )^{\frac{1}{2}}~, \quad \forall l \in [L-2] \\
     =&~ \left ( \frac{1}{m^2} \| W_{t}^{(l)} - W_{0}^{(l)} - \Tilde{W}_{t}^{(l)} \|_{\Frob}^2 \right )^{\frac{1}{2}} \\
     \geq &~ \left ( \frac{1}{m^2} \| (W_{t}^{(l)} - W_{0}^{(l)} - \Tilde{W}_{t}^{(l)})^{\intercal} (W_{t}^{(l)} - W_{0}^{(l)} - \Tilde{W}_{t}^{(l)}) \|_{2} \right )^{\frac{1}{2}}~,
\end{align*}
and finally,
\begin{equation*}
    \begin{split}
        \Delta_{m, t} =&~ \sup_{k \in [n]} \Delta \zeta_{m, t}(\xb_{k}) + \Delta \zb_{m, t} + \Delta a_{m, t} \\
        &~ + \summ{l}{L-1} \left ( \sup_{k \in [n]} \Delta \hitl{m}{t}{l}(\xb_{k}) + \sup_{k \in [n]} \Delta \qitl{m}{t}{l}(\xb_{k}) + \Delta \bitl{m}{t}{l} \right ) + \summ{l}{L-2} \Delta W_{m, t}^{(l)}~,
    \end{split}
\end{equation*}
At initial time, we see that $\Delta_{m, 0} = 0$. For $t \geq 0$, we will bound its growth by examining each term on the right-hand side.
\begin{enumerate}[\hspace{-6pt}(1)]

\item $\Delta \hitl{m}{t}{l}$ \label{item:Delta_h}

When $l=1$,
\begin{align*}
    \Delta \hitl{m}{t}{1}(\xb) = O(\Delta \zb_{m, t} + \Delta \bitl{m}{t}{1})
\end{align*}
For $l \in \{2, ..., L-3 \}$,
\begin{align*}
    \hitl{i}{t}{l+1}(\xb) - \tilhitl{i}{t}{l+1}(\xb)
    =&~ (\bitl{i}{t}{l+1} - \tilbitl{i}{t}{l+1}) + \frac{1}{m} \summ{j}{m} \Wijtl{i}{j}{0}{l} \sigbig{\hitlxb{j}{t}{l}{\xb} + \bitl{j}{t}{l}} \\
    &~ + \frac{1}{m} \summ{j}{m} \big ( \Wijtl{i}{j}{0}{l} - \Wijtl{i}{j}{0}{l} - \tilWijtl{i}{j}{t}{l} \big ) \sigbig{\hitlxb{j}{t}{l}{\xb} + \bitl{j}{t}{l}} \\
    &~ + \left ( \frac{1}{m} \summ{j}{m} \tilWijtl{i}{j}{t}{l} \sigbig{\hitlxb{j}{t}{l}{\xb} + \bitl{j}{t}{l}} - \frac{1}{m} \summ{j}{m} \tilWijtl{i}{j}{t}{l} \sigbig{\tilhitlxb{j}{t}{l}{\xb} + \tilbitl{j}{t}{l}} \right ) \\
    & + \left ( \frac{1}{m} \summ{j}{m} \tilWijtl{i}{j}{t}{l} \sigbig{\tilhitlxb{j}{t}{l}{\xb} + \tilbitl{j}{t}{l}} - \Hdtl{t}{l+1}(\xb, \bitl{i}{0}{l+1}) \right )
\end{align*}
By the Marchenko-Pastur law of the eigenvalues of sample covariance matrices \citep{marchenko1967pastur, bai2010spectral}, under the assumption that $\rhoW$ has a finite fourth moment, $\frac{1}{m} \| (W_{0}^{(l)})^{\intercal} W_{0}^{(l)} \|$ converges almost surely to some finite number, and hence
\begin{equation*}
    \begin{split}
        &~ \frac{1}{m} \summ{i}{m} \left | \frac{1}{m} \summ{j}{m} \Wijtl{i}{j}{0}{l} \sigbig{\hitlxb{j}{t}{l}{\xb} + \bitl{j}{t}{l}} \right |^2 \\
        = &~ O \left ( \frac{1}{m} \left ( 1 + \Delta h^{(l)}_{m, t}(\xb) + \Delta b^{(l)}_{m, t} \right )^2 \left ( \frac{1}{m} \| (W_{0}^{(l)})^{\intercal} W_{0}^{(l)} \| \right ) \right ) \\
        =&~ o_{\PP} \big (1 + (\Delta_{m, t})^2 \big )~.
    \end{split}
\end{equation*}
In addition,
\begin{align*}
    &~ \frac{1}{m} \summ{i}{m} \left | \frac{1}{m} \summ{j}{m} \big ( \Wijtl{i}{j}{0}{l} - \Wijtl{i}{j}{0}{l} - \tilWijtl{i}{j}{t}{l} \big ) \sigbig{\hitlxb{j}{t}{l}{\xb} + \bitl{j}{t}{l}} \right |^2 \\
    \leq&~ \frac{1}{m^2} \left \| (W_{t}^{(l)} - W_{0}^{(l)} - \Tilde{W}_{t}^{(l)})^{\intercal} (W_{t}^{(l)} - W_{0}^{(l)} - \Tilde{W}_{t}^{(l)}) \right \|_{2} \frac{1}{m} \summ{j}{m}  \big | \sigbig{\hitlxb{j}{t}{l}{\xb} + \bitl{j}{t}{l}} \big |^2 \\
    =&~ O ( (\Delta W^{(l)}_{m, t})^2 (\Delta h^{(l)}_{m, t}(\xb) + \Delta \bitl{j}{t}{l})^2 )
\end{align*}
Moreover, since the deterministic maps $\Hdtl{t}{l}$ and $\Xidtl{t}{l}$ are Lipschitz at any finite $t \geq 0$, we can deduce from the law of large numbers that $\forall i \in [m]$,
\begin{equation*}
    \begin{split}
        & ~ \left | \frac{1}{m} \summ{j}{m} \tilWijtl{i}{j}{t}{l} \sigbig{\tilhitlxb{j}{t}{l}{\xb} + \tilbitl{j}{t}{l}} - \Hdtl{t}{l+1}(\xb, \bitl{i}{0}{l+1}) \right | \\
        =& ~ \bigg |  \frac{1}{m} \summ{j}{m} \Xidtl{t}{l}(\bitl{i}{0}{l+1}, \bitl{j}{0}{l}) \sigbig{\Hdtl{t}{l}(\xb, \bitl{j}{0}{l}) + \Bdtl{t}{l}(\bitl{j}{0}{l})} \\
        &~ - \EEElr{\Xidtl{t}{l}(\bitl{i}{0}{l+1}, \Btl{0}{l}) \sigbig{\Hdtl{t}{l}(\xb, \Btl{0}{l}) + \Bdtl{t}{l}(\Btl{0}{l})} } \bigg | \\
        =&~ o_{\PP}(1)~.
    \end{split}
\end{equation*}
Thus,
\begin{align*}
    (\Delta \hitl{m}{t}{l+1}(\xb))^2 = &~ O \big ( (\Delta \btl{t}{l+1})^2 + (\Delta W_{t}^{(l)})^2 + (\Delta \hitl{m}{t}{l}(\xb))^2 + (\Delta \btl{t}{l})^2 \big ) \\
    &~+ o_{\PP} \big (1 + (\Delta h^{(l)}_{m, t}(\xb) + \Delta \btl{t}{l})^2 \big )
\end{align*}
and so
\begin{equation*}
    \Delta \hitl{m}{t}{l+1}(\xb) = O \big (\Delta \bitl{m}{t}{l+1} + \Delta W_{m, t}^{(l)} + \Delta \hitl{m}{t}{l}(\xb) + \Delta \bitl{m}{t}{l} \big ) + o_{\PP} \big (1 + \Delta h^{(l)}_{m, t}(\xb) + \Delta \bitl{m}{t}{l} \big )~.
\end{equation*}
With a similar argument, we can obtain the same bound for $\Delta \hitl{m}{t}{2}$ and $\Delta \hitl{m}{t}{l+1}$.

So, by induction, $\forall l \in [L-1]$, $\forall \xb \in \domX$,
\begin{equation}
\label{eq:Delta_h_mt_bound_by_Delta_mt}
\begin{split}
    \Delta \hitl{m}{t}{l}(\xb) =&~ O \left ( \Delta \zb_{m, t} + \summ{l'}{l} \Delta \bitl{m}{t}{l'} + \summ{l'}{l-1} \Delta W_{t}^{(l')} \right ) + o_{\PP}(1 + \Delta h^{(l)}_{m, t}(\xb) + \Delta b^{(l)}_{m, t}) \\
     =&~ O(\Delta_{m, t}) + o_{\PP}(\Delta_{m, t}) ~.
\end{split}
\end{equation}

\item $\Delta \qitl{m}{t}{l}$

For $l = L-1$,
\begin{equation*}
\begin{split}
    &~\big | \qitl{i}{t}{L-1}(\xb) \sigpbig{\hitl{i}{t}{L-1}(\xb) + \bitl{i}{t}{L-1}} - \tilqitl{i}{t}{L-1}(\xb) \sigpbig{\tilhitlxb{i}{t}{L-1}{\xb} + \tilbitl{i}{t}{L-1}} \big | \\
    =&~ \big | a_{i, t} \sigpbig{\hitl{i}{t}{L-1}(\xb) + \bitl{i}{t}{L-1}} - \tilde{a}_{i, t} \sigpbig{\tilhitlxb{i}{t}{L-1}{\xb} + \tilbitl{i}{t}{L-1}} \big |~,
\end{split}
\end{equation*}
and hence
\begin{equation*}
    \Delta \qitl{m}{t}{L-1}(\xb) = O (\Delta a_{m, t} + \Delta \hitl{m}{t}{L-1}(\xb) +  \Delta \bitl{m}{t}{L-1}))~.
\end{equation*}
For $l \in \{3, ..., L-2 \}$, 
\begin{align*}
    &~ \qitl{j}{t}{l-1}(\xb) \sigpbig{\hitl{j}{t}{l-1}(\xb) + \bitl{j}{t}{l-1}} - \tilqitlxb{j}{t}{l-1}{\xb} \sigpbig{\tilhitl{j}{t}{l-1}(\xb) + \tilbitl{j}{t}{l-1}} \\
    =&~ \left ( \frac{1}{m} \summ{i}{m} \Wijtl{i}{j}{0}{l-1} \qitlxb{i}{t}{l}{\xb} \sigpbig{\hitlxb{i}{t}{l}{\xb} + \bitl{i}{t}{l}} \right ) \sigpbig{\hitlxb{j}{t}{l-1}{\xb} + \bitl{j}{t}{l-1}} \\
    &~+ \left ( \frac{1}{m} \summ{i}{m} \left (\Wijtl{i}{j}{t}{l-1} - \Wijtl{i}{j}{0}{l-1}- \tilWijtl{i}{j}{t}{l-1} \right ) \qitlxb{i}{t}{l}{\xb} \sigpbig{\hitlxb{i}{t}{l}{\xb} + \bitl{i}{t}{l}} \right ) \sigpbig{\hitlxb{j}{t}{l-1}{\xb} + \bitl{j}{t}{l-1}} \\
    &~+ \left ( \frac{1}{m} \summ{i}{m} \tilWijtl{i}{j}{t}{l-1} \left (\qitlxb{i}{t}{l}{\xb} \sigpbig{\hitlxb{i}{t}{l}{\xb} + \bitl{i}{t}{l}} - \tilqitlxb{i}{t}{l}{\xb} \sigpbig{\tilhitlxb{i}{t}{l}{\xb} + \tilbitl{i}{t}{l}} \right ) \right ) \sigpbig{\hitlxb{j}{t}{l-1}{\xb} + \bitl{j}{t}{l-1}} \\
    &~+ \left ( \frac{1}{m} \summ{i}{m} \tilWijtl{i}{j}{t}{l-1} \tilqitlxb{i}{t}{l}{\xb} \sigpbig{\tilhitlxb{i}{t}{l}{\xb} + \tilbitl{i}{t}{l}} \right ) \left ( \sigpbig{\hitlxb{j}{t}{l-1}{\xb} + \bitl{j}{t}{l-1}}  - \sigpbig{\tilhitlxb{j}{t}{l-1}{\xb} + \tilbitl{j}{t}{l-1}} \right ) \\
    &~+ \left ( \frac{1}{m} \summ{i}{m} \tilWijtl{i}{j}{t}{l-1} \tilqitlxb{i}{t}{l}{\xb} \sigpbig{\tilhitlxb{i}{t}{l}{\xb} + \tilbitl{i}{t}{l}} - \tilqitlxb{j}{t}{l-1}{\xb} \right ) \sigpbig{\tilhitlxb{j}{t}{l-1}{\xb} + \tilbitl{j}{t}{l-1}} ~.
\end{align*}
Note that $\forall j \in [m]$, by the Lipschitzness of the deterministic maps at finite $t$ and the law of large numbers,
\begin{equation*}
    \begin{split}
        &~ \bigg | \frac{1}{m} \summ{i}{m} \tilWijtl{i}{j}{t}{l-1} \tilqitlxb{i}{t}{l}{\xb} \sigpbig{\tilhitlxb{i}{t}{l}{\xb} + \tilbitl{i}{t}{l}} - \tilqitlxb{j}{t}{l-1}{\xb} \bigg |\\
        \leq &~ \Bigg | \frac{1}{m} \summ{i}{m} \Xidtl{t}{l-1}(\bitl{i}{0}{l}, \bitl{j}{0}{l-1}) \Qdtl{t}{l}(\xb, \bitl{i}{0}{l}) \sigpbig{\Hdtl{t}{l}(\xb, \bitl{i}{0}{l}) + \Bdtl{t}{l}(\bitl{i}{0}{l})} \\
        &~ - \EEElr{\Xidtl{t}{l-1}(\Btl{0}{l}, \bitl{j}{0}{l-1}) \Qdtl{t}{l}(\xb, \Btl{0}{l}) \sigpbig{\Hdtl{t}{l}(\xb, \Btl{0}{l}) + \Bdtl{t}{l}(\Btl{0}{l})}} \Bigg | \\
        =&~ o_{\PP}(1)~.
    \end{split}
\end{equation*}
Via other techniques analogous to those used in part \eqref{item:Delta_h}, we see that
\begin{align*}
    \Delta \qitl{m}{t}{l-1}(\xb) =&~ O (\Delta \qitl{m}{t}{l}(\xb) + \Delta \hitl{m}{t}{l-1}(\xb) + \Delta \bitl{m}{t}{l-1} + \Delta W_{m, t}^{(l)} \\
    &~ + (\Delta \qitl{m}{t}{l}(\xb) + \Delta \hitl{m}{t}{l-1}(\xb) + \Delta \bitl{m}{t}{l-1} + \Delta W_{m, t}^{(l)})^2 ) + o_{\PP}(1 + \Delta \qitl{m}{t}{l}(\xb))
\end{align*}
Similarly, we can obtain the same bound for $\Delta \qitl{m}{t}{L-2}(\xb)$ and $\Delta \qitl{m}{t}{1}(\xb)$.
Thus, by induction,
\begin{equation*}
    \Delta \qitl{m}{t}{l} = O \left ( \Delta_{m, t} + (\Delta_{m, t})^{2^{L-l-1}} \right ) + o_{\PP}\left (1 + \Delta_{m, t} + (\Delta_{m, t})^{2^{L-l-2}} \right )
\end{equation*}

\item $\Delta \bitl{m}{t}{l}$

For $l \in [L-1]$,
\begin{align*}
    \ddt (\bitl{i}{t}{l} - \tilbitl{i}{t}{l}) =&~ - \beta \Big (\expenu{\xb}{\zeta_{m, t}(\xb) \qitlxb{i}{t}{l}{\xb} \sigpbig{\hitlxb{i}{t}{l}{\xb} + \bitl{i}{t}{l}}} \\
    & \hspace{27pt} - \expenu{\xb}{\zeta_{t}(\xb) \tilqitlxb{i}{t}{l}{\xb} \sigpbig{\tilhitlxb{i}{t}{l}{\xb} + \tilbitl{i}{t}{l}}} \Big ) \\
    =&~ - \beta \expenu{\xb}{\zeta_{m, t}(\xb) \left (\qitlxb{i}{t}{l}{\xb} \sigpbig{\hitlxb{i}{t}{l}{\xb} + \bitl{i}{t}{l}} - \tilqitlxb{i}{t}{l}{\xb} \sigpbig{\tilhitlxb{i}{t}{l}{\xb} + \tilbitl{i}{t}{l}} \right ) } \\
    &~- \beta \expenu{\xb}{(\zeta_{m, t}(\xb) - \zeta_{t}(\xb)) \tilqitlxb{i}{t}{l}{\xb} \sigpbig{\tilhitlxb{i}{t}{l}{\xb} + \tilbitl{i}{t}{l}}}
\end{align*}
Thus,
\begin{align*}
    \left ( \frac{1}{m} \summ{i}{m} \left | \ddt (\bitl{i}{t}{l} - \tilbitl{i}{t}{l}) \right |^2 \right )^{\frac{1}{2}} = &~ O \left ( \expenu{\xb}{\Delta \qitl{m}{t}{l}(\xb) + \Delta \zeta_{m, t}(\xb) + \Delta \zeta_{m, t}(\xb) \Delta \qitl{m}{t}{l}(\xb)} \right )~,
\end{align*}
which implies that
\begin{align*}
    \ddt \bitl{m}{t}{l} =&~ O \left ( \expenu{\xb}{\Delta \qitl{m}{t}{l}(\xb)  + \Delta \zeta_{m, t}(\xb) + \Delta \zeta_{m, t}(\xb) \Delta \qitl{m}{t}{l}(\xb)} \right ) \\
    = &~ O \left ( \Delta_{m, t} + (\Delta_{m, t})^2 \right )
\end{align*}

\item $\Delta W_{m, t}^{(l)}$

For $l \in [L-2]$,
\begin{equation*}
\begin{split}
    &~ \ddt (\Wijtl{i}{j}{t}{l} - \tilWijtl{i}{j}{t}{l}) \\
    =&~ - \expenu{\xb}{\zeta_{m, t}(\xb) \qitlxb{i}{t}{l+1}{\xb} \sigpbig{\hitlxb{i}{t}{l+1}{\xb} + \bitl{i}{t}{l+1}} \sigbig{\hitlxb{j}{t}{l}{\xb} + \bitl{j}{t}{l}}} \\
    &~+ \expenu{\xb}{\zeta_{t}(\xb) \tilqitlxb{i}{t}{l+1}{\xb} \sigpbig{\tilhitlxb{i}{t}{l+1}{\xb} + \tilbitl{i}{t}{l+1}} \sigbig{\tilhitlxb{j}{t}{l}{\xb} + \tilbitl{j}{t}{l}}}  \\
    =&~ - \mathcal{E}_{\xb} \Big \{\zeta_{m, t}(\xb) \left (\qitlxb{i}{t}{l+1}{\xb} \sigpbig{\hitlxb{i}{t}{l+1}{\xb} + \bitl{i}{t}{l+1}} - \tilqitlxb{i}{t}{l+1}{\xb} \sigpbig{\tilhitlxb{i}{t}{l+1}{\xb} + \tilbitl{i}{t}{l+1}} \right ) \\
    &~ \hspace{305pt} \cdot \sigbig{\hitlxb{j}{t}{l}{\xb} + \bitl{j}{t}{l}} \Big \} \\
    &~ - \expenu{\xb}{\zeta_{m, t}(\xb) \tilqitlxb{i}{t}{l+1}{\xb} \sigpbig{\tilhitlxb{i}{t}{l+1}{\xb} + \tilbitl{i}{t}{l+1}} \left (\sigbig{\hitlxb{j}{t}{l}{\xb} + \bitl{j}{t}{l}} - \sigbig{\tilhitlxb{j}{t}{l}{\xb} + \tilbitl{j}{t}{l}} \right )} \\
    &~ - \expenu{\xb}{(\zeta_{m, t}(\xb) - \zeta_{t}(\xb)) \tilqitlxb{i}{t}{l+1}{\xb}\sigbig{\tilhitlxb{j}{t}{l}{\xb} + \tilbitl{j}{t}{l}}}
\end{split}
\end{equation*}
Thus,
\begin{align*}
    &~ \frac{1}{m^2} \summ{i,j}{m} \left ( \ddt (\Wijtl{i}{j}{t}{l} - \tilWijtl{i}{j}{t}{l}) \right )^2 \\
    = &~ O\left ( \expenu{\xb}{(1 + \Delta \zeta_{m, t}(\xb)) (\Delta \qitl{m}{t}{l}(\xb) + \Delta \hitl{m}{t}{l-1}(\xb) + \Delta \bitl{m}{t}{l-1})^2 + \Delta \zeta_{m, t}(\xb)} \right ) 
\end{align*}
and so
\begin{equation*}
\begin{split}
    \ddt \Delta W_{m, t}^{(l)} =&~ O \left ( \expenu{\xb}{(1 + \Delta \zeta_{t}(\xb)) (\Delta \qitl{m}{t}{l}(\xb) + \Delta \hitl{m}{t}{l-1}(\xb) + \Delta \bitl{m}{t}{l-1} + \Delta \zeta_{m, t}(\xb)} \right ) \\
    = &~ O \left ( \Delta_{m, t} + (\Delta_{m, t})^2 \right )~.
\end{split}
\end{equation*}

\item $\Delta \zb_{m, t}$
\begin{align*}
    \ddt \Delta \zb_{m, t} =&~ O \left ( \expenu{\xb}{\Delta \zeta_{m, t}(\xb) + \Delta \qitl{m}{t}{1}(\xb) + \Delta \zeta_{m, t}(\xb) \Delta \qitl{m}{t}{1}(\xb)} \right ) \\
     = &~ O \left ( \Delta_{m, t} + (\Delta_{m, t})^2 \right )~.
\end{align*}

\item $\Delta a_{m, t}$
\begin{align*}
    &~ \ddt \Delta a_{m, t} \\
    =&~ O \left ( \expenu{\xb}{\Delta \zeta_{m, t}(\xb) + \Delta \hitl{m}{t}{L-1}(\xb) + \Delta \bitl{m}{t}{L-1} + \Delta \zeta_{m, t}(\xb) (\Delta \hitl{m}{t}{L-1}(\xb) + \Delta \bitl{m}{t}{L-1})} \right ) \\
     = &~ O \left ( \Delta_{m, t} + (\Delta_{m, t})^2 \right )~.
\end{align*}

\item $\Delta \zeta_{m, t}$
\begin{align*}
    \Delta \zeta_{m, t}(\xb) = &~ O \left (\expenu{\xb}{\Delta a_{m, t} + \Delta \hitl{m}{t}{L-1}(\xb) + \Delta \bitl{m}{t}{L-1} + \Delta a_{m, t} (\Delta \hitl{m}{t}{L-1}(\xb) + \Delta \bitl{m}{t}{L-1} )} \right ) \\
    = &~ O \left ( \Delta_{m, t} + (\Delta_{m, t})^2 \right )~.
\end{align*}
\end{enumerate}
Therefore, for the duration in which $\Delta_{m, t} \leq 1$, it holds that
\begin{align*}
    \ddt \Delta_{m, t}
    =&~  O(\Delta_{m, t}) +  o_{\PP}(1)~,
\end{align*}
and hence, by Gr\"onwall's inequality,
\begin{align}
\label{eq:Delta_mt_bound}
    \Delta_{m, t} = o_{\PP}(1)~.
\end{align}
Thus, for any finite $t \geq 0$, when $m$ is large enough, we can always ensure that $\Delta_{m, t} \leq 1$. This implies that \eqref{eq:Delta_mt_bound} holds for all finite $t \geq 0$.

Finally, applying \eqref{eq:Delta_h_mt_bound_by_Delta_mt} to each $\xb \in \{ \xbg{1}, ..., \xbg{N} \}$~, we arrive at Lemma~\ref{lem:Delta_h_mt_N}.

\subsection{Derivation of the Training Dynamics of Deep Linear NNs}
\label{app:linear}
In the linear NN case,
\begin{equation*}
    \begin{split}
        &~ \kappa^{(l)}_{t, s}(\xb, \xb') \\
        =&~ \EEElr{\Htl{t}{l}(\xb) \Htl{s}{l}(\xb')} \\
        =&~ \EEE \bigg[ \left (\Htl{0}{l}(\xb) - \int_{0}^t \expenu{\xb''}{\zeta_r(\xb'') \kappa^{(l-1)}_{t, r}(\xb, \xb'') \Qtl{r}{l}(\xb'')} dr \right ) \\
        & \hspace{20pt} \left (\Htl{0}{l}(\xb') - \int_{0}^s \expenu{\xb''}{\zeta_r(\xb'') \kappa^{(l-1)}_{s, r}(\xb, \xb'') \Qtl{r}{l}(\xb'')} dr \right )  \bigg ] \\
        =&~ \kappa^{(l)}_{0, 0}(\xb, \xb') + \int_{0}^t \int_{0}^s \expenu{\xb'', \xb'''}{\zeta_r(\xb'') \zeta_p(\xb''') \gamma^{(l)}_{r, p}(\xb'', \xb''') \kappa^{(l-1)}_{t, r}(\xb, \xb'') \kappa^{(l-1)}_{s, p}(\xb', \xb''')} dr ~ dp \\
        &~ - \int_{0}^t \expenu{\xb''}{\zeta_r(\xb'') \kappa^{(l-1)}_{t, r}(\xb, \xb'') \EEElb{l}{\Htl{0}{l}(\xb') \Qtl{r}{l}(\xb'')}} dr \\
        &~ - \int_{0}^s \expenu{\xb''}{\zeta_r(\xb'') \kappa^{(l-1)}_{s, r}(\xb', \xb'') \EEElb{l}{\Htl{0}{l}(\xb) \Qtl{r}{l}(\xb'')}} dr 
    \end{split}
\end{equation*}
Since
\begin{equation*}
    \begin{split}
        \EEElr{\Htl{0}{l}(\xb) \Qtl{r}{l}(\xb'')} =&~ \EEElr{\Htl{0}{l}(\xb') \int_{0}^r \expenu{\xb'''}{\zeta_p(\xb''') \gamma^{(l+1)}_{r, p}(\xb'', \xb''') \Htl{p}{l}(\xb''')} dp} \\
        =&~ \int_{0}^r \expenu{\xb'''}{\zeta_p(\xb''') \gamma^{(l+1)}_{r, p}(\xb'', \xb''') \kappa^{(l)}_{0, p}(\xb, \xb''') dp }~,
    \end{split}
\end{equation*}
we then have
\begin{equation*}
    \begin{split}
        &~ \kappa^{(l)}_{t, s}(\xb, \xb') \\
        =&~ \kappa^{(l)}_{0, 0}(\xb, \xb') + \int_{0}^t \int_{0}^s \expenu{\xb'', \xb'''}{\zeta_r(\xb'') \zeta_p(\xb''') \gamma^{(l)}_{r, p}(\xb'', \xb''') \kappa^{(l-1)}_{t, r}(\xb, \xb'') \kappa^{(l-1)}_{s, p}(\xb', \xb''')} dr ~ dp \\
        &~ - \int_{0}^t \expenu{\xb'', \xb'''}{\zeta_r(\xb'') \zeta_p(\xb''') \gamma^{(l+1)}_{r, p}(\xb'', \xb''')  \kappa^{(l-1)}_{t, r}(\xb, \xb'') \kappa^{(l)}_{0, p}(\xb', \xb''')} dr \\
        &~ - \int_{0}^s \expenu{\xb'', \xb'''}{\zeta_r(\xb'') \zeta_p(\xb''') \gamma^{(l+1)}_{r, p}(\xb'', \xb''')  \kappa^{(l-1)}_{s, r}(\xb', \xb'') \kappa^{(l)}_{0, p}(\xb, \xb''')} dr ~.
    \end{split}
\end{equation*}
Notice that $\kappa^{(l)}_{0, p}(\xb, \xb') = 0$, $\forall l > 1, \forall p \geq 0$ while $\kappa^{(1)}_{0, 0}(\xb, \xb') = \kappa^{(0)}_{0, p}(\xb, \xb') = \xb^{\intercal} \cdot \xb'$, $\forall p \geq 0$. Thus, using the linearity, we can derive \eqref{eq:K} for $l \in [L-2]$, and moreover, 
\begin{align*}
    K^{(1)}_{t, s} =&~ 1 + \int_{0}^t \int_{0}^s c^{(1)}_{r, p} \zetab_r \cdot \zetab_p^{\intercal} ~ dp ~dr~ \\
        &~ - \int_{0}^t \int_{0}^r c^{(2)}_{r, p} \zetab_r \cdot \zetab_p^{\intercal} \cdot K^{(1)}_{p, 0} ~ dp ~dr~ \\
        &~ - \int_{0}^s \int_{0}^r c^{(2)}_{r, p} K^{(1)}_{0, p} \cdot \zetab_p \cdot \zetab_r^{\intercal} ~ dp ~dr~.
\end{align*}
With a similar argument, we can derive \eqref{eq:c} for $l \in [L-2]$, and moreover,
\begin{align*}
        c^{(L-1)}_{t, s} =&~ 1 +  \int_{0}^t \int_{0}^s \zetab_r^{\intercal} \cdot K^{(L-1)}_{r, p} \cdot \zetab_p^{\intercal} ~dp ~dr \\
        &~ - \int_{0}^t \int_{0}^r c^{(L-1)}_{p, 0} \zetab_r^{\intercal} \cdot K^{(L-2)}_{r, p} \cdot \zetab_p^{\intercal} ~dp ~dr \\
        &~ - \int_{0}^s \int_{0}^r c^{(L-1)}_{p, 0} \zetab_r^{\intercal} \cdot K^{(L-2)}_{r, p} \cdot \zetab_p^{\intercal} ~dp ~dr~. 
\end{align*}

\section{Supplementary Materials for Section~\ref{sec:depth}}
\subsection{Proof of Lemma~\ref{lem:relu_grows}}
\label{app:pf_lem_relu_grows}

For any function $f \in \Ladder{L}$, we let $(\Hilb^{(l)})_{l \in [L]}$ be an $L$-level NHL that it belongs to. Hence, there exist $\mu^{(1)}$, ..., $\mu^{(L-1)}$ that satisfy the condition given in Definition~\ref{def:nhl}, and furthermore, there exists $\xi \in L^2(\Hilb^{(L-1)}, \mu^{(L-1)})$ such that
\begin{equation}
\label{eq:xi_L-1}
    f(\xb) = \int_{\Hilb^{(L-1)}} \xi(h) \sigbig{h(\xb)} \mu^{(L-1)}(dh)~, \quad \forall \xb \in \domX~.
\end{equation}
We can decompose $\xi = \xi_+ - \xi_-$, where $\xi_+$ and $\xi_- \in L^2(\Hilb^{(L-1)}, \mu^{(L-1)})$ are both non-negative on $\Hilb^{(L-1)}$, and then define
\begin{align*}
    f_+(\xb) \coloneqq&~ \int_{\Hilb^{(L-1)}} \xi_+(h) \sigbig{h(\xb)} \mu^{(L-1)}(dh)~, \quad \forall \xb \in \domX~, \\
    f_-(\xb) \coloneqq&~ \int_{\Hilb^{(L-1)}} \xi_-(h) \sigbig{h(\xb)} \mu^{(L-1)}(dh)~, \quad \forall \xb \in \domX~.
\end{align*}
Since ReLU has non-negative outputs, we see that $f_+$ and $f_-$ are both non-negative functions on $\domX$. Furthermore, they satisfy $f = f_+ - f_-$, $\| f_+ \|_{\Hilb^{(L)}}^2 \leq \int_{\Hilb^{(L-1)}} |\xi_+(h)|^2 \mu^{(L-1)}(dh)$ and $\| f_- \|_{\Hilb^{(L)}}^2 \leq \int_{\Hilb^{(L-1)}} |\xi_-(h)|^2 \mu^{(L-1)}(dh)$, and hence 
\begin{equation*}
    \| f_+ \|_{\Hilb^{(L)}}^2 + \| f_- \|_{\Hilb^{(L)}}^2 \leq \int_{\Hilb^{(L-1)}} |\xi(h)|^2 \mu^{(L-1)}(dh)~.
\end{equation*}
Thus, we can further define $\mu^{(L)} \coloneqq \frac{1}{2} \delta_{f_+} + \frac{1}{2} \delta_{f_-} \in \Pcal(\Hilb^{(L)})$ as well as $\bar{\xi} \in L^2(\Hilb^{(L)}, \mu^{(L)})$ as $\bar{\xi} = 2 \mathds{1}_{f_+} - 2 \mathds{1}_{f_-}$, and let $\Hilb^{(L+1)} = \Hilb_{\mu^{(L)}}$. It holds that
\begin{equation*}
    \begin{split}
        \int_{\Hilb^{(L)}} \bar{\xi}(h) \sigbig{h(\xb)} \mu^{(L)}(dh) =&~ \frac{1}{2} \cdot 2 \sigbig{f_+(\xb)} - \frac{1}{2} \cdot 2 \sigbig{f_-(\xb)} = f(\xb)~.
    \end{split}
\end{equation*}
and therefore,
\begin{equation*}
\begin{split}
    \| f \|_{\Hilb^{(L+1)}}^2 \| \mu^{(L)} \|_{\Hilb^{(L)}, 2}^2 \leq &~ \left ( \int_{\Hilb^{(L)}} |\bar{\xi}(h)|^2 \mu^{(L)}(dh) \right ) \| \mu^{(L)} \|_{\Hilb^{(L)}, 2}^2 \\
    \leq &~ 4 \cdot \frac{1}{2} (\| f_+ \|_{\Hilb^{(L)}}^2 + \| f_- \|_{\Hilb^{(L)}}^2) \\
    \leq &~ 2 \int_{\Hilb^{(L-1)}} |\xi(h)|^2 \mu^{(L-1)}(dh)~.
\end{split}
\end{equation*}
By optimizing over all $\xi$ such that \eqref{eq:xi_L-1} holds, we derive that the left-hand side can be upper-bounded by $2 \| f \|_{\Hilb^{(L)}}^2$. Therefore, $f$ also belongs to the $(L+1)$-level NHL,  $(\Hilb^{(l)})_{l \in [L+1]}$, with $\CompL{L+1}(f) \leq \sqrt{2} \CompL{L}(f)$.

\end{document}